\newcommand{\shortonly}{0} 

\newcommand{\shortversion}[1]{\ifthenelse{\equal{\shortonly}{1}}{#1}{}}
\newcommand{\longversion}[1]{\ifthenelse{\equal{\shortonly}{1}}{}{#1}}

\documentclass{article}
\usepackage{ifthen}
\usepackage{msrtr}\msrtrno{2001-72}

\longversion{\usepackage{fullname}}

\newcommand{\mymarginpar}[1]{\marginpar{\raggedright #1}}
\newcommand{\longversionboring}[1]{\ifthenelse{\equal{\shortonly}{1}}{}{\mymarginpar{Begin boring details}#1\mymarginpar{End boring details}}}

\input{psfig}

\shortversion{\newcommand{\namecite}{\citet}}
\shortversion{\newcommand{\mycite}{\citep}}
\longversion{\newcommand{\mycite}{\cite}}

\shortversion{
\title{A Bit of Progress in Language Modeling}
\author{Joshua T. Goodman}
\address{Machine Learning and Applied Statistics Group \\
         Microsoft Research \\
         One Microsoft Way \\
         Redmond, WA 98052 \\
         {\tt joshuago@microsoft.com} }

\shortauthor{J. Goodman}
\shorttitle{A Bit of Progress in Language Modeling}

\date{August 8, 2001}

\usepackage{natbib}
}

\longversion{
\title{A Bit of Progress in Language Modeling\\ Extended Version}
\author{Joshua T. Goodman\\
Machine Learning and Applied Statistics Group \\
         Microsoft Research \\
         One Microsoft Way \\
         Redmond, WA 98052 \\
         {\tt joshuago@microsoft.com} }
}

\newcommand{\Px}[1]{P_{\mbox{\tt \footnotesize #1}}}
\newcommand{\Psmooth}{P_{\mbox{\tt \footnotesize Smooth}}}
\newcommand{\Pkatz}{P_{\mbox{\tt \footnotesize Katz}}}
\newcommand{\Pbkn}{P_{\mbox{\tt \footnotesize BKN}}}
\newcommand{\Pikn}{P_{\mbox{\tt \footnotesize IKN}}}

\newcommand{\Pkn}{P_{\mbox{\tt \footnotesize KN}}}
\newcommand{\text}{\it}
\newcommand{\mboxtext}[1]{\mbox{\text #1}}

\newcommand{\un}{\mbox{\underline{ }}}
\newcommand{\myetal}{{et al.\ }}

\begin{document}
\maketitle
\begin{abstract}

In the past several years, a number of different language modeling
improvements over simple trigram models have been found, including
caching, higher-order n-grams, skipping, interpolated Kneser-Ney
smoothing, and clustering.  We present explorations of variations on,
or of the limits of, each of these techniques, including showing that
sentence mixture models may have more potential.  While all of these
techniques have been studied separately, they have rarely been studied
in combination.  \longversion{We find some significant interactions,
especially with smoothing and clustering techniques.}  We compare a
combination of all techniques together to a Katz smoothed trigram
model with no count cutoffs.  We achieve perplexity reductions between
38\% and 50\% (1 bit of entropy), depending on training data size, as
well as a word error rate reduction of 8.9\%.
Our perplexity reductions are perhaps the highest reported compared to
a fair baseline.  \longversion{This is the extended version of the
paper; it contains additional details and proofs, and is designed to
be a good introduction to the state of the art in language modeling.}
\end{abstract}

\section{Introduction}

\subsection{Overview}

Language modeling is the art of determining the probability of a
sequence of words.  This is useful in a large variety of areas
including speech recognition, optical character recognition,
handwriting recognition, machine translation, and spelling correction
\mycite{Church:88a,Brown:90b,Hull:92a,Kernighan:90a,Srihari:92a}.  The
most commonly used language models are very simple (e.g.\ a
Katz-smoothed trigram model).  There are many improvements over this
simple model however, including caching, clustering, higher-order
n-grams, skipping models, and sentence-mixture models, all of which we
will describe below.  Unfortunately, these more complicated techniques
have rarely been examined in combination.  It is entirely possible
that two techniques that work well separately will not work well
together, and, as we will show, even possible that some techniques
will work better together than either one does by itself.  In this
paper, we will first examine each of the aforementioned techniques
separately, looking at variations on the technique, or its limits.
Then we will examine the techniques in various combinations, and
compare to a Katz smoothed trigram with no count cutoffs.  On a small
training data set, 100,000 words, we can get up to a 50\% perplexity
reduction, which is one bit of entropy.  On larger data sets, the
improvement declines, going down to 41\% on our largest data set,
284,000,000 words.  On a similar large set without punctuation, the
reduction is 38\%.  On that data set, we achieve an 8.9\% word error
rate reduction.  These are perhaps the largest reported perplexity
reductions for a language model, versus a fair baseline.

The paper is organized as follows.  First, in this section, we will
describe our terminology, briefly introduce the various techniques we
examined, and describe our evaluation methodology.  In the following
sections, we describe each technique in more detail, and give
experimental results with variations on the technique, determining for
each the best variation, or its limits.  In particular, for caching,
we show that trigram caches have nearly twice the potential of unigram
caches.  For clustering, we find variations that work slightly better
than traditional clustering, and examine the limits.  For n-gram
models, we examine up to 20-grams, but show that even for the largest
models, performance has plateaued by 5 to 7 grams.  For skipping
models, we give the first detailed comparison of different skipping
techniques, and the first that we know of at the 5-gram level.  For
sentence mixture models, we show that mixtures of up to 64 sentence
types can lead to improvements.  We then give experiments comparing
all techniques, and combining all techniques in various ways.  All of
our experiments are done on three or four data sizes, showing which
techniques improve with more data, and which get worse.  In the
concluding section, we discuss our results.  \longversion{Finally, in
the appendices, we give a proof helping to justify Kneser-Ney
smoothing and we describe implementation tricks and details for
handling large data sizes, for optimizing parameters, for clustering,
and for smoothing. }

\shortversion{There is also an extended version of this paper
\mycite{Goodman:01b} that goes into much more detail than this
version.  The extended version contains more tutorial information,
more details about the experiments presented here, interesting
implementation details, and a few additional experiments and proofs.
It is meant to be a reasonable introduction to the field of
language modeling.}

\subsection{Technique introductions}

The goal of a language model is to determine the probability of a word
sequence $w_1 ... w_n$, $P(w_1 ... w_n)$.  This probability is
typically broken down into its component probabilities:
$$
P(w_1 ... w_i) = P(w_1) \times P(w_2 | w_1) \times ... \times P(w_i | w_1...w_{i-1})
$$
Since it may be difficult to compute a probability of the form $P(w_i
| w_1...w_{i-1})$ for large $i$, we typically assume that the
probability of a word depends on only the two previous words, the {\em
trigram} assumption:
$$
P(w_i | w_1...w_{i-1}) \approx P(w_i | w_{i-2} w_{i-1})
$$
which has been shown to work well in practice.  The trigram
probabilities can then be estimated from their counts in a training
corpus.  We let $C(w_{i-2} w_{i-1} w_i)$ represent the number of
occurrences of $w_{i-2} w_{i-1} w_i$ in our training corpus, and
similarly for $C(w_{i-2} w_{i-1})$.  Then, we can approximate:
$$
P(w_i | w_{i-2} w_{i-1}) \approx \frac{C(w_{i-2} w_{i-1} w_i)}{C(w_{i-2} w_{i-1})}
$$
Unfortunately, in general this approximation will be very noisy,
because there are many three word sequences that never occur.
Consider, for instance, the sequence {\it ``party on Tuesday''}.  What
is $P(\mboxtext{Tuesday} | \mboxtext{party on})$?  Our training corpus
might not contain any instances of the phrase, so $C(\mboxtext{party on
Tuesday})$ would be 0, while there might still be 20 instances of the
phrase {\it ``party on''}.  Thus, we would predict
$P(\mboxtext{Tuesday} | \mboxtext{party on}) = 0$, clearly an
underestimate.  This kind of 0 probability can be very problematic in
many applications of language models.  For instance, in a speech
recognizer, words assigned 0 probability cannot be recognized no
matter how unambiguous the acoustics.

{\em Smoothing} techniques take some probability away from some
occurrences.  Imagine that we have in our training data a single
example of the phrase 
{\it ``party on Stan Chen's
birthday''}.\longversion{\footnote{\tiny Feb. 25.  Stan will provide cake.
You are all invited.}}
Typically,
when something occurs only one time, it is greatly overestimated.  In
particular,
$$
P(\mboxtext{Stan} | \mboxtext{party on}) \ll \frac{1}{20} = \frac{C(\mboxtext{party on Stan})}{C(\mboxtext{party on})}
$$
By taking some probability away from some words, such as {\text
``Stan''} and redistributing it to other words, such as {\text
``Tuesday''}, zero probabilities can be avoided.  In a smoothed
trigram model, the extra probability is typically distributed
according to a smoothed bigram model, etc.  While the most commonly
used smoothing techniques, Katz smoothing \mycite{Katz:87a} and
Jelinek-Mercer smoothing \mycite{Jelinek:80a} (sometimes called
deleted interpolation) work fine, even better smoothing techniques
exist.  In particular, we have previously shown \mycite{Chen:99a} that
versions of Kneser-Ney smoothing \mycite{Ney:94a} outperform all other
smoothing techniques.  In the appendix\shortversion{ of the extended
version of this paper}, we give a proof partially explaining this
optimality.  In Kneser-Ney smoothing, the backoff distribution is
modified: rather than a normal bigram distribution, a special
distribution is used.  Using Kneser-Ney smoothing instead of more
traditional techniques is the first improvement we used.

The most obvious extension to trigram models is to simply move to
{\em higher-order n-grams}, such as four-grams and five-grams.  We will show
that in fact, significant improvements can be gotten from moving to
five-grams.  Furthermore, in the past, we have shown that there is a
significant interaction between smoothing and n-gram order
\mycite{Chen:99a}: higher-order n-grams work better with Kneser-Ney
smoothing than with some other methods, especially Katz smoothing.  We
will also look at how much improvement can be gotten from higher order
n-grams, examining up to 20-grams.

Another simple extension to n-gram models is {\em skipping} models
\mycite{Rosenfeld:94a,Huang:93a,Ney:94a}, in which we condition on
a different context than the previous two words.  For instance, instead
of computing $P(w_i | w_{i-2} w_{i-1})$, we could instead compute
$P(w_i | w_{i-3} w_{i-2})$.  This latter model is probably not as
good, but can be combined with the standard model to yield some
improvements.

{\em Clustering} (also called {\em classing}) models attempt to make
use of the similarities between words.  For instance, if we have seen
occurrences of phrases like ``{\text party on Monday}'' and ``{\text
party on Wednesday}'', then we might imagine that the word ``{\text
Tuesday}'', being similar to both ``{\text Monday}'' and ``{\text Wednesday}'', is
also likely to follow the phrase ``{\text party on}.''  The majority
of the previous research on word clustering has focused on how to get
the best clusters.  We have concentrated our research on the best way
to {\em use} the clusters, and will report results showing some novel
techniques that work a bit better than previous methods.  \longversion{We also show
significant interactions between clustering and smoothing.}

{\em Caching} models \mycite{Kuhn:88a,Kuhn:90a,Kuhn:92a} make use of the
observation that if you use a word, you are likely to use it again.
They tend to be easy to implement and to lead to relatively large
perplexity improvements, but relatively small word-error rate
improvements.  We show that by using a trigram cache, we can get
almost twice the improvement as from a unigram cache.

{\em Sentence Mixture} models \mycite{Iyer:99a,Iyer:94a} make use of the
observation that there are many different sentence types, and that
making models for each type of sentence may be better than using one
global model.  Traditionally, only 4 to 8 types of sentences are used,
but we show that improvements can be gotten by going to 64 mixtures,
or perhaps more.

\subsection{Evaluation}

In this section, we first describe and justify our use of perplexity
or entropy as an evaluation technique.  We then describe the data and
experimental techniques used in the experiments in the following
sections.

\longversion{
The most commonly used method for measuring language model performance
is {\em perplexity}.  A language model that assigned equal probability
to 100 words would have perplexity 100.  In general, the perplexity of
a language model is equal to the geometric average of the inverse
probability of the words measured on test data:
$$
\sqrt[N]{\prod_{i=1}^N \frac{1}{P(w_i | w_1...w_{i-1})}}
$$
Perplexity has many properties that make it attractive as a measure of
language model performance; among others, the ``true'' model for any
data source will have the lowest possible perplexity for that source.
Thus, the lower the perplexity of our model, the closer it is, in some
sense, to the true model.  While several alternatives to perplexity
have been shown to correlate better with speech recognition
performance, they typically have free variables that need to be
optimized for a particular speech recognizer; others are significantly
more difficult to compute in our framework.

An alternative, but equivalent measure to perplexity is entropy, which
is simply $\log_2$ of perplexity.  Entropy has the nice property that
it is the average number of bits per word that would be necessary to
encode the test data using an optimal coder.  For those familiar with
information theory, what we actually measure is the cross entropy of
the test data given the model.  Since this is by far the most common
type of entropy measured, we abuse the term by simply saying entropy,
when what we really mean is a particular cross entropy.
}
\shortversion{
We will primarily measure our performance by the entropy of test data
as given by the model (which should be called cross-entropy):
$$
\frac{1}{N}\sum_{i=1}^N{-\log_2 P(w_i | w_1...w_{i-1})}
$$
Entropy has several nice properties.  First, it is the average number of bits
that would be required to encode the test data using an optimal coder.
Also, assuming the test data is generated by some random process, a
perfect model of this process would have the lowest possible entropy,
so the lower the entropy, the closer we are, in some sense, to this
true model.  Sometimes, we will also measure perplexity, which is
simply $2^{\mbox{\em entropy}}$, and corresponds to the weighted average
number of choices for each word.  Several alternatives to
entropy have been shown to correlate better with speech recognition
performance, but they are typically speech-recognizer-specific and much
harder to compute in our framework.}

\longversion{We will use both entropy and perplexity measurements in
this paper: entropy reductions have several nice properties, including
being additive and graphing well, but perplexity reductions are more
common in the literature.\footnote{Probably because it sounds much better to get a 20\% perplexity reduction than to get a 0.32 bit entropy reduction.}  The following table may be helpful.  Notice
that the relationship between entropy and perplexity reductions is
roughly linear up through about .2 bits.
$$
\begin{tabular}{llllllllllll} 
reduction \\
entropy  & .01      &  .1   & .16 &  .2  &  .3 & .4 & .5   & .75  & 1 \\
perplexity & 0.69\% & 6.7\% & 10\% & 13\% & 19\% & 24\% & 29\% & 41\% & 50\% 
\end{tabular}
$$
} 

All of our experiments were performed on the NAB (North American
Business news) corpus \mycite{Stern:96a}.  We performed most experiments at 4 different
training data sizes: 100,000 words, 1,000,000 words, 10,000,000 words,
and the whole corpus -- except 1994 WSJ data -- approximately
284,000,000 words.  In all cases, we performed parameter optimization
on a separate set of heldout data, and then performed testing on a set
of test data.  None of the three data sets overlapped.  The heldout
and test sets were always every fiftieth sentence from two
non-overlapping sets of 1,000,000 words, taken from the 1994 section.
In the appendix, we describe implementation tricks we used; these
tricks made it possible to train very complex models on very large
amounts of training data, but made it hard to test on large test sets.
For this reason, we used only 20,000 words total for testing or
heldout data.  On the other hand, we did not simply want to use, say,
a 20,000 word contiguous test or heldout set, since this would only
constitute a few articles, and thus risk problems from too much
homogeneity; thus we chose to use every 50'th sentence from
non-overlapping 1,000,000 word sets.  All of our experiments were done
using the same 58,546 word vocabulary.  End-of-sentence,
end-of-paragraph, and end-of-article symbols were included in perplexity
computations, but out-of-vocabulary words were not.

It would have been interesting to try our experiments on other
corpora, as well as other data sizes.  In our previous work
\mycite{Chen:99a}, we compared both across corpora and across data
sizes.  We found that different corpora were qualitatively similar,
and that the most important differences were across training data
sizes.  We therefore decided to concentrate our experiments on
different training data sizes, rather than on different corpora.

Our toolkit is unusual in that it allows all parameters to be jointly
optimized.  In particular, when combining many techniques, there are
many interpolation and smoothing parameters that need to be optimized.
We used Powell's algorithm \mycite{Press:88a} over the heldout data to
jointly optimize all of these parameters.

\section{Smoothing}
\longversion{\mymarginpar{This section is a recap of Chen and Goodman (1999).  If you have read that, skip this.}}
There are many different smoothing techniques that can be used, and
the subject is a surprisingly subtle and complicated one.  Those
interested in smoothing should consult our previous work
\mycite{Chen:99a}, where detailed descriptions and detailed comparisons
of almost all commonly used smoothing algorithms are done.  We will
limit our discussion here to four main techniques: simple
interpolation, Katz smoothing, Backoff Kneser-Ney smoothing, and
Interpolated Kneser-Ney smoothing.  In this section, we describe those
four techniques, and recap previous results, including the important
result that Interpolated Kneser-Ney smoothing, or minor variations on
it, outperforms all other smoothing techniques.

The simplest way to combine techniques in language modeling is to
simply interpolate them together.  For instance, if one has a trigram
model, a bigram model, and a unigram model, one can use
\begin{eqnarray*}
\lefteqn{\Px{interpolate}(w|w_{i-2}w_{i-1}) = }\\
&& \lambda
\Px{trigram}(w|w_{i-2}w_{i-1}) + (1-\lambda)[ \mu \Px{bigram}(w|w_{i-1}) +
(1-\mu) \Px{unigram}(w)]
\end{eqnarray*}
where $\lambda$ and $\mu$ are constants such that $0 \leq \lambda, \mu
\leq 1$. \longversion{In practice, we also interpolate with the
uniform distribution $\Px{uniform}(w) = \frac{1}{\mbox{size of
vocabulary}}$; this ensures that no word is assigned probability 0.
Also, we need to deal with the case when, for instance, the trigram
context $w_{i-2}w_{i-1}$ has never been seen, $C(w_{i-2}w_{i-1})=0$.
In this case, we use an interpolated bigram model, etc.}  Given its
simplicity, simple interpolation works surprisingly well, but other
techniques, such as Katz smoothing, work even better.

\newcommand{\disc}{\mbox{\it disc}}

Katz smoothing \mycite{Katz:87a} is based on the Good-Turing formula
\mycite{Good:53a}.  Notice that if a particular word sequence
(i.e. ``{\text party on Stan}'') occurs only once (out of perhaps a
billion words) it is probably significantly overestimated -- it
probably just showed up by chance, and its true probability is much
less than one one billionth.  It turns out that the same thing is true
to a lesser degree for sequences that occurred twice, and so on.  Let
$n_r$ represent the number of n-grams that occur $r$ times, i.e.
$$
n_r = \vert \{ w_{i-n+1}...w_i | C(w_{i-n+1}...w_i) = r \} \vert
$$
Good proved that under some very weak assumptions that for any n-gram
that occurs $r$ times, we should {\it discount} it, pretending that it occurs
$\disc(r)$ times where
$$
\disc(r) = (r+1)\frac{n_{r+1}}{n_r}
$$
($\disc(r)$ is more typically written as $r^*$).  In language
modeling, the estimate $\disc(r)$ will almost always be less
than $r$.  This will leave a certain amount of probability
``left-over.''  In fact, letting $N$ represent the total size of the
training set, this left-over probability will be equal to
$\frac{n_1}{N}$; this represents the amount of probability to be
allocated for events that were never seen. \longversion{This is really
quite an amazing and generally useful fact, that we can predict how
often we expect something to happen that has never happened before, by
looking at the proportion of things that have occurred once.}

For a given context, Katz smoothing uses one of two formulae.  If the
word sequence $w_{i-n+1}...w_i$ has been seen before, then Katz
smoothing uses the discounted count of the sequence, divided by the
the counts of the context $w_{i-n+1}...w_{i-1}$.  On the other hand, if
the sequence has never been seen before, then we back off to the next
lower distribution, $w_{i-n+2}...w_{i-1}$.  Basically,
we use the following
formula:
\begin{eqnarray*}
\lefteqn{\Pkatz(w_i|w_{i-n+1}...w_{i-1}) } \\
&=&\left\{
\begin{array}{ll}
\frac{\disc(C(w_{i-n+1}...w_i))}{C(w_{i-n+1}...w_{i-1})} &
\mbox{if $C(w_{i-n+1}...w_i) > 0$} \\ \\
\alpha(w_{i-n+1}...w_{i-1}) \times \Pkatz(w_i|w_{i-n+1}...w_{i-1}) &
\mbox{otherwise}
\end{array}
\right.
\end{eqnarray*}
where $\alpha(w_{i-n+1}...w_{i-1})$ is a normalization constant chosen
so that the probabilities sum to 1.\footnote{\namecite{Chen:99a} as
well as the appendix \shortversion {of the extended version of this
paper} give the details of our implementation of Katz smoothing.
Briefly, we also smooth the unigram distribution using additive
smoothing; we discount counts only up to $k$, where we determine $k$
to be as large as possible, while still giving reasonable discounts
according to the Good-Turing formula; we add pseudo-counts $\beta$ for
any context with no discounted counts.  Tricks are used to estimate
$n_r$.}

Katz smoothing is one of the most commonly used smoothing techniques,
but it turns out that other techniques work even better.
\namecite{Chen:99a} performed a detailed comparison of many smoothing techniques
and found that a modified interpolated form of
Kneser-Ney smoothing \mycite{Ney:94a} consistently outperformed all
other smoothing techniques.  The basic insight behind Kneser-Ney
smoothing is the following.  Consider a conventional bigram model of a
phrase such as $\Pkatz(\mboxtext{Francisco}|\mboxtext{on})$.  Since
the phrase {\text San Francisco} is fairly common, the conventional
unigram probability (as used by Katz smoothing or techniques like deleted
interpolation) $\frac {C(\mboxtext{Francisco})}{\sum_w C(w)}$ will
also be fairly high.  This means that using, for instance, a 
model such as Katz smoothing, the probability
\begin{eqnarray*}
\Pkatz(\mboxtext{on Francisco}) &=&
 \left\{
\begin{array}{ll}
\frac{\disc(C(\mboxtext{on Francisco}))}{C(\mboxtext{on})} &
\mbox{if $C(\mboxtext{on Francisco}) > 0$} \\
\alpha(\mboxtext{on})\!\times\! \Pkatz(\mboxtext{Francisco}) &
\mbox{otherwise}
\end{array}
\right. \\
&=&
\alpha(\mboxtext{on}) \times \Pkatz(\mboxtext{Francisco})
\end{eqnarray*}
will also be fairly high.  But, the word {\text Francisco} occurs in
exceedingly few contexts, and its probability of occuring in a new one
is very low.  Kneser-Ney smoothing uses a modified backoff
distribution based on the number of contexts each word occurs in,
rather than the number of occurrences of the word.  Thus, a
probability such as $\Pkn(\mboxtext{Francisco}|\mboxtext{on})$
would be fairly low, while for a word like {\text Tuesday} that occurs in
many contexts, $\Pkn(\mboxtext{Tuesday}|\mboxtext{on})$ would be
relatively high, even if the phrase {\text on Tuesday} did not
occur in the training data.  Kneser-Ney smoothing also uses a simpler
discounting scheme than Katz smoothing: rather than computing the
discounts using Good-Turing, a single discount, $D$, (optimized on
held-out data) is used.  In particular, Backoff Kneser-Ney smoothing
uses the following formula (given here for a bigram) where $\vert \{ v
| C(v w_i) > 0 \} \vert$ is the number of words $v$ that $w_i$ can
occur in the context of.
$$
\Pbkn(w_i|w_{i-1}) = 
\left\{
\begin{array}{ll}
\frac{C(w_{i-1} w_i)-D}{C(w_{i-1})} & 
\mbox{if $C(w_{i-1} w_i) > 0$} \\
\alpha(w_{i-1}) 
\frac{\vert \{ v | C(v w_i) > 0 \} \vert}
     {\sum_w \vert \{ v | C(v w) > 0 \} \vert} &
\mbox{otherwise}
\end{array}
\right.
$$
Again, $\alpha$ is a normalization constant such that the
probabilities sum to 1.  The formula can be easily extended to higher
order n-grams in general.  For instance, for trigrams, both the
unigram and bigram distributions are modified.  

\namecite{Chen:99a} showed that methods like Katz smoothing and
Backoff Kneser-Ney smoothing that backoff to lower order distributions
only when the higher order count is missing do not do well on low
counts, such as one counts and two counts.  This is because the
estimates of these low counts are fairly poor, and the estimates ignore
useful information in the lower order distribution.  {\it
Interpolated} models always combine both the higher-order and the
lower order distribution, and typically work better.  In particular,
the {\longversion basic} formula for Interpolated Kneser-Ney smoothing is
$$
\begin{array}{@{}l}
\Pikn(w_i|w_{i-1}) = 
\frac{C(w_{i-1}w_i)-D}{C(w_{i-1})} + 
\lambda(w_{i-1})\frac{\vert \{ v | C(v w_i) > 0 \} \vert}
{\sum_w \vert \{ v | C(v w) > 0 \} \vert}
\end{array}
$$
where $\lambda(w_{i-1})$ is a normalization constant such that the
probabilities sum to 1.  \namecite{Chen:99a} proposed one additional
modification to Kneser-Ney smoothing, the use of multiple discounts,
one for one counts, another for two counts, and another for three or
more counts.  This formulation, Modified Kneser-Ney smoothing,
typically works slightly better than Interpolated Kneser-Ney.
However, in our experiments on combining techniques, it would have
nearly tripled the number of parameters our system needed to search,
and in a pilot study, when many techniques were combined, it did not
work better than Interpolated Kneser-Ney.  Thus, in the rest of this
paper, we use Interpolated Kneser-Ney instead of Modified Kneser-Ney.
In the \longversion{appendix,}\shortversion{appendix of the long
version of this paper,} we give a few more details about our
implementation of our smoothing techniques, including standard
refinements used for Katz smoothing.  We also give arguments
justifying Kneser-Ney smoothing, and example code, showing that
interpolated Kneser-Ney smoothing is easy to implement.

\longversion{For completeness, we show the exact formula used for an
interpolated Kneser-Ney smoothed trigram.  In practice, to avoid zero
probabilities, we always smooth the unigram distribution with the
uniform distribution, but have ommitted unigram smoothing from other
formulas for simplicity; we include it here for completeness.  Let
$\vert V \vert$ represent the size of the vocabulary.

$$
\begin{array}{@{}l}
\Pikn(w_i|w_{i-2}w_{i-1}) =
\frac{C(w_{i-2}w_{i-1}w_i)-D_3}{C(w_{i-2}w_{i-1})} +
\lambda(w_{i-2}w_{i-1}) \Px{ikn-mod-bigram}(w_i|w_{i-1})
\end{array}
$$
$$
\begin{array}{@{}l}
\Px{ikn-mod-bigram}(w_i|w_{i-1}) =
\frac{\vert \{ v | C(v w_{i-1} w_i) > 0 \} \vert - D_2}
{\sum_w \vert \{ v | C(v w_{i-1} w) > 0 \} \vert}
+ \lambda(w_{i-1}) \Px{ikn-mod-unigram}(w_i)
\end{array}
$$
$$
\begin{array}{@{}l}
\Px{ikn-mod-unigram}(w_i|w_{i-1}) =
\frac{\vert \{ v | v w_i) > 0 \} \vert - D_1}
{\sum_w \vert \{ v | C( v w) > 0 \} \vert}
+ \lambda \frac{1}{\vert V \vert}
\end{array}
$$
}

\begin{figure}
$$
\psfig{figure=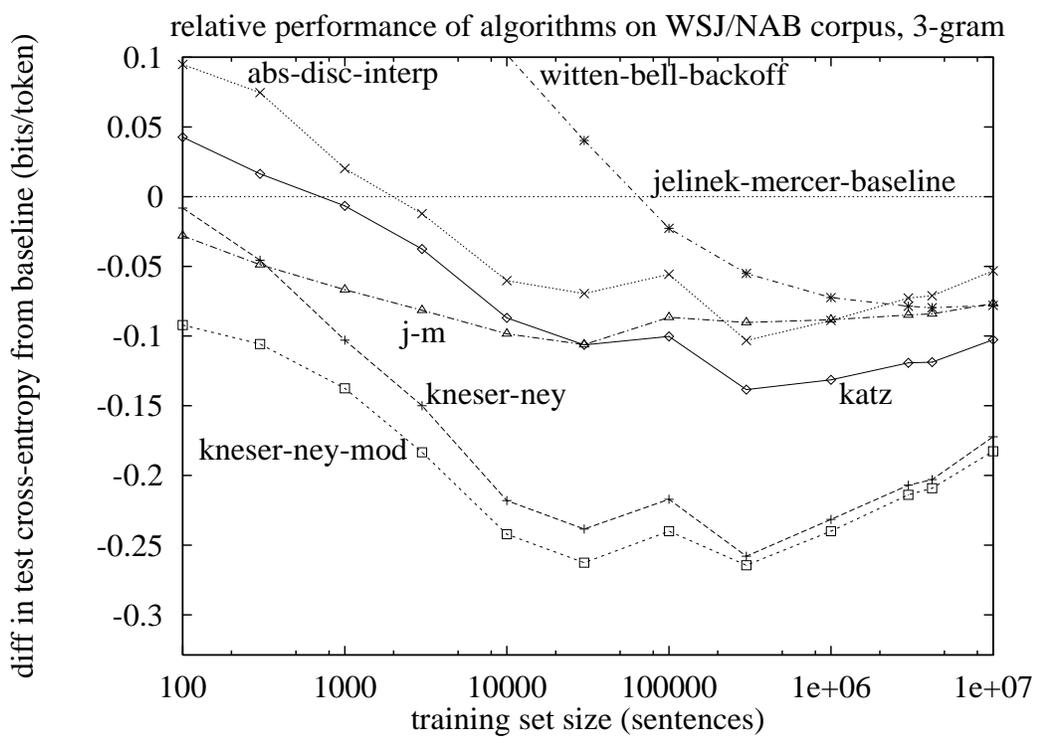,width=5.5in,angle=0}
$$
\caption{Smoothing results across data sizes}
\label{fig:smoothing}
\end{figure}

In Figure \ref{fig:smoothing}, we repeat results from
\namecite{Chen:99a}.  These are the only results in this paper not run
on exactly the same sections of the corpus for heldout, training, and
test as the rest of the paper, but we expect them to be very
comparable.  The baseline used for these experiments was a simple
version of Jelinek-Mercer smoothing, using a single bucket; that
version is identical to the first smoothing technique we described,
simple interpolation.  Kneser-Ney smoothing is the interpolated version
of Kneser-Ney smoothing used throughout this paper, and Kneser-Ney mod
is the version with three discounts instead of a single discount.  Katz
smoothing is essentially the same as the version in this paper.  j-m
is short for Jelinek Mercer smoothing, sometimes called deleted
interpolation elsewhere; abs-disc-interp is the interpolated version
of absolute discounting.  Training set size was measured in sentences,
rather than in words, with about 20 words per sentence.  Notice
that Jelinek-Mercer smoothing and Katz smoothing cross, one being
better at lower data sizes, the other at higher sizes.  This was part
of our motivation for running all experiments in this paper on
multiple data sizes.  On the other hand, in those experiments, which
were done on multiple corpora, we did not find any techniques where
one technique worked better on one corpus, and another worked better
on another one.  Thus, we feel reasonably confident in our decision
not to run on multiple corpora.  \namecite{Chen:99a} give a much more
complete comparison of these techniques, as well as much more in depth
analysis.  \namecite{Chen:98a} gives a superset that also serves as a
tutorial introduction.

\section{Higher-order n-grams}

\begin{figure}
$$
\psfig{figure=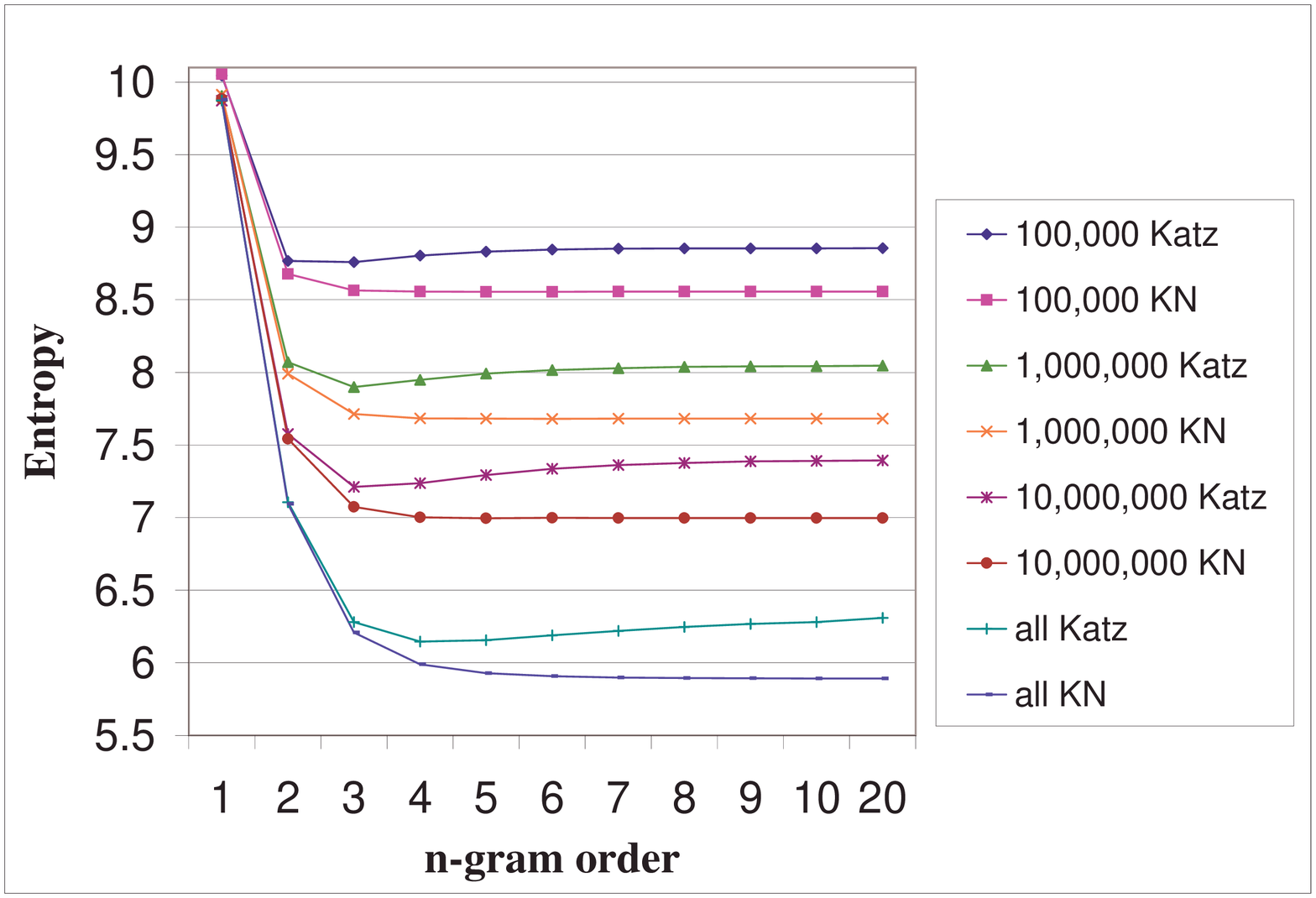,width=5.5in,angle=-90}
$$
\caption{N-gram Order vs. Entropy}
\label{fig:ngramorder}
\end{figure}

While the trigram assumption has proven, in practice, to be
reasonable, there are many cases in which even longer contexts can be
helpful.  It is thus natural to relax the trigram assumption, and,
rather than computing $P(w_i|w_{i-2}w_{i-1})$, use a longer context,
such as $P(w_i|w_{i-4}w_{i-3}w_{i-2}w_{i-1})$, a five-gram model.  In
many cases, no sequence of the form $w_{i-4}w_{i-3}w_{i-2}w_{i-1}$
will have been seen in the training data, and the system will need to
backoff to or interpolate with four-grams, trigrams, bigrams, or even
unigrams, but in those cases where such a long sequence has been seen,
it may be a good predictor of $w_i$.

Some earlier experiments with longer contexts showed little benefit
from them.  This turns out to be partially due to smoothing.  As shown
by \namecite{Chen:99a}, some smoothing methods work significantly
better with higher-order n-grams than others do.  In particular,
the advantage of Interpolated Kneser-Ney smoothing is much larger with
higher-order n-grams than with lower-order ones.

We performed a variety of experiments on the relationship between
n-gram order and perplexity.  In particular, we tried both Katz
smoothing and Interpolated Kneser-Ney smoothing on n-gram orders from
1 to 10, as well as 20, and over our standard data sizes.  The results
are shown in Figure \ref{fig:ngramorder}.

As can be seen, and has been previously observed \mycite{Chen:99a}, the
behavior for Katz smoothing is very different than the behavior for
Kneser-Ney smoothing.  Chen and Goodman determined that the main cause
of this difference was that backoff smoothing techniques, such as Katz
smoothing, or even the backoff version of Kneser-Ney smoothing (we use
only interpolated Kneser-Ney smoothing in this work), work poorly on low
counts, especially one counts, and that as the n-gram order increases,
the number of one counts increases.  In particular, Katz smoothing has
its best performance around the trigram level, and actually gets worse
as this level is exceeded.  Kneser-Ney smoothing on the other hand is
essentially monotonic even through 20-grams.  

The plateau point for Kneser-Ney smoothing depends on the amount of
training data available.  For small amounts, 100,000 words, the
plateau point is at the trigram level, whereas when using the full
training data, 280 million words, small improvements occur even into
the 6-gram (.02 bits better than 5-gram) and 7-gram (.01 bits better
than 6-gram.)  Differences of this size are interesting, but not of
practical importance.  The difference between 4-grams and 5-grams, .06
bits, is perhaps important, and so, for the rest of our experiments,
we often use models built on 5-gram data, which appears to give a good
tradeoff between computational resources and performance. 

Note that in practice, going beyond trigrams is often impractical.
The tradeoff between memory and performance typically requires heavy
pruning of 4-grams and 5-grams, reducing the potential improvement
from them.  Throughout this paper, we ignore memory-performance
tradeoffs, since this would overly complicate already difficult
comparisons.  We seek instead to build the single best system
possible, ignoring memory issues, and leaving the more practical, more
interesting, and very much more complicated issue of finding the best
system at a given memory size, for future research (and a bit of past
research, too \mycite{Goodman:00a}).  Note that many of the experiments
done in this section could not be done at all without the special tool
described briefly at the end of this paper, and in more detail in the
\longversion{appendix.}\shortversion{appendix of the extended version
of this paper.}

\section{Skipping} 

As one moves to larger and larger n-grams, there is less and less
chance of having seen the exact context before; but the chance of
having seen a similar context, one with most of the words in it,
increases.  Skipping models
\mycite{Rosenfeld:94a,Huang:93a,Ney:94a,Martin:99a,Siu:00a} make use of
this observation.  There are also variations on this technique, such
as techniques using lattices \mycite{Saul:97a,Dupont:97a}, or models
combining classes and words \mycite{Blasig:99a}.

{\sloppy When considering a five-gram context, there are many
subsets of the five-gram we could consider, such as
$P(w_i|w_{i-4}w_{i-3}w_{i-1})$ or $P(w_i|w_{i-4}w_{i-2}w_{i-1})$.
Perhaps we have never seen the phrase ``{\text Show John a good time}'' but we
have seen the phrase ``{\text Show Stan a good time.}''  A normal 5-gram
predicting $P(\mboxtext{time} | \mboxtext{show John a good})$ would
back off to $P(\mboxtext{time} | \mboxtext{John a good})$ and from
there to $P(\mboxtext{time} | \mboxtext{a good})$, which would have a
relatively low probability.  On the other hand, a skipping model of
the form $P(w_i|w_{i-4}w_{i-2}w_{i-1})$ would assign high probability
to $P(\mboxtext{time} | \mboxtext{show \rule{2em}{.1em} a good})$.

}

These skipping 5-grams are then interpolated with a normal 5-gram,
forming models such as 
$$
\lambda P(w_i|w_{i\!-\!4}w_{i\!-\!3}w_{i\!-\!2}w_{i\!-\!1}) + \mu P(w_i|w_{i\!-\!4}w_{i\!-\!3}w_{i\!-\!1}) + (1\!-\!\lambda\!-\!\mu)P(w_i|w_{i\!-\!4}w_{i\!-\!2}w_{i\!-\!1})
$$
where, as usual, $0 \leq \lambda \leq 1$ and $0 \leq \mu \leq 1$
and $0 \leq (1-\lambda-\mu) \leq 1$.

Another (and more traditional) use for skipping is as a sort of poor man's higher order
n-gram.  One can, for instance, create a model of the form 
$$
\lambda P(w_i|w_{i-2}w_{i-1}) + \mu P(w_i|w_{i-3}w_{i-1}) + (1-\lambda-\mu)P(w_i|w_{i-3}w_{i-2})
$$
In a model of this form, no component probability depends on more than
two previous words, like a trigram, but the overall probability is
4-gram-like, since it depends on $w_{i-3}$, $w_{i-2}$, and $w_{i-1}$.
We can extend this idea even further, combining in all pairs of
contexts in a 5-gram-like, 6-gram-like, or even 7-gram-like way, with
each component probability never depending on more than the previous
two words.

\begin{figure}
$$\psfig{figure=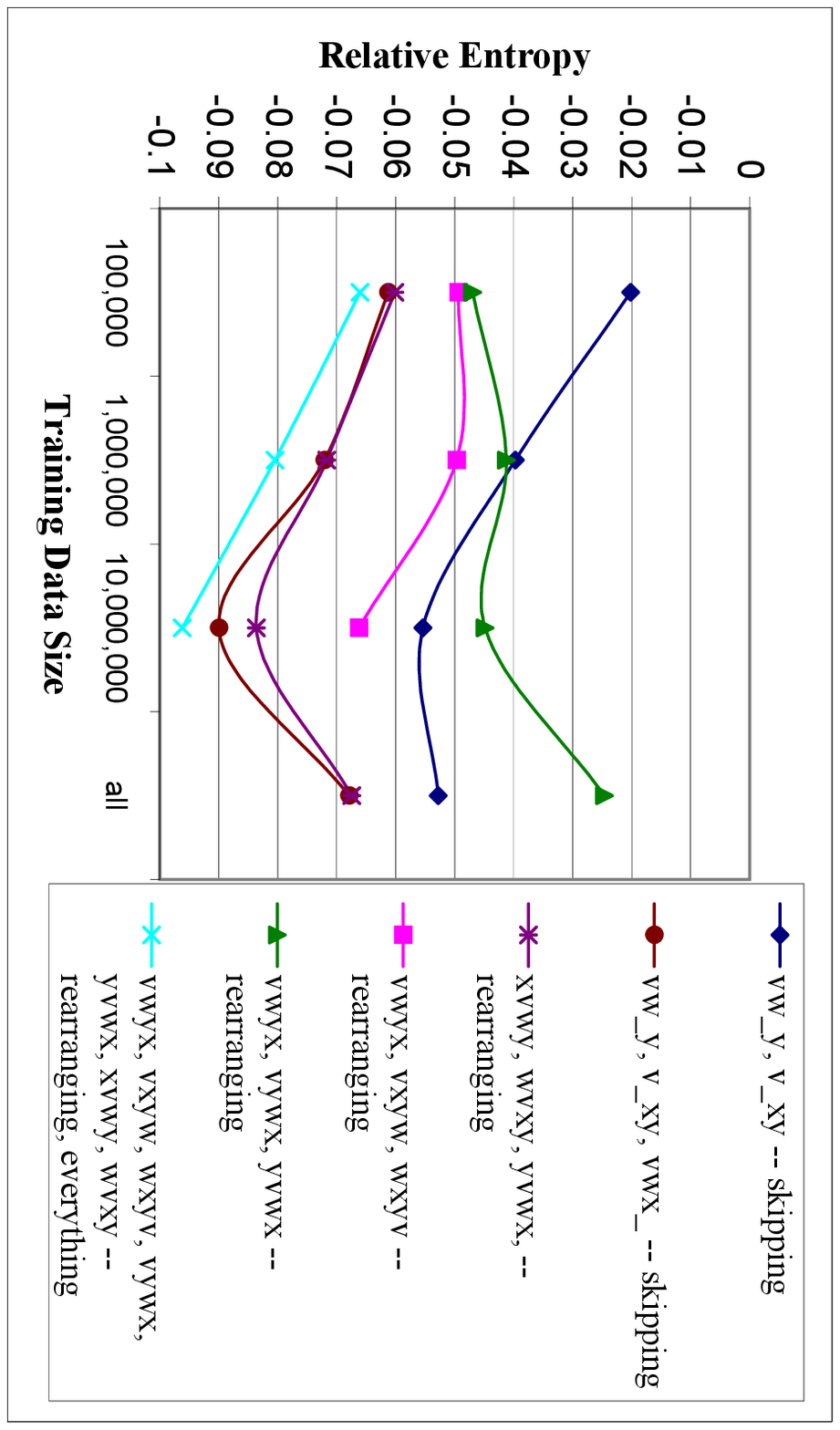,width=5.5in,angle=0}$$
\caption{5-gram Skipping Techniques versus 5-gram Baseline}
\label{fig:fivegramskpping}
\end{figure}

We performed two sets of experiments, one on 5-grams and one on
trigrams.  For the 5-gram skipping experiments, all contexts depended on at most
the previous four words, $w_{i-4}$, $w_{i-3}$, $w_{i-2}$, $w_{i-1}$,
but used the four words in a variety of ways.  We tried six models,
all of which were interpolated with a baseline 5-gram model.  For
readability and conciseness, we
define a new notation, letting $v=w_{i-4}$, $w=w_{i-3}$, $x=w_{i-2}$
and $y=w_{i-1}$, allowing us to avoid numerous subscripts in what
follows.  The results are shown in Figure \ref{fig:fivegramskpping}.

The first model interpolated dependencies on $vw \un y$ and $v \un
xy$.  This simple model does not work well on the smallest training
data sizes, but is competitive for larger ones.  Next, we tried a
simple variation on this model, which also interpolated in $vwx \un$.
Making that simple addition leads to a good-sized improvement at all
levels, roughly .02 to .04 bits over the simpler skipping model.  Our
next variation was analogous, but adding back in the dependencies on
the missing words.  In particular, we interpolated together $xvwy$, $wvxy$,
and $yvwx$; that is, all models depended on the same variables, but
with the interpolation order modified.  For instance, by $xvwy$, we
refer to a model of the form $P(z|vwxy)$ interpolated with $P(z|vw\un
y$) interpolated with $P(z|w\un y$) interpolated with $P(z|y$)
interpolated with $P(z|y$) interpolated with $P(z)$.  All of these
experiments were done with Interpolated Kneser-Ney smoothing, so all
but the first probability uses the modified backoff distribution.  This model
is just like the previous one, but for each component starts the
interpolation with the full 5-gram.  We had hoped that in the case
where the full 5-gram had occurred in the training data, this would
make the skipping model more accurate, but it did not help at
all.\footnote{In fact, it hurt a tiny bit, 0.005 bits at the
10,000,000 word training level.  This turned out to be due to
technical smoothing issues.  \longversion{
In particular, after
some experimentation, this turned out to be due to our use of
Interpolated Kneser-Ney smoothing with a single discount, even though
we know that using multiple discounts is better.  When using multiple
discounts, the problem goes away.  }}

We also wanted to try more radical approaches.  For instance, we tried
interpolating together $vwyx$ with $vxyw$ and $wxyv$ (along with the
baseline $vwxy$).  This model puts each of the four preceding words in
the last (most important) position for one component.  This model does
not work as well as the previous two, leading us to conclude that the
$y$ word is by far the most important.  We also tried a model with
$vwyx$, $vywx$, $yvwx$, which puts the $y$ word in each possible
position in the backoff model.  This was overall the worst model,
reconfirming the intuition that the $y$ word is critical.  However, as
we saw by adding $vwx \un$ to $vw \un y$ and $v \un xy$, having a
component with the $x$ position final is also important.  This will
also be the case for trigrams.

Finally, we wanted to get a sort of upper bound on how well 5-gram
models could work.  For this, we interpolated together $vwyx$, $vxyw$,
$wxyv$, $vywx$, $yvwx$, $xvwy$ and $wvxy$.  This model was chosen as
one that would include as many pairs and triples of combinations of
words as possible.  The result is a marginal gain -- less than 0.01
bits -- over the best previous model.  

We do not find these results particularly encouraging.  In particular,
when compared to the sentence mixture results that will be presented
later, there seems to be less potential to be gained from skipping
models.  Also, while sentence mixture models appear to lead to larger
gains the more data that is used, skipping models appear to get their
maximal gain around 10,000,000 words.  Presumably, at the largest data
sizes, the 5-gram model is becoming well trained, and there are fewer
instances where a skipping model is useful but the 5-gram is not.

\begin{figure}
$$\psfig{figure=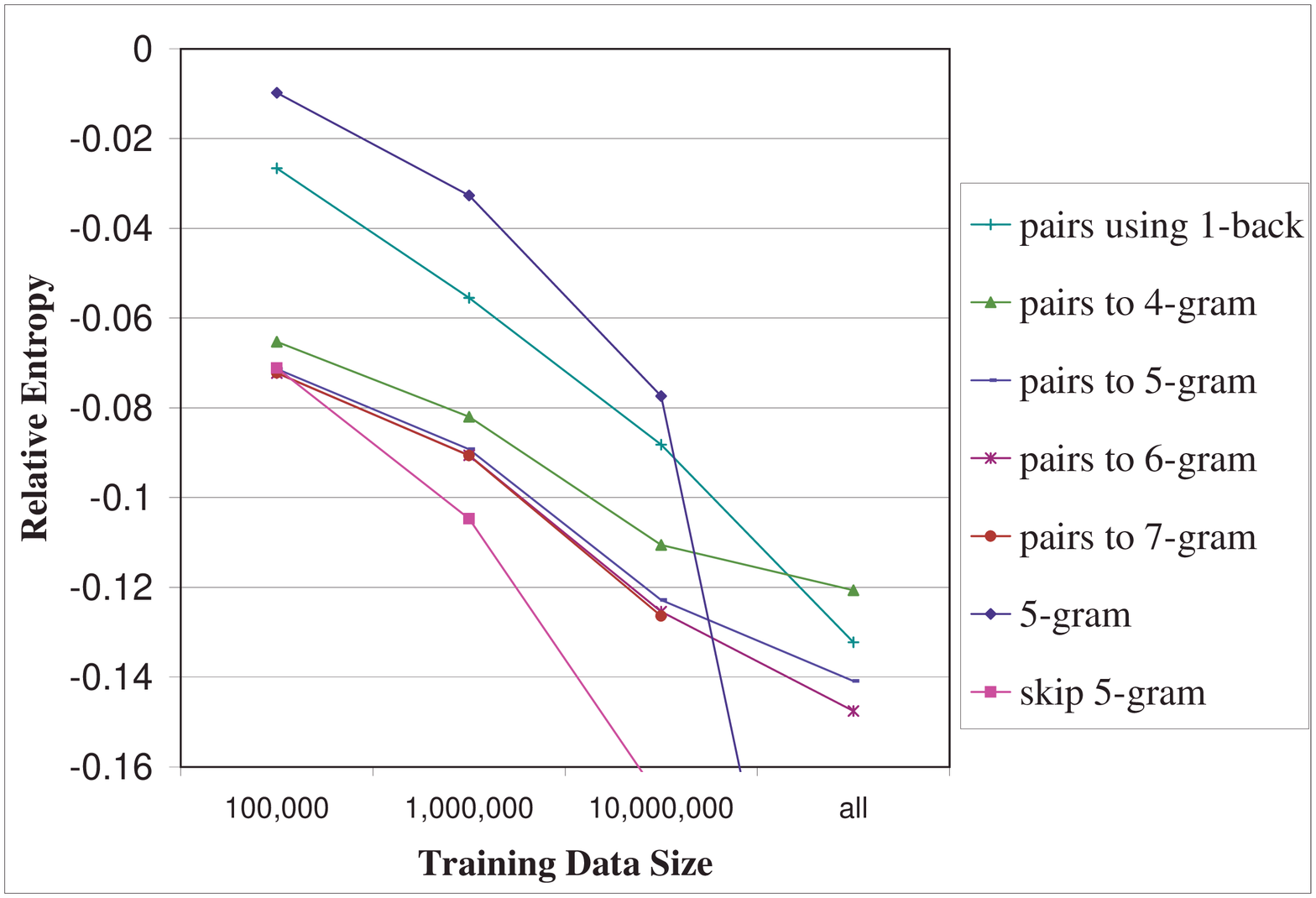,width=5.5in,angle=-90}$$
\caption{Trigram Skipping Techniques versus Trigram Baseline}
\label{fig:trigramskipping}
\end{figure}

We also examined trigram-like models.  These results are shown in
Figure \ref{fig:trigramskipping}.  The baseline for comparison was a
trigram model.  For comparison, we also show the relative improvement
of a 5-gram model over the trigram, and the relative improvement of
the skipping 5-gram with $vw \un y$, $v \un xy$ and $vwx \un$.  For
the trigram skipping models, each component never depended on more
than two of the previous words.  We tried 5 experiments of this form.
First, based on the intuition that pairs using the 1-back word ($y$) are
most useful, we interpolated $xy$, $wy$, $vy$, $uy$ and $ty$ models.
This did not work particularly well, except at the largest sizes.
Presumably at those sizes, a few appropriate instances of the 1-back
word had always been seen.  Next, we tried using all pairs of words
through the 4-gram level: $xy$, $wy$ and $wx$.  Considering its
simplicity, this worked very well.  We tried similar models using all
5-gram pairs, all 6-gram pairs and all 7-gram pairs; this last model
contained 15 different pairs.  However, the improvement over 4-gram
pairs was still marginal, especially considering the large number of
increased parameters.

The trigram skipping results are, relative to their baseline, much
better than the 5-gram skipping results.  They do not appear to have
plateaued when more data is used and they are much more comparable to
sentence mixture models in terms of the improvement they get.
Furthermore, they lead to more improvement than a 5-gram alone does
when used on small amounts of data (although, of course, the best
5-gram skipping model is always better than the best trigram skipping
model.)  This makes them a reasonable technique to use with small and
intermediate amounts of training data, especially if 5-grams cannot be
used.

\section{Clustering} 

\subsection{Using Clusters}

Next, we describe our clustering techniques, which are a bit different
(and, as we will show, slightly more effective) than traditional
clustering \mycite{Brown:90a,Ney:94a}.  Consider a probability such as
$P(\mboxtext{Tuesday} | \mboxtext{party on})$.  Perhaps the training
data contains no instances of the phrase {\text ``party on Tuesday''},
although other phrases such as {\text ``party on Wednesday''} and {\text
``party on Friday''} do appear.  We can put words into classes,
such as the word {\text ``Tuesday''} into the class {\text WEEKDAY}.  Now,
we can consider the probability of the word {\text ``Tuesday''} given
the phrase {\text ``party on''}, and also given that the next word is
a {\text WEEKDAY}.  We will denote this probability by
$P(\mboxtext{Tuesday} | \mboxtext{party on WEEKDAY})$.  We can then decompose the probability
\begin{eqnarray*}
\lefteqn{P(\mboxtext{Tuesday} | \mboxtext{party on}) } \nonumber \\
& = & P(\mboxtext{WEEKDAY} | \mboxtext{party on}) \times 
P(\mboxtext{Tuesday} | \mboxtext{party on WEEKDAY}) 
\end{eqnarray*}
When each word belongs to only one class, which is called hard clustering, this decompostion is a
strict 
\longversion{equality a fact that can be trivially proven. Let $W_i$
represent the cluster of word $w_i$.  Then,
\begin{eqnarray}
P(W_i | w_{i-2} w_{i-1}) \times P(w_i | w_{i-2} w_{i-1} W_i) & = &
\frac{P(w_{i-2} w_{i-1} W_i)}{P(w_{i-2} w_{i-1})} \times \frac{P(w_{i-2} w_{i-1}
W_i w_i)}{P(w_{i-2} w_{i-1} W_i)} \nonumber \\ 
& = & 
\frac{P(w_{i-2} w_{i-1} W_i w_i)}{P(w_{i-2} w_{i-1})} \label{eqn:bigcluster}
\end{eqnarray}
Now, since each word belongs to a single cluster, $P(W_i | w_i)$ =
1, and thus
\begin{eqnarray}
P(w_{i-2} w_{i-1} W_i w_i) &=& P(w_{i-2} w_{i-1} w_i) \times P(W_i | w_{i-2} w_{i-1} w_i) \nonumber \\
& = & P(w_{i-2} w_{i-1} w_i) \times P(W_i | w_i) \nonumber \\
& = & P(w_{i-2} w_{i-1} w_i)
\label{eqn:smallcluster}
\end{eqnarray}
Substituting Equation \ref{eqn:smallcluster} into Equation
\ref{eqn:bigcluster}, we get
\begin{equation}
P(W_i | w_{i-2} w_{i-1}) \times P(w_i | w_{i-2} w_{i-1} W_i) =
\frac{P(w_{i-2} w_{i-1} w_i)}{P(w_{i-2} w_{i-1})} =
P(w_i | w_{i-2} w_{i-1})
\label{eqn:predictequation}
\end{equation}
}
\shortversion{equality.
Notice that, because we are using hard clusters, if we know $w_i$, we
also know $W_i$, meaning that $P(w_{i-2} w_{i-1} W_i w_i) = P(w_{i-2} w_{i-1} w_i)$.
With this fact,
\begin{eqnarray}
P(W_i | w_{i\!-\!2} w_{i\!-\!1})\! \times \! P(w_i | w_{i\!-\!2} w_{i\!-\!1} W_i) \! & = & 
\frac{P(w_{i-2} w_{i-1} W_i)}{P(w_{i-2} w_{i-1})} \times \frac {P(w_{i-2} w_{i-1} W_i w_i)}{P(w_{i-2} w_{i-1} W_i)} \nonumber \\
& = & \frac{P(w_{i-2} w_{i-1} W_i w_i)}{P(w_{i-2} w_{i-1})} \nonumber \\
& = & \frac{P(w_{i-2} w_{i-1} w_i)}{P(w_{i-2} w_{i-1})} \nonumber \\
& = & P(w_i | w_{i-2} w_{i-1})
\label{eqn:predictequation}
\end{eqnarray}
The extended version of the paper gives a slightly more detailed derivation.}

Now, although Equation \ref{eqn:predictequation} is a strict equality,
when smoothing is taken into consideration, using the clustered
probability will be more accurate than the non-clustered probability.
For instance, even if we have never seen an example of {\text ``party
on Tuesday''}, perhaps we have seen examples of other phrases, such
as {\text ``party on Wednesday''} and thus, the probability
$P(\mboxtext{WEEKDAY} | \mboxtext{party on})$ will be relatively
high.  And although we may never have seen an example of {\text ``party
on WEEKDAY Tuesday''}, after we backoff or interpolate with a lower
order model, we may be able to accurately estimate $P(\mboxtext{Tuesday}
| \mboxtext{on WEEKDAY})$.  Thus, our smoothed clustered estimate may be a
good one.  We call this particular kind of clustering {\em predictive
clustering}.  (On the other hand, we will show that if the clusters
are poor, predictive clustering can also lead to degradation.)

Note that predictive clustering has other uses as well as for
improving perplexity.  Predictive clustering can be used to
significantly speed up maximum entropy training \mycite{Goodman:01a}, by
up to a factor of 35, as well as to compress language models
\mycite{Goodman:00a}.

Another type of clustering we can do is to cluster the words in the
contexts.  For instance, if {\text ``party''} is in the class  {\text
EVENT} and {\text ``on''} is in the class {\text PREPOSITION},
then we could write
$$
P(\mboxtext{Tuesday} | \mboxtext{party on}) \approx
P(\mboxtext{Tuesday} | \mboxtext{EVENT PREPOSITION}) 
$$
or more generally
\begin{equation}
P(w|w_{i-2}w_{i-1}) \approx P(w|W_{i-2}W_{i-1})
\label{eqn:IBMish}
\end{equation}
Combining Equation \ref{eqn:IBMish} with Equation \ref{eqn:predictequation} we get
\begin{equation}
P(w|w_{i-2}w_{i-1}) \approx P(W|W_{i-2}W_{i-1}) \times P(w|W_{i-2}W_{i-1}W)
\label{eqn:fullIBM}
\end{equation}
Since Equation \ref{eqn:fullIBM} does not take into account the exact
values of the previous words, we always (in this work) interpolate it
with a normal trigram model.  We call the interpolation of Equation
\ref{eqn:fullIBM} with a trigram {\em fullibm} clustering.  We call it fullibm
because it is a generalization of a technique invented at IBM
\mycite{Brown:90a}, which uses the approximation $P(w|W_{i-2}W_{i-1}W)
\approx P(w|W)$ to get 
\begin{equation}
P(w|w_{i-2}w_{i-1}) \approx P(W|W_{i-2}W_{i-1}) \times P(w|W)
\label{eqn:IBM}
\end{equation}
which, when interpolated with a normal trigram, we refer to as {\em ibm}
clustering.  Given that fullibm clustering uses more information than
regular ibm clustering, we assumed that it would lead to
improvements.\longversion{\footnote{In fact, we originally used the name goodibm
instead of fullibm on the assumption that it {\it must} be better.}}
As will be shown, it works about the same, at least when interpolated
with a normal trigram model.

Alternatively, rather than always discarding information, we could
simply change the backoff order, called {\em index} clustering:
\begin{equation}
\Px{index}(\mboxtext{Tuesday} | \mboxtext{party on}) =  
P(\mboxtext{Tuesday} | \mboxtext{party EVENT on PREPOSITION}) 
\label{eqn:WEEKDAYcondition}
\end{equation}
Here, we abuse notation slightly to use the order of the words on the
right side of the $|$ to indicate the backoff/interpolation order.
Thus, Equation \ref{eqn:WEEKDAYcondition} implies that we would go from
$P(\mboxtext{Tuesday} | \mboxtext{party EVENT on PREPOSITION})$ to
$P(\mboxtext{Tuesday} | \mboxtext{EVENT on PREPOSITION})$ to
$P(\mboxtext{Tuesday} | \mboxtext{on PREPOSITION})$ to
$P(\mboxtext{Tuesday} | \mboxtext{PREPOSITION})$ to $P(\mboxtext{Tuesday})$.
Notice that since each word belongs to a single cluster, some of these
variables are redundant.  For instance, in our notation
$$
C(\mboxtext{party EVENT on PREPOSITION}) =
C(\mboxtext{party on})
$$ and
$$
C(\mboxtext{EVENT on PREPOSITION}) =
C(\mboxtext{EVENT on})
$$
We generally write an index clustered model as $P(w_i |
w_{i-2} W_{i-2} w_{i-1} W_{i-1})$.

There is one especially noteworthy technique, {\em fullibmpredict}.  This is the
best performing technique we have found (other than combination
techniques.)  This technique makes use of the intuition behind predictive
clustering, factoring the problem into prediction of the cluster,
followed by prediction of the word given the cluster.  In addition, at
each level, it smooths this prediction by combining a word-based and a
cluster-based estimate.  It is not interpolated with a normal trigram model.  It is of the form
$$
\Px{fullibmpredict}(w|w_{i-2}w_{i-1}) = \! \begin{array}[t]{l}
(\lambda P(W|w_{i-2}w_{i-1}) + (1\!-\!\lambda) P(W|W_{i-2}W_{i-1})) \times \\
(\mu P(w|w_{i-2}w_{i-1}W) + (1\!-\!\mu) P(w|W_{i-2}W_{i-1}W)
\end{array}
$$

There are many variations on these themes.  As it happens, none of the
others works much better than ibm clustering, so we describe them only
very briefly.  One is {\em indexpredict}, combining index and predictive
clustering:
\begin{eqnarray*}
\lefteqn{\Px{indexpredict}(w_i|w_{i-2}w_{i-1}) = } \\
&&P(W_i | w_{i-2} W_{i-2} w_{i-1} W_{i-1}) \times P(w_i | w_{i-2} W_{i-2} w_{i-1} W_{i-1} W_i)
\end{eqnarray*}
Another is {\em combinepredict}, interpolating a normal trigram with a predictive clustered trigram:
\begin{eqnarray*}
\lefteqn{\Px{combinepredict}(w_i|w_{i-2}w_{i-1}) =}\\
&& \lambda P(w_i|w_{i-1}w_{i-2}) + (1-\lambda)P(W_i | w_{i-2} w_{i-1}) \times P(w_i | w_{i-2} w_{i-1} W_i)
\end{eqnarray*}
Finally, we wanted to get some sort of upper bound on how much could
be gained by clustering, so we tried combining all these clustering
techniques together, to get what we call {\em allcombinenotop}, which
is an interpolation of a normal trigram, a fullibm-like model, an
index model, a predictive model, a true fullibm model, and an
indexpredict model.  \shortversion{A variation,
{\em allcombine}, interpolates the predict-type models first at the
cluster level, before interpolating with the word level models.  Exact
formulae are given in the extended version of the paper.
}
\longversionboring{
\begin{eqnarray*}
\lefteqn{\Px{allcombinenotop}(w_i|w_{i-2}w_{i-1}) =} \\
&& \lambda P(w_i|w_{i-2}w_{i-1})  \\
&& + \mu P(w_i | W_{i-2}W_{i-1})  \\
&& + \nu P(w_i | w_{i-2} W_{i-2} w_{i-1} W_{i-1}) \\
&& + \alpha P(W_i | w_{i-2} w_{i-1}) \times P(w_i | w_{i-2} w_{i-1} W_i) \\
&& + \beta P(W_i | W_{i-2}W_{i-1}) \times P(w_i | W_{i-1}W_{i-2}W_i) \\
&& + (1-\lambda - \nu - \alpha - \beta) P(W_i | w_{i-2} W_{i-2} w_{i-1} W_{i-1}) \times P(w_i | w_{i-2} W_{i-2} w_{i-1} W_{i-1} W_i)
\end{eqnarray*}
or, for a different way of combining, {\em allcombine}, which
interpolates the predict-type models first at the cluster level,
before inteprolating with the word level models.
\begin{eqnarray*}
\lefteqn{\Px{allcombine}(w_i|w_{i-2}w_{i-1}) =} \\
&& \lambda P(w_i|w_{i-2}w_{i-1})  \\
&& + \mu P(w_i | W_{i-2}W_{i-1})  \\
&& + \nu P(w_i | w_{i-2} W_{i-2} w_{i-1} W_{i-1}) \\
&& +  \begin{array}[t]{l} (1\!-\!\lambda\!-\!\mu\!-\!\nu)
 \\
   \left[ \times \alpha P(W_i | w_{i-2} w_{i-1}) + \beta P(W_i | W_{i-2}W_{i-1}) + (1\!-\! \alpha \!-\! \beta) P(W_i | w_{i-2} W_{i-2} w_{i-1} W_{i-1}) 
\right] \\
\times 
\left[
\gamma P(w_i | w_{i-2} w_{i-1} W_i) \!+\! \rho  P(w_i | W_{i-2}W_{i-1}W_i) \!+\!   (1\!-\!\gamma \!-\! \rho) P(w_i | w_{i-2} W_{i-2} w_{i-1} W_{i-1} W_i)
\right] 
\end{array}
\end{eqnarray*}
}

\longversion{
\begin{figure}
$$\psfig{figure=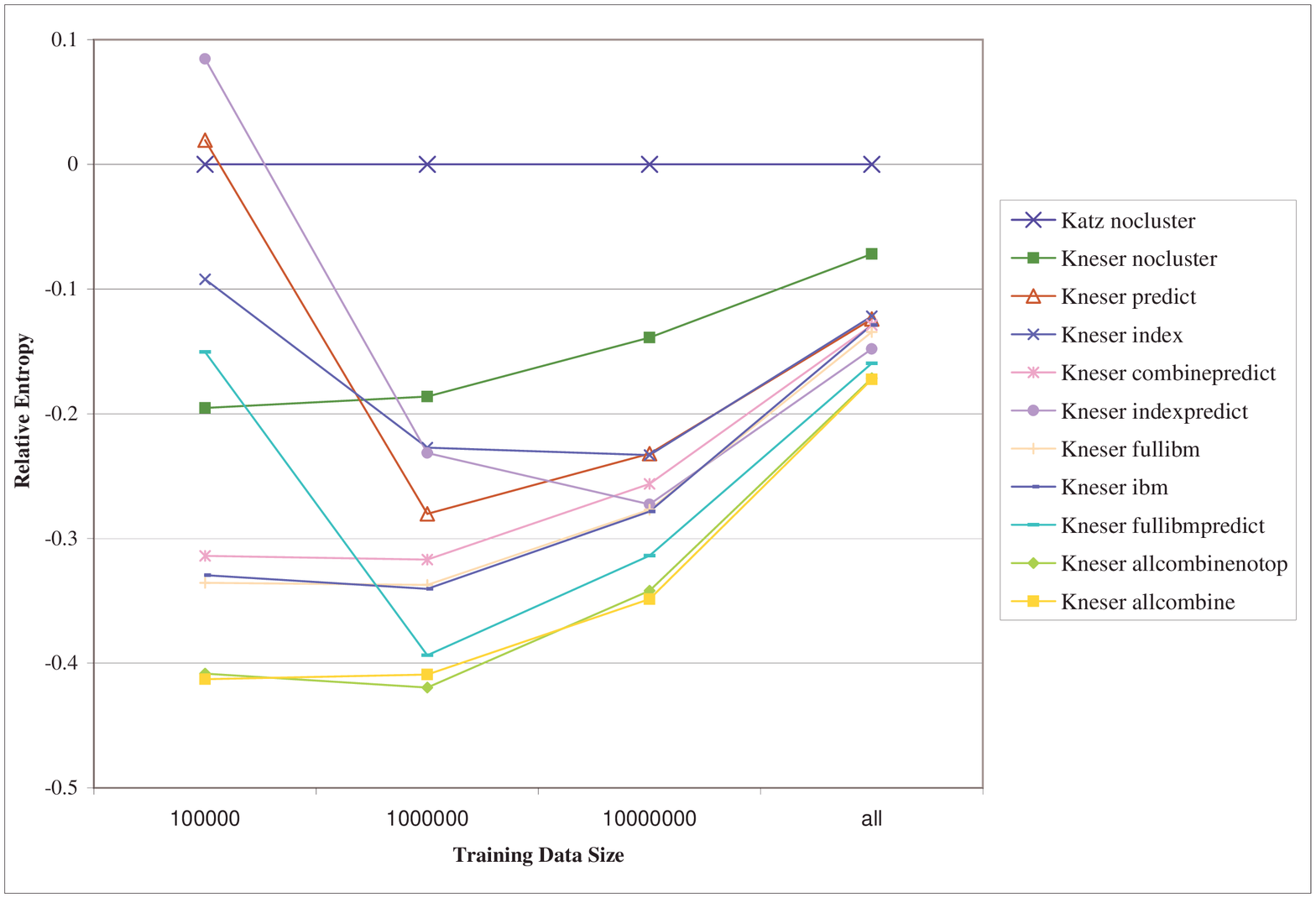,width=5.5in,angle=-90}$$
\caption{Comparison of nine different clustering techniques, Kneser-Ney smoothing}
\label{fig:clusterkneserall}
\end{figure}

\begin{figure}
$$\psfig{figure=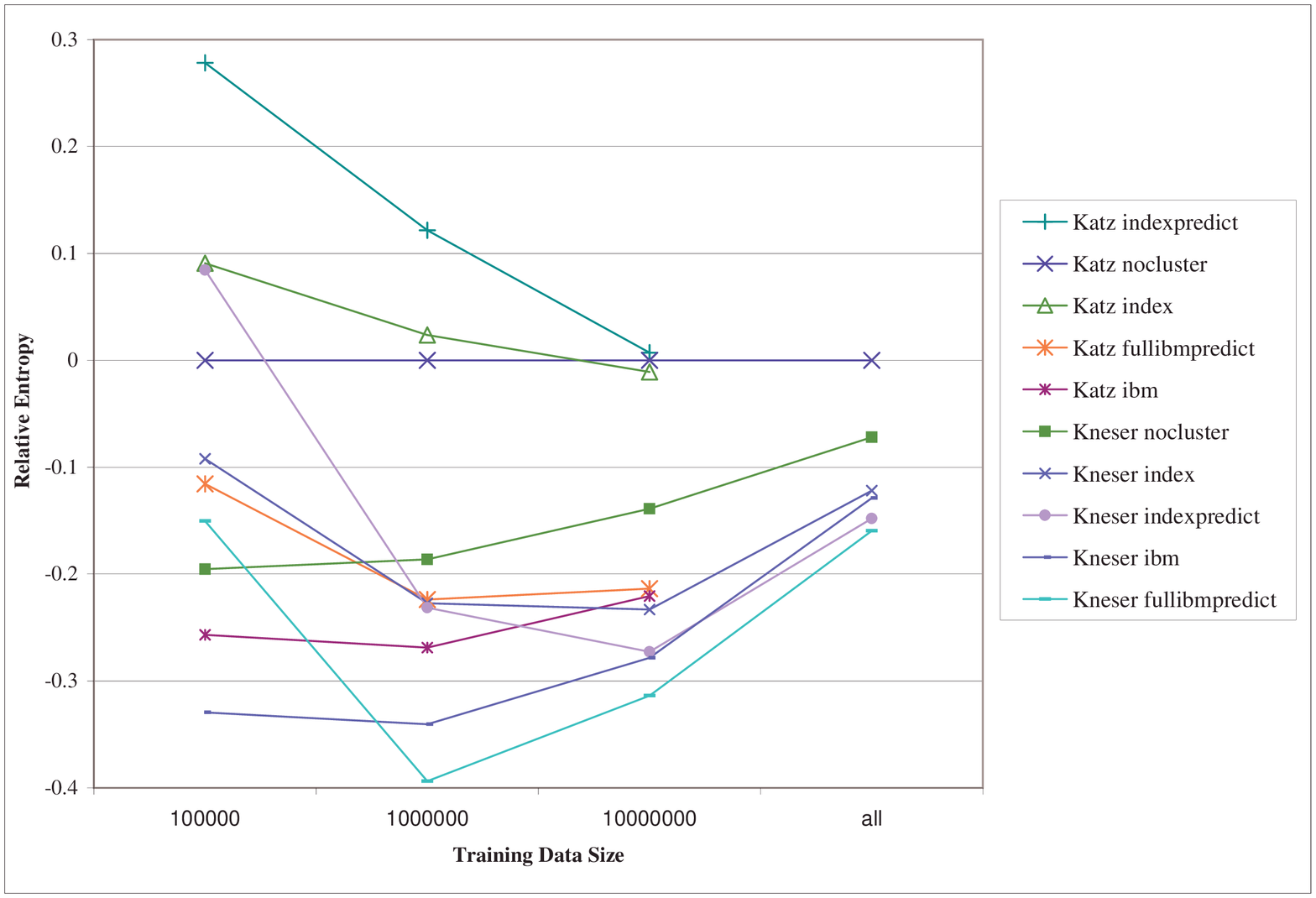,width=5.5in,angle=-90}$$
\caption{Comparison of Kneser-Ney smoothed Clustering to Katz smoothed}
\label{fig:clusterkatzkneser}
\end{figure}
}
\shortversion{
\begin{figure}
$$\psfig{figure=clusterchartsentropykneseronly.eps,width=5.5in,angle=-90}$$
\caption{Comparison of nine different clustering techniques, Kneser-Ney smoothing}
\label{fig:clusterkneserall}
\end{figure}
}
In Figure \ref{fig:clusterkneserall}, we show a comparison of nine
different clustering techniques, all using Kneser-Ney smoothing.  The
clusters were built separately for each training size.  \longversion{To keep these
comparable to our next chart, we use Katz smoothing as the baseline
for the relative entropy comparisons. } Notice that the value of
clustering decreases with training data; at small data sizes, it is
about 0.2 bits for the best clustered model; at the largest sizes, it
is only about 0.1 bits.  Since clustering is a technique for dealing
with data sparseness, this is unsurprising.  Next, notice that ibm
clustering consistently works very well.  Of all the other techniques
we tried, only 4 others worked as well or better: fullibm clustering,
which is a simple variation; allcombine and allcombinenotop, which
interpolate in a fullibm; and fullibmpredict.  Fullibmpredict works
very well -- as much as 0.05 bits better than ibm clustering.  However,
it has a problem at the smallest training size, in which case it is
worse.  We believe that the clusters at the smallest training size are
very poor, and that predictive style clustering gets into trouble when
this happens, since it smooths across words that may be unrelated,
while ibm clustering interpolates in a normal trigram model, making it
more robust to poor clusters.  All of the models that use predict
clustering and do not interpolate an unclustered trigram are actually
worse than the baseline at the smallest training size.

\longversion{Note that our particular experiments, which use a fixed
vocabulary, are a severe test of clustering at the smallest levels.
Many of the words in the 58,000 word vocabulary do not occur at all in
100,000 words of training data.  We attempted to partially deal with
this by adding a ``pseudo-count'' to every word, a co-occurrence with
a fake word.  This would have the property of making all unseen words,
and words with very low counts, similar, hopefully putting them in the
same cluster.  The data with 100,000 words of training should be
interpreted more as how well a system will work with bad clusters, and
less about how a realistic system, in which the vocabulary would match
the training data, would work.}

\shortversion{In the extended version of this paper, we also show
results compared to a Katz smoothed model.}\longversion{In Figure
\ref{fig:clusterkatzkneser} we show a comparison of several techniques
using Katz smoothing and the same techniques with Kneser-Ney
smoothing.}  The results are similar, with some interesting
exceptions: in particular, indexpredict works well for the Kneser-Ney
smoothed model, but very poorly for the Katz smoothed model.  This
shows that smoothing can have a significant effect on other
techniques, such as clustering.  The other result is that across all
nine clustering techniques, at every size, the Kneser-Ney version
always outperforms the Katz smoothed version.  In fact, the Kneser-Ney
smoothed version also outperformed both interpolated and backoff
absolute discounting versions of each technique at every size.

There are two other ways to perform clustering, which 
we will not explore here.  First, one can cluster groups of
words -- complete contexts -- instead of individual words.  That is,
to change notation for a moment, instead of computing
$$
P(w | \mbox{\it word-cluster}(w_{i-2}) \mbox{\it word-cluster}(w_{i-1}))
$$
one could compute
$$
P(w | \mbox{\it context-cluster}(w_{i-2} w_{i-1}))
$$
For
instance, in a trigram model, one could cluster contexts like {\text
``New York''} and {\text ``Los Angeles''} as {\text ``CITY''}, and
{\text ``on Wednesday''} and {\text ``late tomorrow''} as {\text
``TIME''}.  There are many difficult issues to solve for this kind of
clustering.  Another kind of conditional
clustering one could do is to empirically determine, for a given
context, the best combination of clusters and words to use, the {\em
varigram} approach \mycite{Blasig:99a}.

\subsection{Finding Clusters}

A large amount of previous research has focused on how best to find
the clusters
\mycite{Brown:90a,Kneser:93a,Yamamoto:99a,Ueberla:95a,Pereira:93a,Bellegarda:96a}.
Most previous research has found only small differences between
different techniques for finding clusters.  One result however is that
automatically derived clusters outperform part-of-speech tags
\mycite{Niesler:98a}, at least when there is enough training data
\mycite{Ney:94a}.  We did not explore different techniques for finding
clusters, but simply picked one we thought would be good, based on
previous research.

There is no need for the clusters used for different positions to be
the same.  In particular, for a model like ibm clustering, with $P(w_i
| W_i) \times P(W_i|W_{i-2}W_{i-1})$, we will call the $W_i$ cluster a
{\em predictive} cluster, and the clusters for $W_{i-1}$ and $W_{i-2}$
{\em conditional} clusters.  The predictive and conditional clusters
can be different \mycite{Yamamoto:99a}.  For instance, consider a pair
of words like {\text a} and {\text an}.  In general, {\text a} and
{\text an} can follow the same words, and so, for predictive
clustering, belong in the same cluster.  But, there are very few words
that can follow both {\text a} and {\text an} -- so for conditional
clustering, they belong in different clusters.  We have also found in
pilot experiments that the optimal number of clusters used for
predictive and conditional clustering are different; in this paper, we
always optimize both the number of conditional and predictive clusters
separately, and reoptimize for each technique at each training data
size.  This is a particularly time consuming experiment, since each
time the number of clusters is changed, the models must be rebuilt
from scratch.  We always try numbers of clusters that are powers of 2,
e.g. 1, 2, 4, etc, since this allows us to try a wide range of numbers
of clusters, while never being more than a factor of 2 away from the
optimal number.  Examining charts of performance on heldout data, this
seems to produce numbers of clusters that are close enough to optimal.

The clusters are found automatically using a tool that attempts to
minimize perplexity.  In particular, for the conditional clusters we
try to minimize the perplexity of training data for a bigram of the
form $P(w_i | W_{i-1})$, which is equivalent to maximizing 
$$
\prod_{i=1}^N P(w_i | W_{i-1})
$$
For the predictive clusters, we try to minimize the perplexity of
training data of $P(W_i | w_{i-1}) \times P(w_i | W_i)$.
\shortversion{In the full version of this paper, we show that this is
equivalent to maximizing the perplexity of $P(w_{i-1} | W_i)$\footnote{As suggested to us by Lillian Lee}, which
is very convenient, since it means we can use the same tool to get
both conditional and predictive clusters, simply switching the order
of the input pairs.}
\longversionboring{
(We do not
minimize $P(W_i | w_{i-1}) \times P(w_i | w_{i-1} W_i)$, because we
are doing our minimization on unsmoothed training data, and the latter
formula would thus be equal to $P(w_i | w_{i-1})$ for any clustering.
If we were to use the method of leaving-one-out \mycite{Kneser:93a}, then we
could use the latter formula, but that approach is more difficult.)
Now,
\begin{eqnarray*}
\prod_{i=1}^N P(W_i | w_{i-1}) \times P(w_i | W_i) & = &
\prod_{i=1}^N \frac{P(w_{i-1} W_i )}{P(w_{i-1})} \times \frac{P(W_i
w_i)}{P(W_i)} \\
& = & 
\prod_{i=1}^N \frac{P(W_i w_i)}{P(w_{i-1})} \times \frac{P(w_{i-1} W_i
)}{P(W_i)}  \\
&=&
\prod_{i=1}^N \frac{P(w_i)}{P(w_{i-1})} \times P(w_{i-1}|W_i)
\end{eqnarray*}
Now, $\frac{P(w_i)}{P(w_{i-1})}$ will be independent of the clustering
used; therefore, it is sufficient to try to maximize $\prod_{i=1}^N
P(w_{i-1}|W_i)$.\footnote{Thanks to Lillian Lee}  This is very convenient, since it is exactly the
opposite of what was done for conditional clustering.  It means that
we can use the same clustering tool for both, and simply switch the
order used by the program used to get the raw counts for clustering.
}
We give more details about the clustering algorithm used in section
\ref{sec:implementationnotes}.

\section{Caching} 

If a speaker uses a word, it is likely that he will use the same word
again in the near future.  This observation is the basis of caching
\mycite{Kuhn:88a,Kuhn:90a,Kuhn:92a,Kupiec:89a,Jelinek:91c}.  In
particular, in a unigram cache, we form a unigram model from the most
recently spoken words (all those in the same article if article
markers are available, or a fixed number of previous words if not.)
This unigram cache can then be linearly interpolated with a
conventional n-gram.

Another type of cache model depends on the context.  For instance, we
could form a smoothed bigram or trigram from the previous words, and
interpolate this with the standard trigram.  In particular, we use
\begin{eqnarray*}
\lefteqn{\Px{trigram-cache}(w|w_1...w_{i-2}w_{i-1}) =} \\
&&\lambda \Psmooth(w|w_{i-2}w_{i-1}) + 
(1-\lambda) \Px{tricache}(w|w_1...w_{i-1}) 
\end{eqnarray*}
where $\Px{tricache}(w|w_1...w_{i-1})$ is a simple interpolated
trigram model, using counts from the preceding words in the same
document.

Yet another technique is to use conditional caching.  In this
technique, we weight the trigram cache differently depending on
whether or not we have previously seen the context or not.
\shortversion{The exact formulae are given in the extended version of
the paper, but basically, we only interpolate the trigram
cache $\Px{tricache}(w|w_{i-2}w_{i-1})$ if we have at least seen
$w_{i-1}$ in the cache.  Alternatively, we can interpolate a unigram,
bigram, and trigram cache, and use the bigram cache only if we have
seen $w_{i-1}$ and the trigram only if we have seen the pair $w_{i-2}w_{i-1}$.}
\shortversion{In addition, the actual
formulas we used allowed the caches to have a variable
weight, depending on the amount of context, but the optimal parameters
found set the variable factor very near zero.}

\longversionboring{We digress here for a moment to mention a trick that we use.  When
interpolating three probabilities $P_1(w)$, $P_2(w)$, and $P_3(w)$,
rather than use
$$
\lambda P_1(w) + \mu P_2(w) + (1-\lambda-\mu)P_3(w)
$$
we actually use
$$
\frac{\lambda}{\lambda+\mu+\nu} P_1(w) + \frac{\mu}{\lambda+\mu+\nu}P_2(w) + \frac{\nu}{\lambda+\mu+\nu} P_3(w)
$$
This allows us to simplify the constraints of the search, and we also
believe aids our parameter search routine, by adding a useful
dimension to search through.  It is particularly useful when sometimes
we do not use one of the three components.  In particular, for
conditional caching, we use the following formula:

\begin{eqnarray}
\lefteqn{\Px{conditionaltrigram}(w|w_1...w_{i-2}w_{i-1}) =}  \nonumber \\
&&
 \left\{ \begin{array}{ll}
\begin{array}{l}
\frac{\lambda}{\lambda+\mu+\nu} \Psmooth(w|w_{i-2}w_{i-1})  \nonumber \\
+ \frac{\mu}{\lambda+\mu+\nu} \Px{unicache}(w|w_1...w_{i-1})  \nonumber \\
+ \frac{\nu}{\lambda+\mu+\nu} \Px{tricache}(w|w_1...w_{i-1}) 
\end{array} 
  & \mbox {if $w_{i-1}$ in cache}  \nonumber \\
 \nonumber \\
\begin{array}{l}
\frac{\lambda}{\lambda+\mu} \Psmooth(w|w_{i-2}w_{i-1})  \nonumber \\
+ \frac{\mu}{\lambda+\mu} \Px{unicache}(w|w_1...w_{i-1})
\end{array}
  & \mbox {otherwise}
\end{array}
\right.
\label{eqn:condcache}
\end{eqnarray}

We tried one additional improvement.  We assume that the more data we
have, the more useful each cache is.  Thus, we make $\lambda$, $\mu$
and $\nu$ be linear functions of the amount of data in the cache
(number of words so far in the current document.)
$$
\lambda(\mathit{wordsincache}) = \lambda_\mathit{startweight} + \lambda_\mathit{multiplier} \times \frac {\min(\mathit{wordsincache}, \lambda_\mathit{maxwordsweight})}{\lambda_\mathit{maxwordsweight}}
$$
where, as usual, $\lambda_\mathit{startweight}$,
$\lambda_\mathit{multiplier}$ and $\lambda_\mathit{maxwordsweight}$
are parameters estimated on heldout data.  However, our parameter
search engine nearly always set $\lambda_\mathit{maxwordsweight}$ to
at or near the maximum value we allowed it to have, 1,000,000, while
assigning $\lambda_\mathit{multiplier}$ to a small value (typically
100 or less) meaning that the variable weighting was essentially
ignored.


Finally, we can try conditionally combinging unigram, bigram, and
trigram caches.  

\begin{eqnarray*}
\lefteqn{\Px{conditionaltrigram}(w|w_1...w_{i-2}w_{i-1}) =} \\
&&
 \left\{ \begin{array}{ll}
\begin{array}{l}
\frac{\lambda}{\lambda+\mu+\nu+\kappa} \Psmooth(w|w_{i-2}w_{i-1}) \\
+ \frac{\mu}{\lambda+\mu+\nu+\kappa} \Px{unicache}(w|w_1...w_{i-1}) \\
+ \frac{\nu}{\lambda+\mu+\nu+\kappa} \Px{bicache}(w|w_1...w_{i-1}) \\
+ \frac{\kappa}{\lambda+\mu+\nu+\kappa} \Px{tricache}(w|w_1...w_{i-1}) 
\end{array} 
  & \mbox {if $w_{i-2}w_{i-1}$ in cache} \\
\\
\begin{array}{l}
\frac{\lambda}{\lambda+\mu+\nu} \Psmooth(w|w_{i-2}w_{i-1}) \\
+ \frac{\mu}{\lambda+\mu+\nu} \Px{unicache}(w|w_1...w_{i-1}) \\
+ \frac{\nu}{\lambda+\mu+\nu} \Px{bicache}(w|w_1...w_{i-1}) 
\end{array} 
  & \mbox {if $w_{i-1}$ in cache} \\
\\
\begin{array}{l}
\frac{\lambda}{\lambda+\mu} \Psmooth(w|w_{i-2}w_{i-1}) \\
+ \frac{\mu}{\lambda+\mu} \Px{unicache}(w|w_1...w_{i-1})
\end{array}
  & \mbox {otherwise}
\end{array}
\right.
\end{eqnarray*}
}

\begin{figure}
$$\psfig{figure=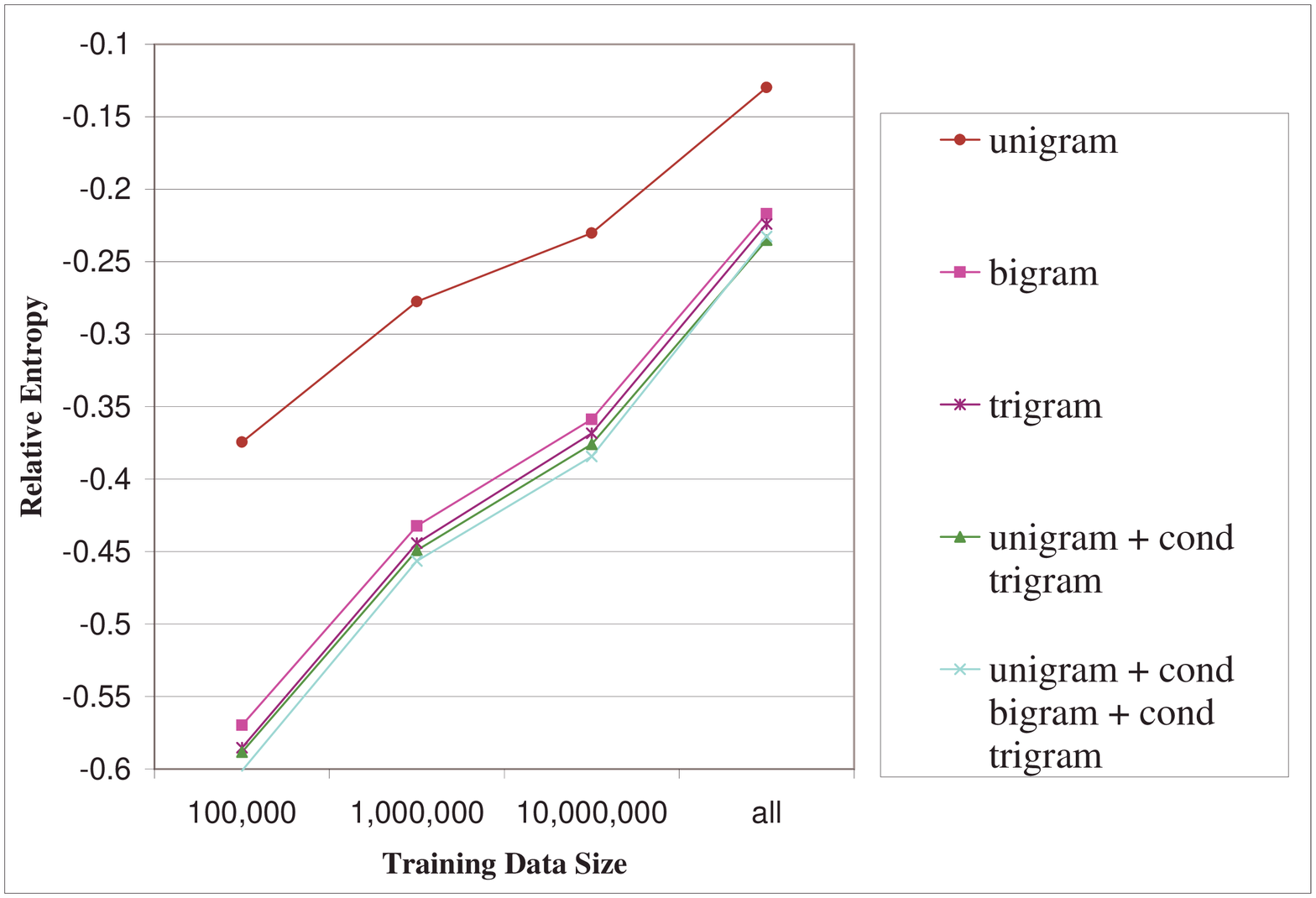,width=5.5in,angle=-90}$$
\caption{Five different cache models interpolated with trigram
compared to trigram baseline}
\label{fig:cachechart}
\end{figure}

Figure \ref{fig:cachechart} gives results of running each of these
five cache models.  All were interpolated with a Kneser-Ney smoothed
trigram.  Each of the n-gram cache models was smoothed using simple
interpolation, for technical reasons.  As can be seen, caching is
potentially one of the most powerful techniques we can apply, leading
to performance improvements of up to 0.6 bits on small data.  Even on
large data, the improvement is still substantial, up to 0.23 bits.  On
all data sizes, the n-gram caches perform substantially better than
the unigram cache, but which version of the n-gram cache is used appears
to make only a small difference.  

It should be noted that all of these results assume that the previous
words are known exactly.  In a speech recognition system however, many
product scenarios do not include user correction.  It is then possible
for a cache to ``lock-in'' errors.  For instance, if the user says
``recognize speech'' and the system hears ``wreck a nice beach'' then,
later, when the user says ``speech recognition'' the system may hear
``beach wreck ignition'', since the probability of ``beach'' will be
significantly raised.  Thus, getting improvements from caching in a
real product is potentially a much harder problem. 


\section{Sentence Mixture Models} 

\namecite{Iyer:99a,Iyer:94a} observed that within a corpus,
there may be several different sentence types; these sentence types
could be grouped by topic, or style, or some other criterion.  No
matter how they are grouped, by modeling each sentence type
separately, improved performance can be achieved.  For instance, in
Wall Street Journal data, we might assume that there are three
different sentence types: financial market sentences (with a great
deal of numbers and stock names), business sentences (promotions,
demotions, mergers), and general news stories.  We can compute the
probability of a sentence once for each sentence type, then take a
weighted sum of the probabilities across sentence types.  Because
long-distance correlations within a sentence (lots of numbers, or lots
of promotions) are captured by such a model, the overall model is
better.  Of course, in general, we do not know the sentence type until
we have heard the sentence.  Therefore, instead, we treat the
sentence type as a hidden variable.

Let $s_j$ denote the condition that the sentence under consideration
is a sentence of type $j$.  Then the probability of the sentence,
given that it is of type $j$ can be written as
$$
\prod_{i=1}^N P(w_i|w_{i-2}w_{i-1} s_j )
$$
Sometimes, the global model (across all sentence types) will be better
than any individual sentence type.  Let $s_0$ be a special context
that is always true:
$$
P(w_i|w_{i-2}w_{i-1} s_0) = P(w_i|w_{i-2}w_{i-1})
$$
Let there be $S$ different sentence types ($4 \leq S \leq 8$ is
typical); let $\sigma_0...\sigma_S$ be sentence interpolation
parameters optimized on held-out data subject to the constraint
$\sum_{j=0}^S \sigma_j = 1$.  The overall probability of a sentence
$w_1...w_n$ is equal to
\begin{equation}
\sum_{j=0}^S \sigma_j \prod_{i=1}^N P(w_i|w_{i-2}w_{i-1} s_j)
\label{eqn:sentence}
\end{equation}
Equation \ref{eqn:sentence} can be read as saying that there is a
hidden variable, the sentence type; the prior probability for each
sentence type is $\sigma_j$.  We compute the probability of a test
sentence once for each sentence type, and then sum these probabilities
according to the prior probability of that sentence type.

The probabilities $P(w_i|w_{i-2}w_{i-1} s_j)$ may suffer from data
sparsity, so they are linearly interpolated with the global model
$P(w_i|w_{i-2}w_{i-1})$, using interpolation weights optimized on
held-out data.  

\longversion{$$
\sum_{j=0}^S \sigma_j \prod_{i=1}^N \lambda_j P(w_i|w_{i-2}w_{i-1} s_j) + (1-\lambda_j)P(w_i|w_{i-2}w_{i-1})
$$}

Sentence types for the training data were found by using the same
clustering program used for clustering words; in this case, we tried
to minimize the sentence-cluster unigram perplexities.  That is, let
$s(i)$ represent the sentence type assigned to the sentence that word
$i$ is part of.  (All words in a given sentence are assigned to the
same sentence type.)  We tried to put sentences into clusters in such
a way that $\prod_{i=1}^N P(w_i | s(i))$ was maximized.  This is a
much simpler technique than that used by \namecite{Iyer:99a}.
\longversion{They use a two stage process, the first stage of which is
a unigram-similarity agglomerative clustering method; the second stage
is an EM-based n-gram based reestimation.  Also, their technique used
soft clusters, in which each sentence could belong to multiple
clusters.}  We assume that their technique results in better models
than ours.

\begin{figure}
$$\psfig{figure=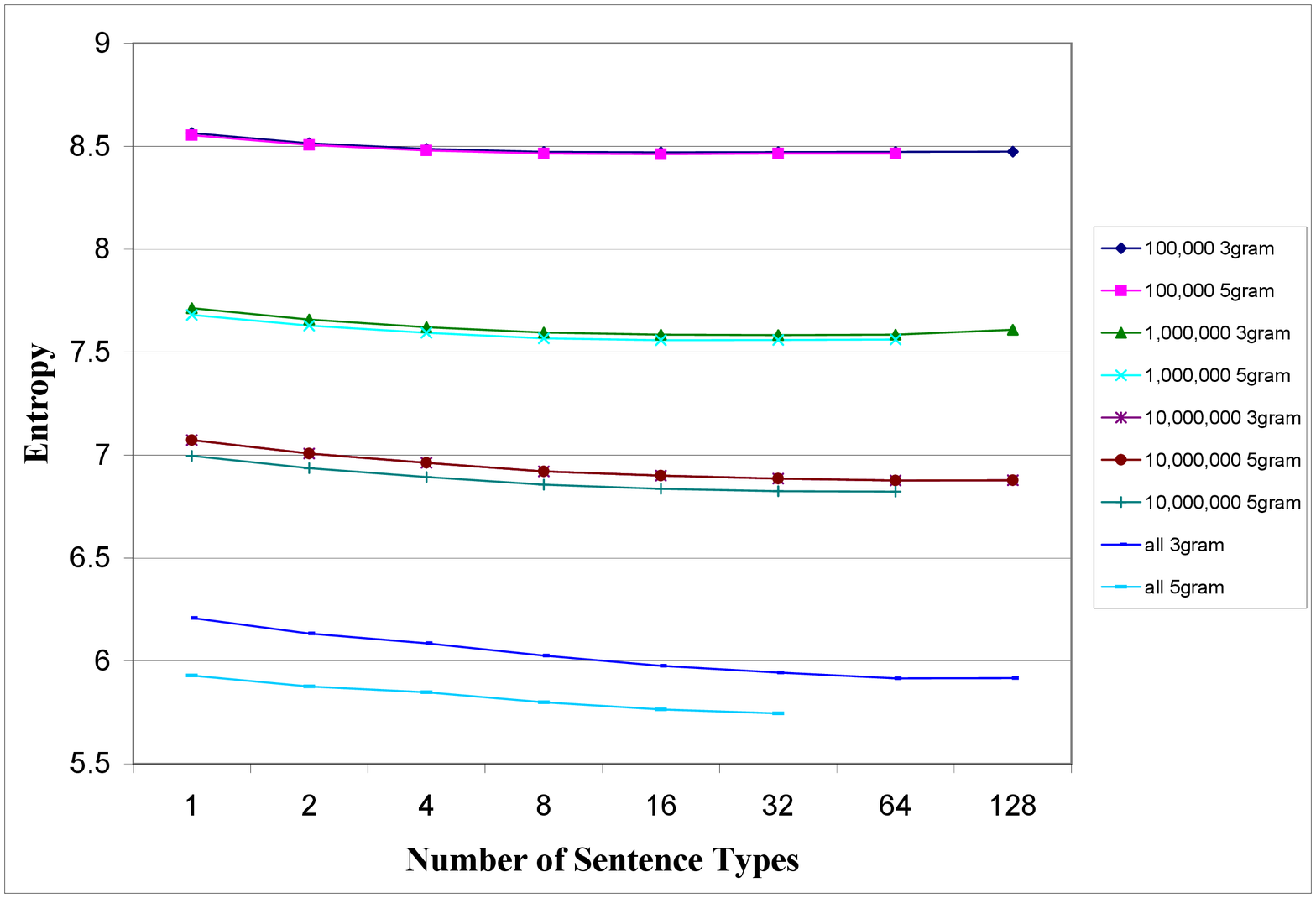,width=5.5in,angle=-90}$$
\caption{Number of sentence types versus entropy}
\label{fig:sentencechart}
\end{figure}

\longversion{
\begin{figure}
$$\psfig{figure=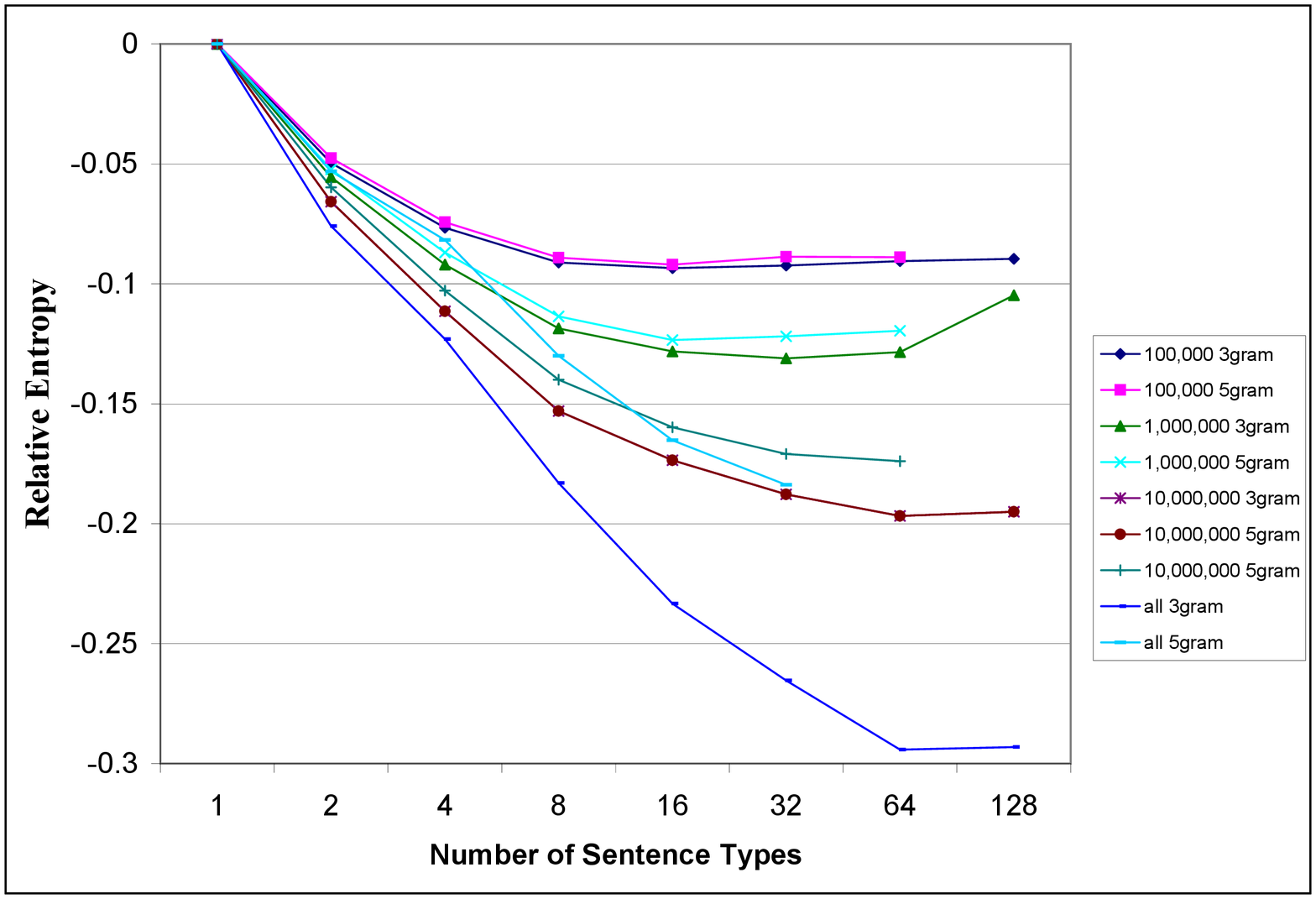,width=5.5in,angle=-90}$$
\caption{Number of sentence types versus entropy, relative to baseline}
\label{fig:sentencechartrelative}
\end{figure}
}

We performed a fairly large number of experiments on sentence mixture
models.  We sought to study the relationship between training data
size, n-gram order, and number of sentence types.  We therefore ran a
number of experiments using both trigrams and 5-grams, at our standard
data sizes, varying the number of sentence types from 1 (a normal
model without sentence mixtures) to 128.  All experiments were done
with Kneser-Ney smoothing.  The results are shown in Figure
\ref{fig:sentencechart}.  \longversion{We give the same results again,
but graphed relative to their respective n-gram model baselines, in
Figure \ref{fig:sentencechartrelative}.  Note that we do not trust
results for 128 mixtures; for these experiments, we used the same
20,000 words of heldout data used in our other experiments.  With 128
sentence types, there are 773 parameters, and the system may not have
had enough heldout data to accurately estimate the parameters.  In
particular, the plateau shown at 128 for 10,000,000 and all training
data does not show up in the heldout data.  Ideally, we would run this
experiment with a larger heldout set, but it already required 5.5 days
with 20,000 words, so this is impractical.}  \shortversion{Note,
however, that we do not trust results for 128 mixtures because there
may not have been enough heldout data to correctly estimate the
parameters; see the extended version of the paper for details.}

The results are very interesting for a number of reasons.  First, we
suspected that sentence mixture models would be more useful on larger
training data sizes, and indeed they are; with 100,000 words, the most
improvement from a sentence mixture model is only about .1 bits, while
with 284,000,000 words, it is nearly .3 bits.  This bodes well for the
future of sentence mixture models: as computers get faster and larger,
training data sizes should also increase.  Second, we had suspected
that because both 5-grams and sentence mixture models attempt to model
long distance dependencies, the improvement from their combination
would be less than the sum of the individual improvements.  As can be
seen in Figure \ref{fig:sentencechart}, for 100,000 and 1,000,000
words of training data, the difference between trigrams and 5-grams is
very small anyway, so the question is not very important.  For
10,000,000 words and all training data, there is some negative
interaction.  For instance, with 4 sentence types on all training
data, the improvement is 0.12 bits for the trigram, and 0.08 bits for
the 5-gram.  Similarly, with 32 mixtures, the improvement is .27 on
the trigram and .18 on the 5-gram.  So, approximately one third of the
improvement seems to be correlated.

\namecite{Iyer:99a} reported experiments on both 5-mixture components and
8 components and found no significant difference, using 38 million
words of training data.  However, our more thorough investigation
shows that indeed there is substantial room for improvement by using
larger numbers of mixtures, especially when using more training data,
and that this potential extends at least to 64 sentence types on our
largest size.  This is an important result, leading to almost twice
the potential improvement of using only a small number of components.

We think this new result is one of the most interesting in our
research.  In particular, the techniques we used here were relatively
simple, and many extensions to these techniques might lead to even
larger improvements.  For instance, rather than simply smoothing a
sentence type with the global model, one could create sentence types
and supertypes, and then smooth together the sentence type with its
supertype and with the global model, all combined.  This would
alleviate the data sparsity effects seen with the largest numbers of
mixtures.  

\shortversion{Our sentence mixture model results are encouraging, but
disappointing when compared to previous results.  While
\namecite{Iyer:99a} achieve about 19\% perplexity reduction and about
3\% word error rate reduction with 5 mixtures, on similar data we
achieve only about 9\% and (as we will show later) 1.3\% reductions
with 4 mixtures.  In the extended version of this paper, we speculate
on the potential causes of the differences.  We suspect that our
clustering technique, much simpler than theirs, or differences in the
exact composition of the data sets, account for the differences.}

\longversionboring{Our sentence mixture model results are encouraging, but
disappointing when compared to previous results.  While
\namecite{Iyer:99a} achieve about 19\% perplexity reduction and about
3\% word error rate reduction with 5 mixtures, on similar data we
achieve only about 9\% and (as we will show later) 1.3\% reductions
with 4 mixtures.  The largest difference we are aware of between their
system and ours is the difference in clustering technique: we used a
fairly simple technique, and they used a fairly complex one.  Other
possibilities include the different smoothing that they used
(Witten-Bell smoothing) versus our Katz or Interpolated Kneser-Ney
smoothing; and the fact that they used five clusters while we used
four.  However, Katz and Interpolated Kneser-Ney are very different
techniques, but, as we will report later, sentence mixture models
produce about the same improvement with both, so we do not think the
difference in smoothing explains the different results.  Also,
\namecite{Iyer:99a} found no significant difference when using 8
clusters instead of 5, and we found only a 4\% difference when using 8
instead of 4 on similar data.  It is noteworthy that Iyer and
Ostendorf have a baseline perplexity of 211, versus our baseline
perplexity of 95, with the same training set, but different
vocabularies and test sets.  Perhaps whatever unknown factor accounts
for this difference in baseline perplexities gives more room for
improvement from sentence mixture models.  It is worth noting that our
improvement from caching is also much less than Iyer and Ostendorf's:
about 7.8\% versus their 16.7\%.  Our cache implementations were very
similar, the main difference being their exclusion of stop-words.
This adds support to the ``different baseline/test condition''
explanation.  If the difference in perplexity reduction is due to some
difference in the mixture model implementation, rather than in the
test conditions, then an additional 10\% perplexity reduction could be
achieved, an amount that merits additional exploration.}

Sentence mixture models can also be useful when combining very
different language model types.  For instance, \namecite{Jurafsky:95a}
uses a sentence mixture model to combine a stochastic context-free
grammar (SCFG) model with a bigram model, resulting in marginally
better results than either model used separately.  The model of
Jurafsky \myetal is actually of the form
\begin{eqnarray*}
\lefteqn{P(w_i|w_1...w_{i-1}) = } \\
&& P(\mbox{SCFG}|w_1...w_{i-1}) \times P(w_i|w_1...w_{i-1},\mbox{SCFG}) \\
&& + P(\mbox{bigram}|w_1...w_{i-1}) \times P(w_i|w_1...w_{i-1},\mbox{bigram})
\end{eqnarray*}
which turns out to be equivalent to a model in the form of Equation
\ref{eqn:sentence}.  \longversion{This version of the equations has
the advantage that when used in a stack-decoder, it allows sentence
mixture models to be used with relatively little overhead, compared to
Equation \ref{eqn:sentence}}  \namecite{Charniak:01a}, as discussed
in Section \ref{sec:structured}, uses a sentence level mixture model
to combine a linguistic model with a trigram model, achieving
significant perplexity reduction.

\section{Combining techniques} 
\label{sec:combination}

In this section, we present additional results on combining
techniques.  While each of the techniques we have presented works well
separately, we will show that some of them work together
synergistically, and that some of them are partially redundant.  For
instance, we have shown that the improvement from Kneser-Ney modeling
and 5-gram models together is larger than the improvement from either
one by itself.  Similarly, as we have already shown, the improvement
from sentence mixture models when combined with 5-grams is only about
$\frac{2}{3}$ of the improvment of sentence mixture models by
themselves, because both techniques increase data sparsity.  In this
section, we systematically study three issues: what effect does
smoothing have on each technique; how much does each technique help
when combined with all of the others; and how does each technique
affect word error rate, separately and together.

There are many different ways to combine techniques.  The most obvious
way to combine techniques is to simply linearly interpolate them, but
this is not likely to lead to the largest possible improvement.
Instead, we try to combine concepts.  To give a simple example, recall
that a fullibmpredict clustered trigram is of the form:
$$
\begin{array}{l}
(\lambda P(W|w_{i-2}w_{i-1}) + (1-\lambda) P(W|W_{i-2}W_{i-1})) \times \\
(\mu P(w|w_{i-2}w_{i-1}W) + (1-\mu) P(w|W_{i-2}W_{i-1}W)
\end{array}
$$
One could simply interpolate this clustered trigram with a normal
5-gram, but of course it makes much more sense to combine the concept
of a 5-gram with the concept of fullibmpredict, using a clustered 5-gram:
$$
\begin{array}{l}
(\lambda P(W|w_{i-4}w_{i-3}w_{i-2}w_{i-1}) + (1-\lambda) P(W|W_{i-4}W_{i-3}W_{i-2}W_{i-1})) \times \\
(\mu P(w|w_{i-4}w_{i-3}w_{i-2}w_{i-1}W) + (1-\mu) P(w|W_{i-4}W_{i-3}W_{i-2}W_{i-1}W)
\end{array}
$$
We will follow this idea of combining concepts, rather than simply
interpolating throughout this section.  This tends to result in good
performance, but complex models. 

\longversionboring{When we combine sentence mixture models with caches, we
need to answer additional questions.  \namecite{Iyer:99a}
used separate caches for each sentence type, putting each word into
the cache for the most likely sentence type.  This would have required
a great deal of additional work in our system; also, we did pilot
experiments combining sentence mixture models and caches with only a
single global cache, and found that the improvement from the
combination was nearly equal to the sum of the individual
improvements.  Since Iyer and Ostendorf also get an improvement nearly
equal to the sum, we concluded that not using separate caches was a
reasonable combination technique.}


Our overall combination technique is somewhat complicated.  At the
highest level, we use a sentence mixture model, in which we sum over
sentence-specific models for each sentence type.  Within a particular
sentence mixture model, we combine different techniques with
predictive clustering.  That is, we combine sentence-specific, global,
cache, and global skipping models first to predict the cluster of the
next word, and then again to predict the word itself given the
cluster. 

For each sentence type, we wish to linearly interpolate the
sentence-specific 5-gram model with the global 5-gram model, the three
skipping models, and the two cache models.  Since we are using
fullibmpredict clustering, we wish to do this based on both words and
clusters.  Let $\lambda_{1,j}...\lambda_{12,j}$,
$\mu_{1,j}...\mu_{12,j}$ be interpolation parameters.  Then, we define
the following two very similar functions.  First,\footnote{This formula
is actually an oversimplification because the values $\lambda_{11,j}$
and $\lambda_{12,j}$ depend on the amount of training data in a linear
fashion, and if the context $w_{i-1}$ does not occur in the cache,
then the trigram cache is not used.  In either case, the values of the
$\lambda$'s have to be renormalized for each context so that they sum
to 1.}

\begin{eqnarray*}
\lefteqn{\mbox{\it sencluster}_j(W, w_{i-4}...w_{i-1}) =  } \\
&& \lambda_{1,j} P(W | w_{i-4} w_{i-3} w_{i-2} w_{i-1} s_j) + \lambda_{2,j} P(W | W_{i-4} W_{i-3} W_{i-2} W_{i-1} s_j) + \\
&& \lambda_{3,j} P(W | w_{i-4} w_{i-3} w_{i-2} w_{i-1} ) + \lambda_{4,j} P(W | W_{i-4} W_{i-3} W_{i-2} W_{i-1} ) + \\
&& \lambda_{5,j} P(W | w_{i-4} w_{i-3} w_{i-1} ) + \lambda_{6,j} P(W | W_{i-4} W_{i-3} W_{i-1} ) + \\
&& \lambda_{7,j} P(W | w_{i-4} w_{i-2} w_{i-1} ) + \lambda_{8,j} P(W | W_{i-4} W_{i-2} W_{i-1} ) + \\
&& \lambda_{9,j} P(W | w_{i-4} w_{i-3} w_{i-2} ) + \lambda_{10,j} P(W | W_{i-4} W_{i-3} W_{i-2} ) + \\
&& \lambda_{11,j} \Px{unicache}(W) + \lambda_{12,j} \Px{tricache}(W | w_{i-2} w_{i-1}) 
\end{eqnarray*}
Next, we define the analogous function for predicting words given
clusters:
\begin{eqnarray*}
\lefteqn{\mbox{\it senword}_j(w, w_{i-4}...w_{i-1}, W) =  } \\
&& \mu_{1,j} P(w | w_{i-4} w_{i-3} w_{i-2} w_{i-1} W s_j) + \mu_{2,j} P(w | W_{i-4} W_{i-3} W_{i-2} W_{i-1} W s_j) + \\
&& \mu_{3,j} P(w | w_{i-4} w_{i-3} w_{i-2} w_{i-1} W) + \mu_{4,j} P(w | W_{i-4} W_{i-3} W_{i-2} W_{i-1} W) + \\
&& \mu_{5,j} P(w | w_{i-4} w_{i-3} w_{i-1} W) + \mu_{6,j} P(w | W_{i-4} W_{i-3} W_{i-1} W) + \\
&& \mu_{7,j} P(w | w_{i-4} w_{i-2} w_{i-1} W) + \mu_{8,j} P(w | W_{i-4} W_{i-2} W_{i-1} W) + \\
&& \mu_{9,j} P(w | w_{i-4} w_{i-3} w_{i-2} W) + \mu_{10,j} P(w | W_{i-4} W_{i-3} W_{i-2} W) + \\
&& \mu_{11,j} \Px{unicache}(w|W) + \mu_{12,j} \Px{tricache}(w | w_{i-2} w_{i-1}W) 
\end{eqnarray*}

Now, we can write out our probability model:
\begin{eqnarray}
\lefteqn{P_{\mbox{\it \footnotesize everything}}(w_1...w_N) = } \nonumber \\
&&\!\!\!\!\!\!\sum_{j=0}^S \!\sigma_j \! \prod_{i=1}^N \!
\mbox{\it sencluster}_j(W_i, w_{i\!-\!4}...w_{i\!-\!1})\! \times\!
\mbox{\it senword}_j(w_i, w_{i\!-\!4}...w_{i\!-\!1}, W_i) \label{eqn:everything}
\end{eqnarray}

Clearly, combining all of these techniques together is not easy, but
as we will show, the effects of combination are very roughly additive,
and the effort is worthwhile.

\begin{figure}
$$\psfig{figure=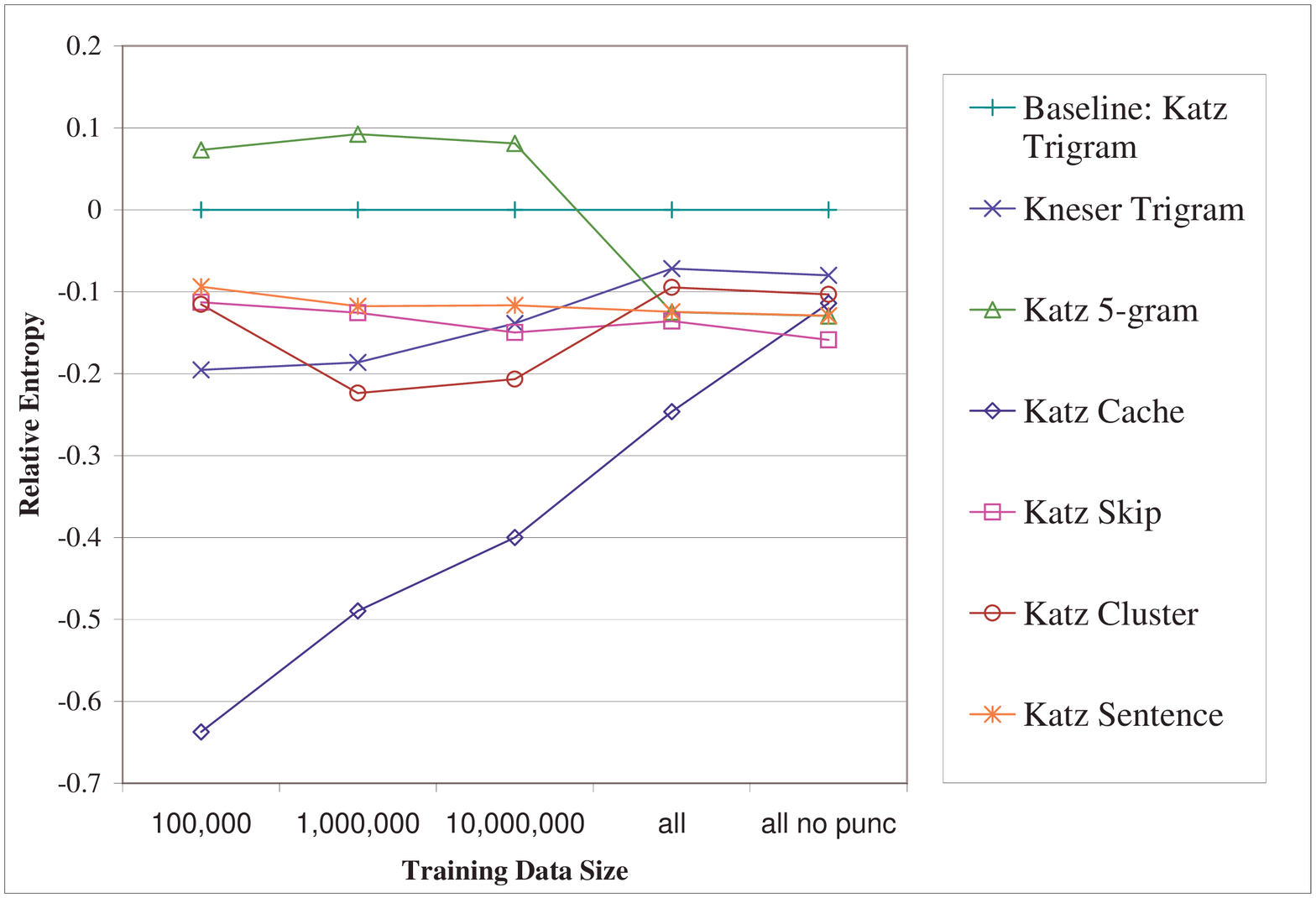,width=5in,angle=-90}$$
\caption{Relative Entropy of Each Technique versus Katz Trigram Baseline}
\label{fig:combinekatz}
\end{figure}

\begin{figure}
$$\psfig{figure=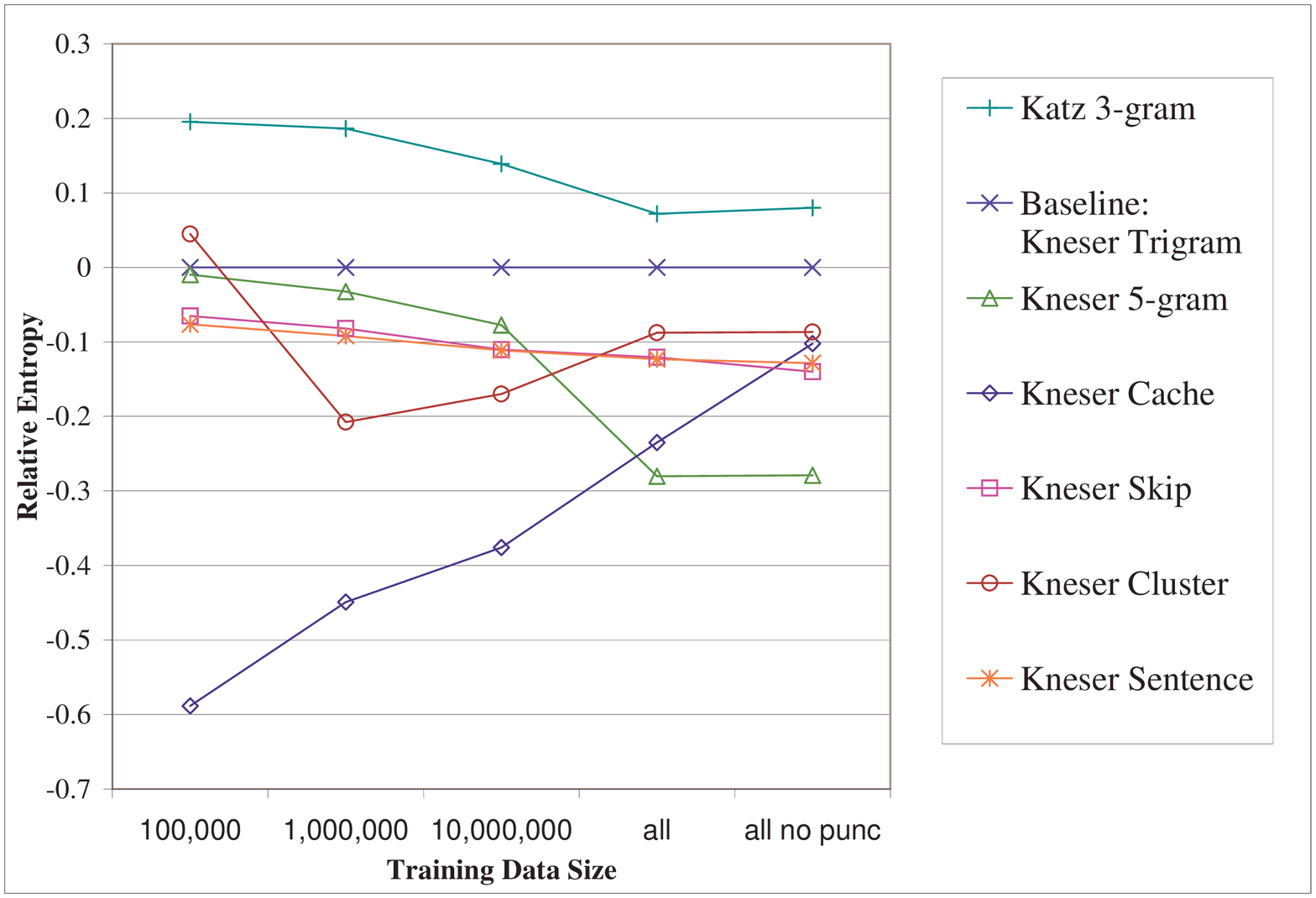,width=5in,angle=-90}$$
\caption{Relative Entropy of Each Technique versus Kneser-Ney Trigram Baseline}
\label{fig:combinekneser}
\end{figure}

\begin{figure}
$$\psfig{figure=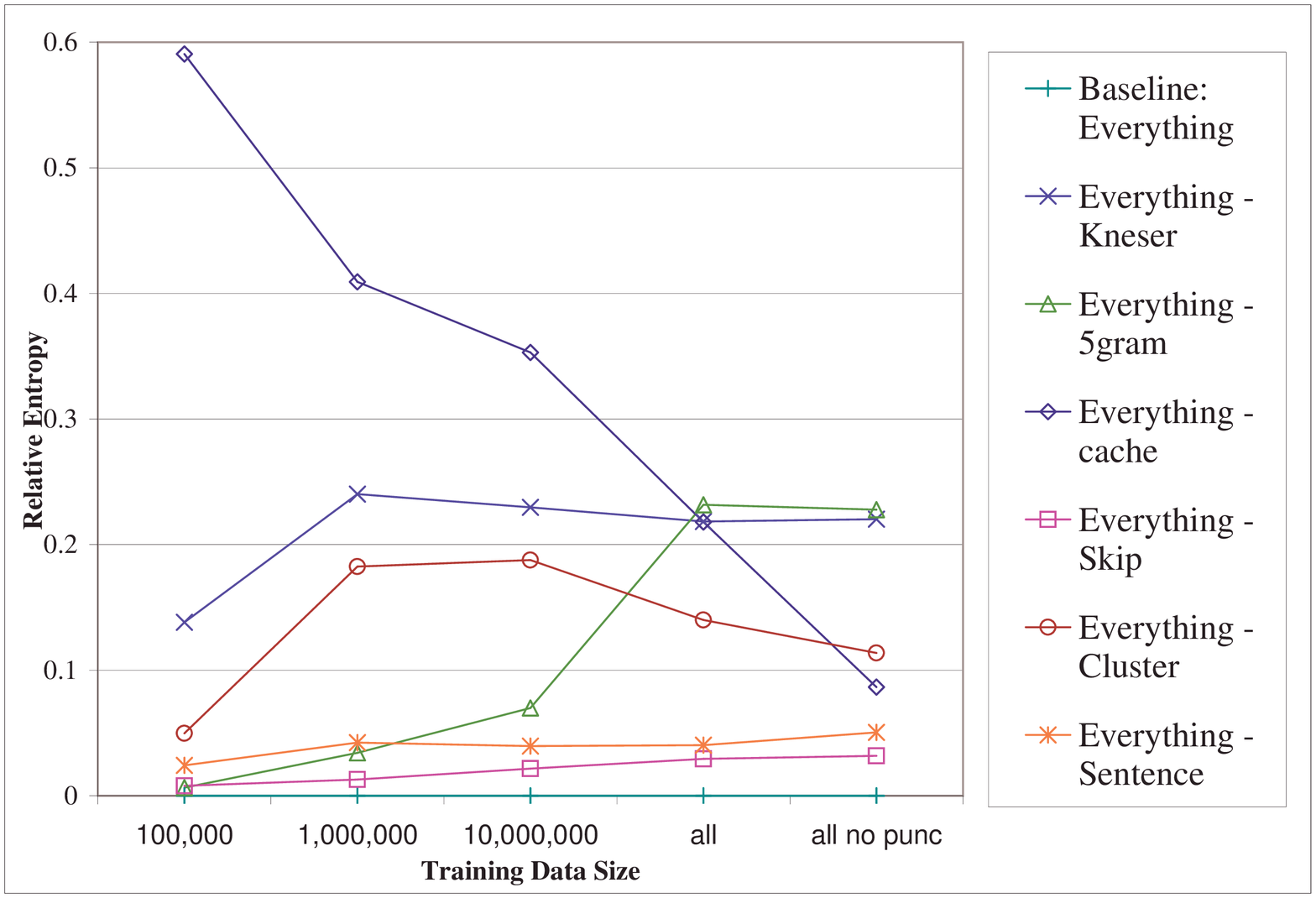,width=5in,angle=-90}$$
\caption{Relative Entropy of {\it Removing} Each Technique versus All
Techniques Combined Baseline}
\label{fig:combineeverything}
\end{figure}

\begin{figure}
$$\psfig{figure=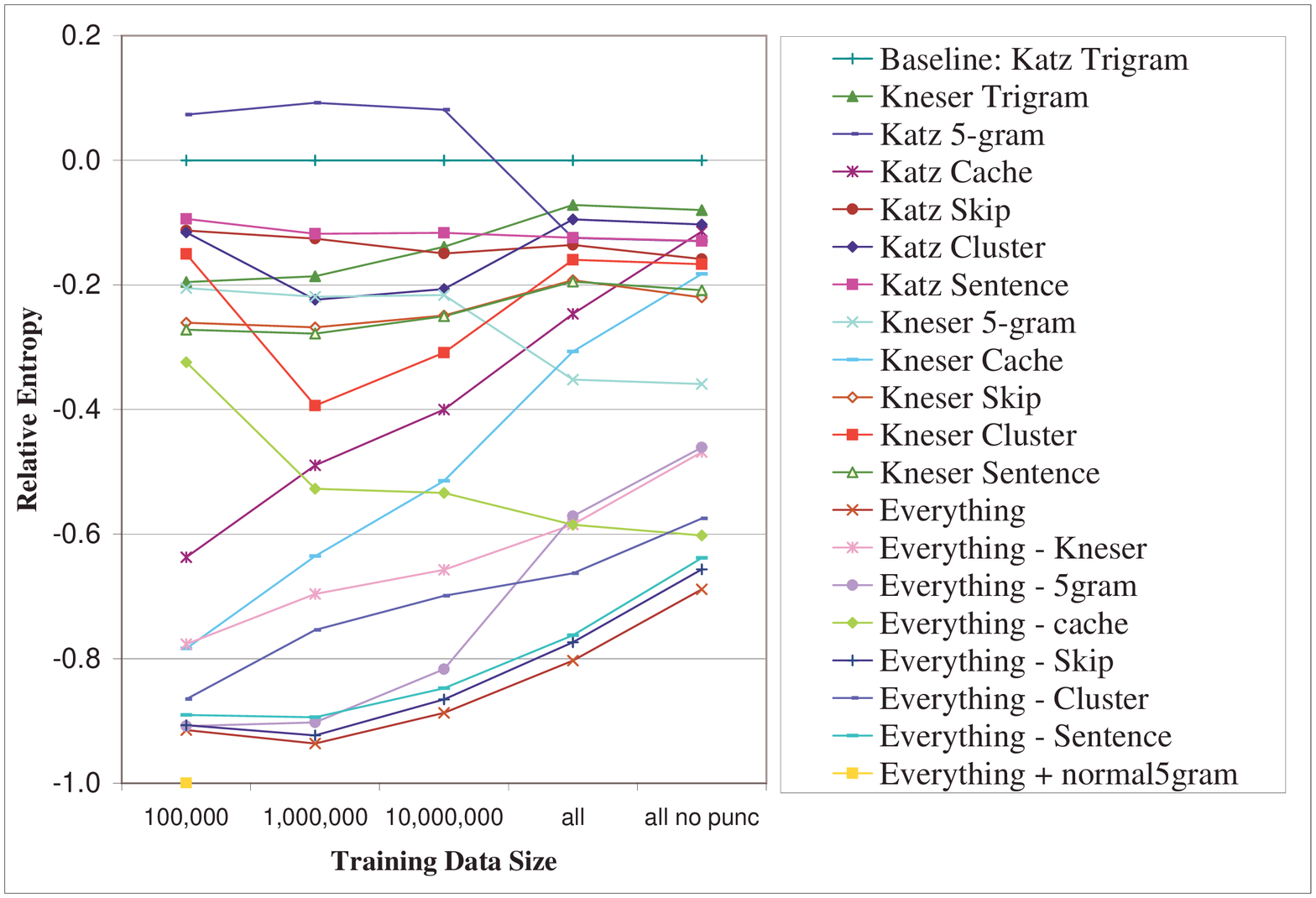,width=5in,angle=-90}$$
\caption{All Techniques Together versus Katz Trigram Baseline}
\label{fig:combineall}
\end{figure}

We performed several sets of experiments.  In these experiments, when
we perform caching, it is with a unigram cache and conditional trigram
cache; when we use sentence mixture models, we use 4 mixtures; when we
use trigram skipping, it is $w \un y$ and $wx \un$; and when we use
5-gram skipping it is $vw \un y$ interpolated with $v \un xy$ and $vwx
\un$.  Our word error rate experiments were done without punctuation,
so, to aid comparisons, we perform additional entropy experiments in
this section on ``all-no-punc'', which is the same as the ``all'' set,
but without punctuation.

In the first set of experiments, we used each technique separately,
and Katz smoothing.  The results are shown in Figure
\ref{fig:combinekatz}.  Next, we performed experiments with the same
techniques, but with Kneser-Ney smoothing; the results are shown in
Figure \ref{fig:combinekneser}.  The results are similar for all
techniques independent of smoothing, except 5-grams, where Kneser-Ney
smoothing is clearly a large gain; in fact, without Kneser-Ney
smoothing, 5-grams actually hurt at small and medium data sizes.  This
is a wonderful example of synergy, where the two techniques together
help more than either one separately.  Caching is the largest gain at
small and medium data sizes, while, when combined with Kneser-Ney
smoothing, 5-grams are the largest gain at large data sizes.  Caching
is still key at most data sizes, but the advantages of Kneser-Ney
smoothing and clustering are clearer when they are combined with the
other techniques.

In the next set of experiments, shown in Figure
\ref{fig:combineeverything}, we tried removing each technique from the
combination of all techniques (Equation \ref{eqn:everything}).  The
baseline is all techniques combined -- ``everything'', and then we
show performance of, for instance, everything except Kneser-Ney,
everything except 5-gram models, etc.  In Figure
\ref{fig:combineeverything} we show all techniques together versus a
Katz smoothed trigram.  We add one additional point to this graph.
With 100,000 words, our Everything model was at .91 bits below a
normal Katz model, an excellent result, but we knew that the 100,000
word model was being hurt by the poor performance of fullibmpredict
clustering at the smallest data size.  We therefore interpolated in a
normal 5-gram at the word level, a technique indicated as ``Everything
+ normal5gram.''  This led to an entropy reduction of 1.0061 -- 1 bit.
This gain is clearly of no real-world value -- most of the entropy
gains at the small and medium sizes come from caching, and caching
does not lead to substantial word error rate reductions.  However, it
does allow a nice title for the paper.  Interpolating the normal
5-gram at larger sizes led to essentially no improvement.

\begin{figure}
$$\psfig{figure=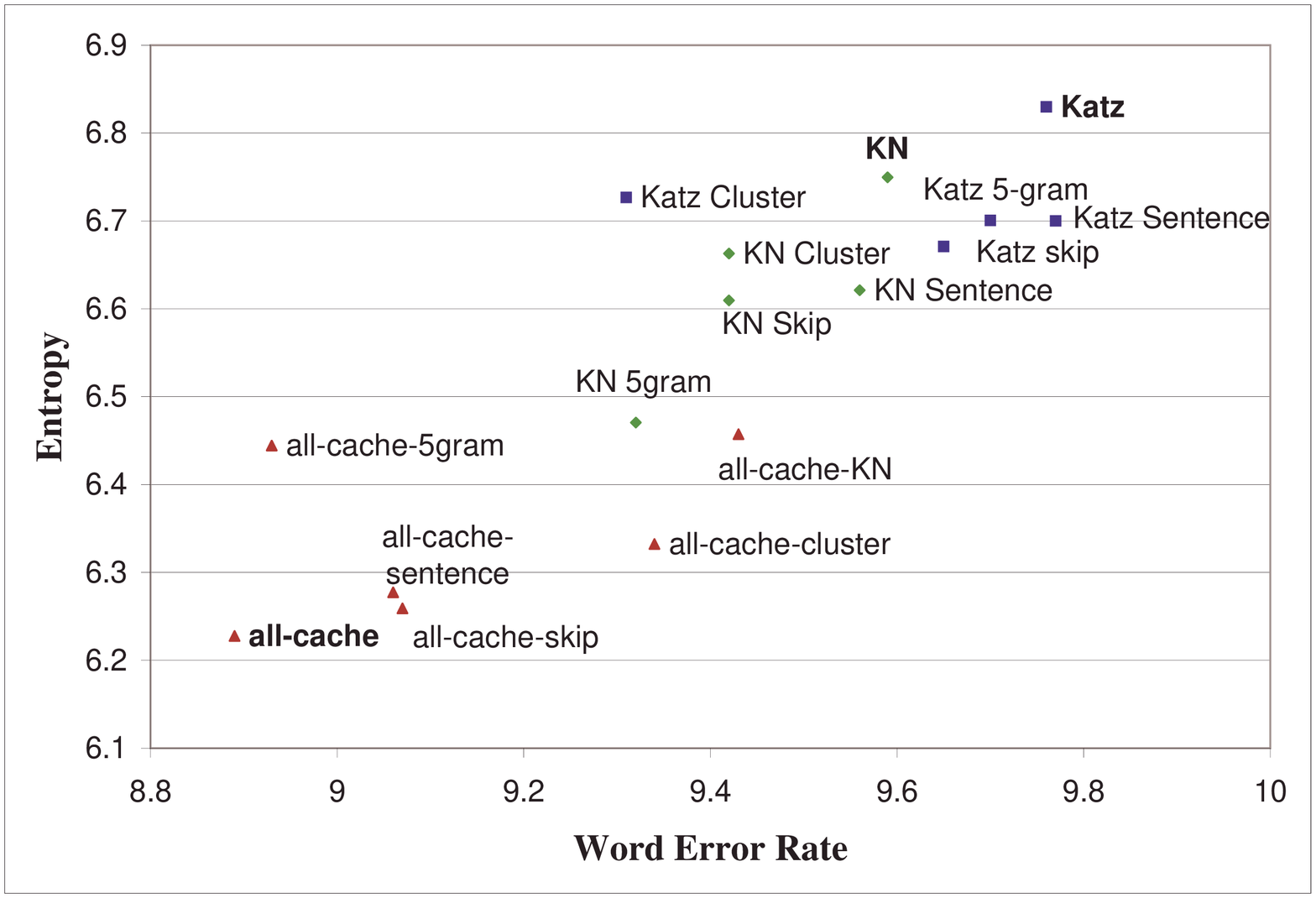,width=5in,angle=-90}$$
\caption{Word Error Rate versus Entropy}
\label{fig:combineerror}
\end{figure}

We also performed word-error rate experiments rescoring 100-best lists
of WSJ94 dev and eval, about 600 utterances.  The 1-best error rate
for the 100-best lists was 10.1\% (our recognizer's models were
slightly worse than even the baseline used in rescoring) and the
100-best error rate (minimum possible from rescoring) was 5.2\%.  We
were not able to get word-error rate improvements by using caching
(when the cache consisted of the output of the recognizer), and were
actually hurt by the use of caching when the interpolation parameters
were estimated on correct histories, rather than on recognized
histories.  Figure \ref{fig:combineerror} shows word-error rate
improvement of each technique, either with Katz smoothing, Kneser-Ney
smoothing, or removed from Everything, except caching.  The most
important single factor for word-error rate was the use of Kneser-Ney
smoothing, which leads to a small gain by itself, but also makes
skipping, and 5-grams much more effective.  Clustering also leads to
significant gains.  
In every case except clustering, the Kneser-Ney smoothed model has
lower word-error rate than the corresponding Katz smoothed model.  The
strange clustering result (the Katz entropy is higher) might be due to
noise, or might be due to the fact that we optimized the number of
clusters separately for the two systems, optimizing perplexity, perhaps leading to a number
of clusters that was not optimal for word error rate reduction.
Overall, we get an 8.9\% word error rate reduction
over a Katz smoothed baseline model.  This is very good, although not
as good as one might expect from our perplexity reductions.  This is
probably due to our rescoring of n-best lists rather than integrating
our language model directly into the search, or rescoring large
lattices.

\section{Implementation Notes} 
\label{sec:implementationnotes}

Actually implementing the model described here is not straightforward.
We give here a few notes on the most significant implementation
tricks, some of which are reasonably novel, and in the appendix\shortversion{ of the extended paper} give more details.  First, we
describe our parameter search technique.  Next, we discuss techniques
we used to deal with the very large size of the models constructed.
Then, we consider architectural strategies that made the research
possible.  Finally, we give a few hints on implementing our clustering
methods.

The size of the models required for this research is very large.  In
particular, many of the techniques have a roughly multiplicative
effect on data sizes: moving to five-grams from trigrams results in at
least a factor of two increase; fullibmpredict clustering results in nearly a factor of 4 increase; and the
combination of sentence-mixture models and skipping leads to about
another factor of four.  The overall model size then, is, very
roughly, 32 times the size of a standard trigram model.  Building and
using such a complex model would be impractical.

Instead, we use a simple trick.  We first make a pass through the test
data (either text, or n-best lists), and the heldout data (used for
parameter optimization), and determine the complete set of values we
will need for our experiments.  Then, we go through the training data,
and record only these values.  This drastically reduces the amount of
memory required to run our experiments, reducing it to a managable 1.5
gigabytes roughly.  Another trick we use is to divide the test data
into pieces -- the less test data there is, the fewer values we need
to store.  The appendix \shortversion{of the extended paper} describes
some ways that we verified that this ``cheating'' resulted in the same
results that a non-cheating model would have gotten.

Careful design of the system was also necessary.  In particular,
we used a concept of a ``model'', an abstract object, with a set of
parameters, that could return the probability of a word or class given
a history.  We created models that could compose other models, by
interpolating them at the word level, the class level, or the sentence
level, or even by multiplying them together as done in predictive
clustering.  This allowed us to compose primitive models that
implemented caching, various smoothing techniques, etc., in a large
variety of ways.

Both our smoothing techniques and interpolation require the
optimization of free parameters.  In some cases, these free parameters
can be estimated from the training data by leaving-one-out techniques,
but better results are obtained by using a Powell search of the
parameters, optimizing the perplexity of held-out data
\mycite{Chen:99a}, and that is the technique used here.  This allowed
us to optimize all parameters jointly, rather than say optimizing one
model, then another, then their interpolation parameters, as is
typically done.  It also made it relatively easy to, in essentially
every experiment in this paper, find the optimal parameter settings
for that model, rather than use suboptimal guesses or results from
related models.\footnote{The only exception was that for Katz smoothed
``everything'' models we estimated the number of clusters from simple
Katz clustered models; large Katz smoothed models are extremely time
consuming because of the need to find the $\alpha$'s after each
potential parameter change.}

Although all smoothing algorithms were reimplemented for this research
(reusing only a small amount of code), the details closely follow
\namecite{Chen:99a}.  This includes our use of additive smoothing of
the unigram distribution for both Katz smoothing and Kneser-Ney
smoothing.  That is, we found a constant $b$ which was added to all
unigram counts; this leads to better performance in small
training-data situations, and allowed us to compare perplexities
across different training sizes, since no unigram received 0 counts,
meaning 0 probabilities were never returned.

There is no shortage of techniques for generating clusters, and there
appears to be little evidence that different techniques that optimize
the same criterion result in a significantly different quality of
clusters.  We note, however, that different algorithms may require
significantly different amounts of run time.  In particular,
agglomerative clustering algorithms may require significantly more
time than top-down, splitting algorithms.  Within top-down, splitting
algorithms, additional tricks can be used, including the techniques of
Buckshot \mycite{Cutting:92a}.  We also use computational tricks adapted
from \namecite{Brown:90a}.  Many more details about the clustering
techniques used are given in \shortversion{the appendix of the
extended version of this paper.}\longversion{Appendix \ref{sec:clusterdetails}.}

\section{Other techniques}
\label{sec:othertechniques}

In this section, we briefly discuss several other techniques that have
received recent interest for language modeling; we have done a few
experiments with some of these techniques.  These techniques include
maximum entropy models, \longversion{whole sentence maximum entropy
models, } latent semantic analysis, parsing based models, and neural
network based models.  \namecite{Rosenfeld:00a} gives a much broader,
different perspective on the field, as well as additional references
for the techniques discussed in this paper.

\subsection{Maximum Entropy Models}
\shortversion{ Maximum entropy models \mycite{Darroch:72a} have
received a fair amount of attention since their introduction for
language modeling by \namecite{Rosenfeld:94a}, who achieved excellent
results -- up to 39\% perplexity reductions.  Maximum entropy models
allow arbitrary constraints to be combined.  For instance, rather than
simply linearly interpolating the components of a skipping model, the
components of a clustering model, and a baseline n-gram model, each of
these models can be expressed as a constraint on how often we expect
words to occur in some context.  We can then build a model that
satisfies all of these constraints, while still being as smooth as
possible.  This is a highly seductive quality.  In addition, recent
improvements in maximum entropy smoothing \mycite{Chen:99b} and
maximum entropy training speedups \mycite{Goodman:01a}, are sources
for optimism.  Unfortunately, typically maximum entropy models are
compared only to trigram models, rather than to comparable n-gram
models that combine the same information using simple interpolation.
Our own pilot experiments comparing maximum entropy models to similar
n-gram models were negative.

The most substantial gain in maximum entropy models comes from word
triggers.  In these models, a word such as ``school'' increases
its own probability, as well as the probability of similar words, such
as ``teacher.''  \namecite{Rosenfeld:94a} gets approximately a 25\%
perplexity reduction by using word triggers, although the gain is
reduced to perhaps 7\%-15\% when combining with a model that already
contains a cache.  \namecite{Tillmann:96a} achieve about a 7\%
perplexity reduction when combining with a model that already has a
cache, and \namecite{Zhang:00a} reports an 11\% reduction.  

A recent variation of the maximum entropy approach is the whole
sentence maximum entropy approach \mycite{Rosenfeld:01a}.  In this
variation, the probability of whole sentences is predicted, instead of
the probabilities of individual words.  This allows features of the
entire sentence to be used, e.g. coherence \mycite{Cai:00a} or parsability, rather than
word level features.  However, training whole sentence maximum entropy
models is particularly complicated \mycite{Chen:99c}, requiring sampling methods such as
Monte Carlo Markov Chain techniques, and we personally do not think
that there are many important features of a sentence that cannot be
rephrased as features of individual words.}

\longversion{ Maximum entropy
models \mycite{Darroch:72a} have received a fair amount of attention
since their introduction for language modeling by
\namecite{Rosenfeld:94a}.  Maximum entropy models are models of the
form
$$
\Px{maxent}(w|w_1...w_{i-1}) = \frac{\exp(\sum_k \lambda_k f_k(w,w_1...w_{i-1}))}{z(w_1...w_{i-1})}
$$
where $z$ is a normalization function, simply set equal to
$$
\sum_w \exp(\sum_k \lambda_k f_k(w, w_1...w_{i-1}))
$$
The $\lambda_k$ are real numbers, learned by a learning algorithm such
as Generalized Iterative Scaling (GIS) \mycite{Darroch:72a}.  The
$f_k$ are arbitrary functions of their input, typically returning 0 or
1.  They can be used to capture n-grams, caches, and clusters, and
skipping models.  They can also be used for more sophisticated
techniques, such as triggers, described below.  The generality and
power of the $f_k$ are one of the major attractions of maximum entropy
models.  There are other attractions to maximum entropy models as
well.  Given training data $w_1$...$w_T$, it is possible to find
$\lambda$s such that for each every $k$,
$$
\sum_w \sum_{1 \leq i \leq T} \Px{maxent}(w|w_1...w_{i-1}) f_k(w, w_1...w_{i-1}) =
\sum_{1 \leq i \leq T} f_k(w_, w_1...w_{i-1})
$$
In other words, the number of times we expect $f_k$ to occur given the
model is the number of times we observe it in training data.  That is,
we can define arbitrary predicates over words $w$ and histories
$w_1...w_{i-1}$, and build a model such that all of these
predicates are true as often as we observed.  In addition, this model
maximizes entropy, in the sense that it is also as smooth as possible
(as close to the uniform distribution) while meeting these
constraints.  This is a quite seductive capability.

Furthermore, there have been two pieces of recent research that have
made us especially optimistic about the use of maximum entropy models.
The first is the the smoothing technique of \namecite{Chen:99b}, which
is the first maximum entropy smoothing technique that works as well
(or better than) Kneser-Ney smoothing.  The second is our own recent
speedup techniques \mycite{Goodman:01a}, which lead to up to a factor of
35, or more, speedup over traditional maximum entropy training
techniques, which can be very slow.

There are reasons both for optimism and pessimism with respect to
maximum entropy models.  On the one hand, maximum entropy models have
lead to some of the largest reported perplexity reductions.
\namecite{Rosenfeld:94a} reports up to a 39\% perplexity reduction
when using maximum entropy models, combined with caching and a
conventional trigram.  On the other hand, our own pilot experiments
with maximum entropy were negative, when we compared to comparable
interpolated n-gram models: we are not aware of any research in which
maximum entropy models yield a significant improvement over comparable
n-gram models.  Furthermore, maximum entropy models are extremely time
consuming to train, even with our speedup techniques, and also slow to
use during test situations.  Overall, maximum entropy techniques may
be too impractical for real-world use.

One special aspect of maximum entropy models worth mentioning is word
triggers \mycite{Rosenfeld:94a}.  Word triggers are the source of the
most substantial gain in maximum entropy models.  In trigger models, a
word such as ``school'' increases the its own probability, as well as
the probability of similar words, such as ``teacher.''
Rosenfeld gets approximately a 25\% perplexity
reduction by using word triggers, although the gain is reduced to
perhaps 7\%-15\% when combining with a model that already contains a
cache.  \namecite{Tillmann:96a} achieves about a 7\% perplexity
reduction when combined with a model that already has a cache, and
\namecite{Zhang:00a} reports an 11\% reduction.

}

\longversion{
\subsection{Whole Sentence Maximum Entropy Models}

A recent variation of the maximum entropy approach is the whole
sentence maximum entropy approach \mycite{Rosenfeld:01a}.  In this
variation, the probability of whole sentences is predicted, instead of
the probabilities of individual words.  This allows features of the
entire sentence to be used, e.g. coherence \mycite{Cai:00a} or parsability, rather than
word level features.  If $s$ represents an entire sentence, then a
whole sentence maximum entropy model is of the form 
$$P(s)=\frac{1}{Z} P_0(s) \exp \sum_k \lambda_k f_k(s)$$ where
$P_0(s)$ is some starting model, e.g. an n-gram model.  Notice that
the normalization factor $Z$ is in this case a constant, eliminating
the need to compute the normalization factor separately for each
context.  In fact, it is not necessary to compute it at all, for most
uses.  This is one of the main benefits of these models, according to
their proponents, although other techniques \mycite{Goodman:01a} also
reduce the burden of normalization.

There are several problems with the whole sentence approach.  First,
training whole sentence maximum entropy models is particularly
complicated \mycite{Chen:99c}, requiring in practice sampling methods such as Monte
Carlo Markov chain techniques.  Second, the benefits of a whole
sentence model may be small when divided over all the words.  Consider
a feature such as parsability: is the sentence parsable according to
some grammar.  Imagine that this feature contributes an entire bit of
information per sentence.  (Note that in practice, a grammar broad
coverage enough to parse almost all grammatical sentences is broad
coverage enough to parse many ungrammatical sentences as well,
reducing the information it provides.)  Now, in an average 20 word
sentence, 1 bit of information reduces entropy by only .05 bits per
word, a relatively small gain for a very complex feature.  Another problem
we see with these models is that most of their features can be
captured in other ways.  For instance, a commonly advocated feature
for whole sentence maximum entropy models is ``coherence'', the notion
that the words of a sentence should all be on similar topics.  But
other techniques, such as caching, triggering, and sentence mixture
models are all ways to improve the coherence of a sentence that do not
require the expense of a whole sentence approach.  Thus, we are
pessimistic about the long term potential of this approach.\footnote{Of
course, we are pessimistic about almost everything.}
}

\subsection{Latent Semantic Analysis}
\namecite{Bellegarda:00a} shows that techniques based on Latent
Semantic Analysis (LSA) are very promising.  LSA is similar to
Principle Components Analysis (PCA) and other dimensionality reduction
techniques, and seems to be a good way to reduce the data sparsity
that plagues language modeling.  The technique leads to significant
perplexity reductions (about 20\%) and word error rate reductions
(about 9\% relative) when compared to a Katz trigram on 42 million
words.  It would be interesting to formally compare these results to
conventional caching results, which exploit similar long term
information.  Bellegarda gets additional improvement over these
results by using clustering techniques based on LSA; the perplexity
reductions appear similar to the perplexity reductions from
conventional IBM-style clustering techniques.

\subsection{Neural Networks}
There has been some interesting recent work on using Neural Networks
for language modeling, by \namecite{Bengio:00a}.  In order to deal
with data sparsity, they first map each word to a vector of 30
continuous features, and then the probability of various outputs is
learned as a function of these continuous features.  The mapping is
learned by backpropagation in the same way as the other weights in the
network.  The best result is about a 25\% perplexity reduction over a
baseline deleted-interpolation style trigram.  \shortversion{We
performed experiments on the same data set as Bengio \myetal  We found
that their techniques appeared to outperform clustered models somewhat
(about 5\% lower perplexity) and we think they have a fair amount of
potential.  It remains to be seen how similar those techniques are to
normal clustering techniques.}

\longversionboring{We wanted to see how the neural network model
compared to standard clustering models.  We performed some experiments
on the same corpus, vocabulary, and splits into training, test, and
heldout data as used by Bengio \myetal and we are gratful to them for
supplying us with the relevant data.  First, we verified that their
baseline was correct; we got a perplexity of 350.6 using simple
interpolation, versus 348 for a deleted-interpolation style baseline,
which seems very reasonable.  On the other hand, a Kneser-Ney smoothed
trigram had a perplexity of 316.  The remaining experiments we did
were with Kneser-Ney smoothing.  We next tried an ibm-clustered 4-gram
model, which is in some ways similar to the neural network model with
three words of context.  The clustered model has perplexity 271,
compared to the neural network perplexity of 291.  However, the 291
number does not interpolate with a trigram, while our clustered model
does.  We also ran a 6-gram ibm-clustered model.  This model had a
perplexity of 271, which is about 5\% worse than the best neural
network model score of 258; the 258 model was interpolated with a
trigram, so this is a fair comparison.  Finally, we tried to build a
``pull-out-all-the-stops'' model.  This was not the very best model we
could build -- it did not include caching, for instance -- but it was
the best model we could build using the same inputs as the best neural
network, the previous 5 words.  Actually, we used only the previous 4
words, since we were not getting much benefit from the 5'th word back.
This model was of the form
\begin{eqnarray*}
\lefteqn{\Px{all-the-stops}(w|w_{n\!-\!4}w_{n\!-\!3}w_{n\!-\!2}w_{n\!-\!1}) = } \\
&& \lambda P(w|w_{n\!-\!4}w_{n\!-\!3}w_{n\!-\!2}w_{n\!-\!1}) + \mu
P(W|W_{n\!-\!4}W_{n\!-\!3}W_{n\!-\!2}W_{n\!-\!1})\times P(w|W) + \\
&& \alpha P(w|w_{n\!-\!4}w_{n\!-\!3}w_{n\!-\!2}W_{n\!-\!1}) + \beta P(w|w_{n\!-\!4}w_{n\!-\!3}W_{n\!-\!2}W_{n\!-\!1}) +\\
&& \gamma P(w|w_{n\!-\!4}W_{n\!-\!4}w_{n\!-\!3}W_{n\!-\!3}w_{n\!-\!2}W_{n\!-\!2}w_{n\!-\!1}W_{n\!-\!1})+\\
&& \epsilon P(W|w_{n\!-\!4}w_{n\!-\!3}w_{n\!-\!2}w_{n\!-\!1}) \times P(w|w_{n\!-\!4}w_{n\!-\!3}w_{n\!-\!2}w_{n\!-\!1}W) + \\
&& (1\!-\!\lambda \!-\! \mu \!-\! \alpha \!-\!\beta  \!-\! \gamma \!-\!\epsilon)
\begin{array}[t]{l}P(W|w_{n\!-\!4}W_{n\!-\!4}w_{n\!-\!3}W_{n\!-\!3}w_{n\!-\!2}W_{n\!-\!2}w_{n\!-\!1}W_{n\!-\!1})
\times \\ P(w|w_{n\!-\!4}W_{n\!-\!4}w_{n\!-\!3}W_{n\!-\!3}w_{n\!-\!2}W_{n\!-\!2}w_{n\!-\!1}W_{n\!-\!1}W) \end{array}
\end{eqnarray*}
and had a perplexity of 254.6, a meaningless 2\% better than the best
neural network.  The best neural network model relied on interpolation
with a trigram model, and used the trigram model exclusively for low
frequency events.  Since their trigram model had a relatively high
perplexity of 348, compared to a Kneser-Ney smoothed trigram's 316, we
assume that a significant perplexity reduction could be gotten simply
from using a Kneser-Ney smoothed trigram.  This means that the neural
network model results really are very good.

It would be very interesting to see how much the neural network model
has in common with traditional clustered models.  One simple
experiment would interpolate a neural network model with a clustered
model, to see how much of the gains were additive.  Given the
relative simplicity of the neural network used, and the very good
results, this seems like a very promising area of research.
}

\subsection{Structured Language Models} \label{sec:structured}
One of the most interesting and exciting new areas of language modeling
research has been structured language models (SLMs).  The first successful
structured language model was the work of \namecite{Chelba:98a}, and other
more recent models have been even more successful \mycite{Charniak:01a}.
The basic idea behind structured language models is that a properly
phrased statistical parser can be thought of as a generative model of
language.  Furthermore, statistical parsers can often take into
account longer distance dependencies, such as between a subject and
its direct or indirect objects.  These dependencies are likely to be
more useful than the previous two words, as captured by a trigram
model.  Chelba is able to achieve an 11\% perplexity reduction over a
baseline trigram model, while Charniak achieves an impressive 24\%
reduction.  \shortversion{We hypothesized that much of the
benefit of a structured language model might be redundant with other
techniques, such as skipping or clustering.  With the help of Chelba,
we performed experiments on his model.  
It turned out to be hard to answer, or even rigorously ask, the
question of how redundant one model was with another, but based on our
experiments, in which we compared how much entropy reduction we
got by combining models, versus how much we might expect, very roughly
half of the model appears to be in common with other techniques,
especially clustering.  Details are given in the extended version of
this paper.}

\longversionboring{ We hypothesized that much of the benefit of a structured
language model might be redundant with other techniques, such as
skipping, clustering, or 5-grams.  The question of how much
information one model has that another model already captures turns
out to be a difficult one.  While formulas for measuring the
conditional entropy of a word given two different models are well
known, they rely on computing joint probabilities.  If the two models
already are sparsely estimated, such as a conventional trigram and a
structured language model, then computing these joint probabilities is
hopeless.  So, we decided to use more approximate measures.  One
simple, practical technique is to simply try interpolating the two
models.  It seems, at first glance, that if the interpolation leads to
no gain, then the models must be capturing the same information, and
if the gains are additive over a common baseline, then the information
is independent.  Unfortunately, at least the first assumption is
incorrect.  In particular, when comparing the structured language
model to a Kneser-Ney smoothed trigram interpolated with a trigram
cache,  with which we assume the overlap in
information is minimal, we end up getting almost no gain from the
interpolation versus the cache model/Kneser-Ney trigram alone.  This
is simply because interpolation is such a crude way to combine
information (although we don't know any that are much better).  The
cache model is so much better than the structured language model that
the interpolation weight all goes to the cache model, and thus the
structured language model cannot help; we learn nothing except that
the cache model had a lower perplexity, which we already knew.

Our strategy then was a bit more complex, and a bit harder to
interpret.  First, we used versions of both systems with simple
interpolation for the smoothing.  This was the only smoothing
technique that was already implemented in both the SLM program and our
toolkit.  This removed smoothing as a factor.  Next, we tried
comparing the SLM to a trigram with various cache sizes (although, of
course, we never cached beyond document boundaries), and interpolating
that with the SLM.  We assumed that the SLM and the cache, of whatever
size, were capturing roughly orthogonal information.  This let us
figure out the amount of perplexity reduction we would expect if the
two models were uncorrelated.  For instance, the baseline SLM, not
interpolated with a trigram, had a perplexity of 167.5.  A fullibm
clustered model had a perplexity of 144.7; similarly, a trigram with
trigram cache with 160 words of context had a perplexity of 143.4 --
about the same as the clustered model.  The combination of the SLM
with the 160 word context cache had a perplexity of 130.7, a reduction
of 8.8\% over the cache alone.  When combined with the fullibm
clustered model, the perplexity was 137.0, a 5.3\% reduction over the
fullibm clustered model.  So, if the improvements were uncorrelated,
we would have assumed an 8.8\% reduction, and instead we only saw a
5.3\% reduction.  This is 60\% of the reduction we would have
expected.  Thus, we say that the overlap with a clustered
model is very roughly 40\%.

The following table shows the results of our various experiments.  The
``model'' column describes the model we interpolated with the SLM.
The first ``perplex'' column shows the perplexity of the model, and the
first ``reduction'' column shows the reduction from interpolating the
model with the SLM.  The second ``perplex'' column shows the
perplexity of the most similar cache model and the second ``reduction''
column shows the perplexity reduction from interpolating this cache
model with the SLM.  The final column, ``overlap'', shows the ration
between the first reduction and the second: the presumed overlap
between the model and the SLM.

\begin{center}
\begin{tabular}{lrrrrr}
      &   \multicolumn{2}{c}{model}     & \multicolumn{2}{c}{closest cache} & \\
model & perplex & reduction & perplex & reduction & overlap \\
fullibm & 144.7 & 5.3\% & 160 & 8.8\% & 40\% \\
Kneser-Ney  & 144.9 & 6.7\% & 143.4 & 8.8\% & 23\% \\
5-gram & 160.0 & 9.9\% & 162.6 & 10.5\% & 6\% \\
trigram skipping & 157.0 & 8.5\% & 155.0 & 9.8\% & 13\% \\
special cluster 1 & 141.4 & 4.5\% & 141.8 & 8.5\% & 45\% \\
special cluster 2 & 143.6 & 5.0\% & 143.4 & 8.8\% & 43\% \\
\end{tabular}
\end{center}

The trigram skipping model was a model with all pairs, through the
5-gram level.  The special cluster 1 and special cluster 2 models were
clustered skipping models designed to capture contexts similar to what
we assumed the structured language model was doing.

We would have liked to have looked at combinations, such as Kneser-Ney
smoothing with fullibm clustering, but the best cache model we tested
had a perplexity of 141.8, while the Kneser-Ney clustered model had a
much lower perplexity.

It is difficult to draw any too strong conclusions from these results.
One odd result is the large overlap with Kneser-Ney smoothing -- 23\%.
We suspect that somehow the breakdown of the language model into
individual components in the structured LM has a smoothing effect.
Or, perhaps our entire evaluation is flawed.

We also looked at the correlation of individual word probabilities.
We examined for each model for each word the difference from the
probability of the baseline trigram model.  We then measured the
correlation of the differences.  These results were similar to the
other results.

It would be interesting to perform these same kind of experiments with
other structured language models.  Unfortunately, the current best,
Charniak's, does not predict words left to right.  Therefore, Charniak
can only interpolate at the sentence level.  Sentence level
interpolation would make these experiments even harder to interpret.

Overall, we are reasonably confident that some large portion of the
increase from the SLM -- on the order of 40\% or more -- comes from
information similar to clustering.  This is not surprising, given that
the SLM has explicit models for part of speech tags and non-terminals,
objects that behave similarly to word clusters.  We performed these
experiments with Ciprian Chelba, and reached opposite conclusions: we
concluded that the glass was half empty, while Chelba concluded that
the glass was half full.
}

\section{Conclusion}

\label{sec:conclusion}

\subsection{Previous combinations}

There has been relatively little previous research that attempted to
combine more than two techniques, and even most of the previous
research combining two techniques was not particularly systematic.
Furthermore, one of the two techniques typically combined was a
cache-based language model.  Since the cache model is simply linearly
interpolated with another model, there is not much room for
interaction.

A few previous papers do merit mentioning.  The most recent is that of
\namecite{Martin:99a}.  They combined interpolated Kneser-Ney
smoothing, classes, word-phrases, and skipping.  Unfortunately, they
do not compare to the same baseline we use, but instead compare to
what they call interpolated linear discounting, a poor baseline.
However, their improvement over Interpolated Kneser-Ney is also given;
they achieve about 14\% perplexity reduction over this baseline,
versus our 34\% over the same baseline.  Their improvement from
clustering is comparable to ours, as is their improvement from
skipping models; their improvement from word-phrases, which we do not
use, is small (about 3\%); thus, the difference in results is due
mainly to our implementation of additional techniques: caching,
5-grams, and sentence-mixture models.  Their word error rate reduction
over Interpolated Kneser-Ney is 6\%, while ours is 7.3\%.  We assume
that the reason our word error rate reduction is not proportional to
our perplexity reduction is two-fold.  First, 4\% of our perplexity
reduction came from caching, which we did not use in our word
error-rate results.  Second, they were able to integrate their simpler
model directly into a recognizer, while we needed to rescore n-best
lists, reducing the number of errors we could correct.

Another piece of work well worth mentioning is that of
\namecite{Rosenfeld:94a}.  In that work, a large number of techniques
are combined, using the maximum entropy framework and interpolation.
Many of the techniques are tested at multiple training data sizes.
The best system interpolates a Katz-smoothed trigram with a cache and
a maximum entropy system.  The maximum entropy system incorporates
simple skipping techniques and triggering.
The best
system has perplexity reductions of 32\% to 39\% on data similar to
ours.  Rosenfeld gets approximately 25\% reduction from word triggers
alone (p. 45), a technique we do not use.  Overall, Rosenfeld's
results are excellent, and would quite possibly exceed ours if more
modern techniques had been used, such as Kneser-Ney smoothing the
trigram model (which is interpolated in), using smaller cutoffs made
possible by faster machines and newer training techniques, or
smoothing the maximum entropy model with newer
techniques.\longversion{ Rosenfeld experiments with some simple class-based
techniques without success; we assume that more modern classing
technology could also be used.}  Rosenfeld achieves about a 10\% word
error rate reduction\longversion{when using unsupervised adaptation
(the same way we adapted.  He achieves 13\% assuming supervsion --
users correct mistakes immediately.)}.

There is surprisingly little other work combining more than two
techniques.  The only other noteworthy research we are aware of is
that of \namecite{Weng:97a}, who performed experiments combining
multiple corpora, fourgrams, and a class-based approach similar to
sentence-mixture models.  Combining all of these techniques leads to
an 18\% perplexity reduction from a Hub4-only language model.  This
model was trained and tested on a different text genre than our
models, and so no comparison to our work can be made.

\longversion{
\subsection{All hope abandon, ye who enter here}
\mymarginpar{Begin long rambling cynical diatribe -- no results or particularly novel ideas.  Grad students thinking about research in language modeling {\em should} read this section}
In this section,\footnote{Given that this extended version is a technical report, rather than a
full paper, we can rant and rave without fear of editorial review,
willy nilly titling sections, and scaring innocent graduate students.
We can express our own bitterness and frustration and vent it on
unsuspecting readers.  Those with a natural predisposition to
pessismism should skip this section, or risk the consequences.}
we argue that meaningful, practical reductions in
word error rate are hopeless.  We point out that trigrams remain the
de facto standard not because we don't know how to beat them, but
because no improvements justify the cost.  We claim with little
evidence that entropy is a more meaningful measure of progress than
perplexity, and that entropy improvements are small.  We conclude that
most language modeling research, including ours, by comparing to a
straw man baseline and ignoring the cost of implementations, presents
the illusion of progress without the substance.  We go on to describe
what, if any, language modeling research is worth the effort.

\subsubsection{Practical word error rate reductions are hopeless}

Most language modeling improvements require significantly more space
than the trigram baseline compared to, and also typically require
significantly more time.  Most language modeling papers contain a
paragraph such as ``Trigram models are state-of-the-art.  Trigram
models are obviously braindamaged.  We did something slightly less
braindamaged.  Our model has much lower perplexity and somewhat lower
word error rates than a trigram.''  However, what the papers don't
say, but what almost universally applies, is that the resulting model
requires much more space and/or time than a simple trigram model.
Trigram models are state-of-the-art in the sense that they are used in
most commercial speech recognizers and as at least the first pass of
most research recognizers.  This is for the following two reasons:
they are fairly space efficient, and all sorts of tricks can be used
to integrate the trigram storage with the search.  For instance, in
our recognizers at Microsoft, we use a storage method in which the
trigrams are stored as phonetic trees \mycite{Alleva:96a}.  Many more
complex language modeling techniques, especially clustering
techniques, are incompatible with this type of storage, and thus much
less efficient in a practical, optimized recognizer.

Consider in practice each of the techniques we implemented.  Sentence
mixture models require substantially more storage than corresponding
global models, since different models have to be stored for each
mixture, and a global model still needs to be stored, to interpolate
with the mixture-specific models.  Furthermore, if integrated into the
first pass of a recognizer, a different path has to be explored for
each mixture type, significantly slowing the recognizer.  Caching has
problems we have already discussed: while it does not significantly
increase the storage for the recognizer, it requires users to correct
all their mistakes after each sentence; otherwise, mistakes are
``locked in.''  Clustered models of the sort we have described here
substantially increase storage, and interact very poorly with the
kinds of prefix trees used for search.  5-grams require substantially
more space than trigrams and also slow the search.  Kneser-Ney
smoothing leads to improvements in theory, but in practice, most
language models are built with high count cutoffs, to conserve space,
and speed the search; with high count cutoffs, smoothing doesn't
matter.  Skipping models require both a more complex search and more
space and lead to marginal improvements.  In short, none of the
improvements we studied is likely to make a difference in a real
system in the forseeable future, with the exception of caching, which
is typically available, but often turned off by default, because users
don't like to or can't be trained to consistently correct recognizer
errors.

Our results have surveyed and reimplemented the majority of promising
language modeling techniques introduced since \namecite{Katz:87a}.  The
improvement in perplexity is one of the largest ever reported, but the
improvement in word error rate is relatively small.  This partially
reflects our own emphasis on perplexity rather than word error rate,
and flaws in the experiments.  Perhaps rescoring lattices rather than
n-best lists would have lead to larger improvements.  Perhaps
integrating the language modeling into the search would have helped.
Perhaps trying to optimize parameters and techniques for word error
rate reduction, rather than perplexity reduction would have worked
better.  But, despite all these flaws, we think our results are
reasonably representative.  Others who have tried their best to get
language modeling improvements
have had mixed results. For instance, \namecite{Martin:99a} gets a 12\%
relative word error rate reduction from a fair baseline (larger from
an easier baseline), slightly better than ours.  But the cost of most
of their improvements is high -- a quadrupling in size of the language
model and 32 times the CPU time.  Larger improvements could probably
be gotten more easily and cheaply simply by increasing the number of
mixture components or loosening search thresholds in the acoustic
model.  In fact, simply by using more language model training data,
comparable improvements could probably be had.\footnote{These
criticisms apply not just to the work of Martin \myetal but to our own
as well.  We use Martin's system as an example because quirks of our
tool (see Appendix \ref{sec:appendiximplementationnotes}) make it hard for us
to measure the size of the models we used and our acoustic modeling
was done as offline n-best rescoring, rather than integrated with the
search.}

\subsubsection{Perplexity is the wrong measure}

In this subsubsection, we claim with little evidence that word error
rate correlates better with entropy than with perplexity.  Since
relative entropy reductions are much smaller than relative perplexity
reductions, it is difficult to get useful word error rate reductions.

We considered doing the following experiment: incrementally add
portions of the test data to our training set, and use this to compute
perplexity/entropy and word error rates.  However, we noticed that
actually conducting the experiment is unneccessary, since (in a
slightly idealized world) we can predict the results.  If we add in
x\% of the test data, then we will have perfect information for that
x\%.  With a sufficiently powerful n-gram model (say a 1000-gram), and
no search errors, our speech recognizer will get that portion exactly
correct, leading to an x\% relative reduction in word error rate.
Similarly, it will require essentially 0 bits to predict the portion
of the training data we have already seen, leading to an x\% reduction
in entropy.  This theoretical analysis leads us to hypothesize a
linear relationship between entropy and word error rate.

\begin{figure}
$$\psfig{figure=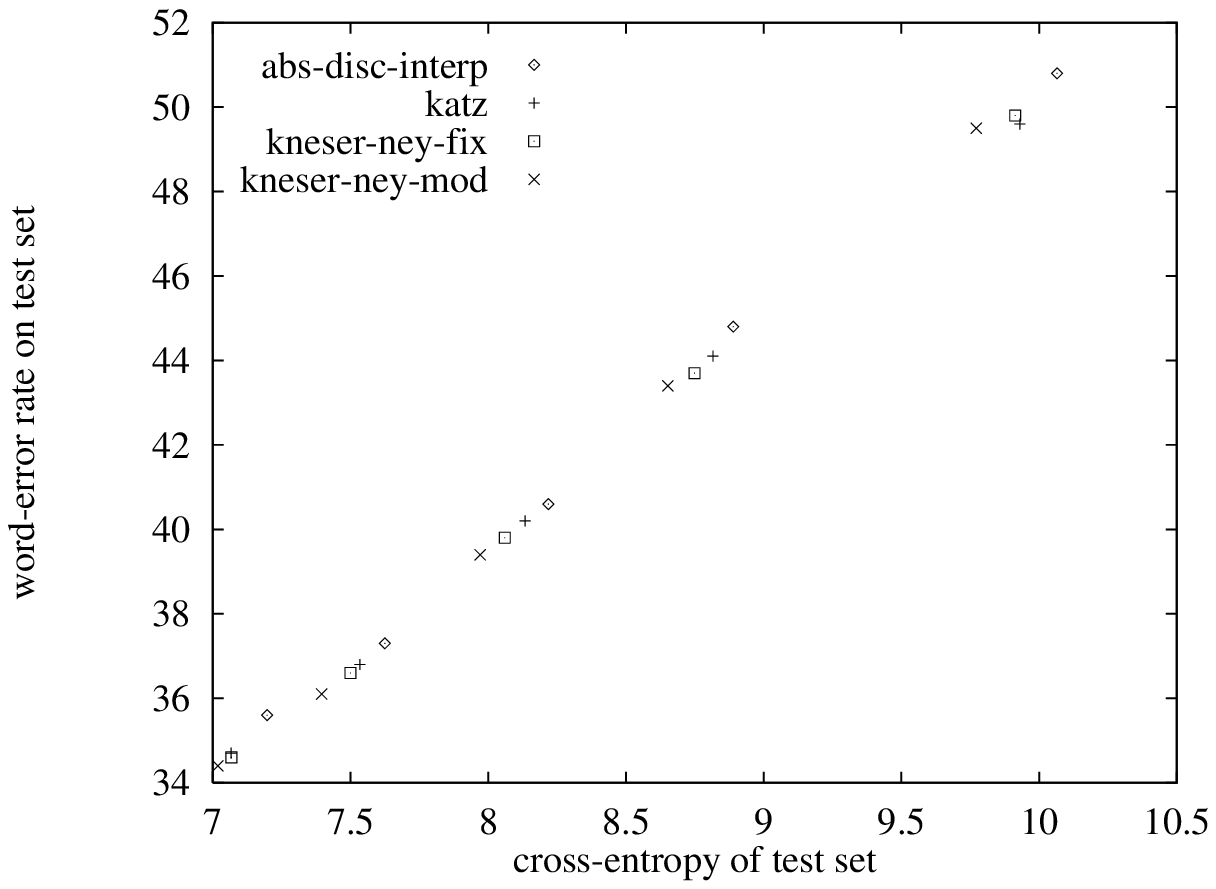,width=5.5in,angle=0}$$
\caption{Word error rate versus entropy}
\label{fig:entropyerror}
\end{figure}

Of course, one could perform more realistic experiments to see whether
perplexity or entropy corellates better with word error rate.  Our own
recent experiments do not shed light on this question -- the
correlation between word error rate and entropy, and the correlation
between word error rate and perplexity are the same, over the
relatively narrow range of perplexities we tested.  However, consider
previous work by \namecite{Chen:98a}.  In Figure
\ref{fig:entropyerror}, we show experiments done on the correlation
between word error rate and entropy.  Obviously, the relationship is
very strong and appears linear over a large range, including a roughly
0 intercept, just as our analysis predicted.\footnote{We prefer to make all of our predictions after seeing the data; this vastly increases their accuracy.}

Consider also other domains, such as text compression.  Here,
trivially, the relationship between entropy and the objective
function, number of bits required to represent the data, is again
linear.

These are not strong arguments, especially in light of the many
researchers who have gotten results where entropy and word error rate
do not correlate.  Still, they are worrisome.  If true, they imply
that we must get very large reductions in perplexity to get meaningful
word error rate reductions.  For instance, a 10\% perplexity
reduction, from say 100 to 90, corresponds to a reduction from 6.64
bits to 6.49 bits, which is only a 2\% entropy reduction.  What
appears as a meaningful perplexity reduction is a trivial entropy
reduction.  We fear that this means that commercially useful
reductions in perplexity are unlikely.

\subsubsection{Progress is Illusory}

A common and related pitfall in language modeling research is the
straw man baseline.  Most language modeling papers point out a problem
with trigram models, typically that they cannot capture long distance
dependencies, and then proceed to give some way to model longer
contexts.  These papers almost never even compare to a well-smoothed
5-gram, which would be a simple, obvious way to make a trigram capture
longer contexts, never mind comparing to a trigram with a cache, a
technique known for over 10 years, or to more recent models, such as
sentence mixture models.  The result is yet another way to beat
trigrams, assuming we don't care about speed or space.  Whether these
new techniques are better than the previous ones, or can be combined
with them for larger improvements, or offer a better speed or space
tradeoff than other techniques, is only very rarely explored.  When a
cache comparison is used, it is typically in the context of combining
the cache with their own model, rather than with the baseline trigram.
Perplexity results in the introduction, conclusion, and abstract will
usually compare to their technique with a cache, versus a trigram
without one.  Sometimes the cache does not look like a cache.  It
might, for instance, be a trigger model if maximum entropy is being
done.  But since the most useful triggers are just self triggers,
which are basically cache features, it is just a cache after all.

Poor experimental technique and misleading papers are by no means
unique to language modeling.  But the prevalence of the trigram
baseline in practice, allowing researchers to call it
``state-of-the-art'' with a straight face, makes it particularly easy
to give the illusion of progress.\footnote{Wait a minute, doesn't this
paper do all those same bad things? No.  Our paper doesn't claim that
trigrams are state of the art (although it does, misleadingly, call
them a fair baseline), and it fairly compares all techniques.
\footnotemark}\footnotetext{\tiny Sounds like weasling to me.
Hypocrite!}

\subsubsection{So, what now?}

Given the previous sections, should anyone do any language modeling
research at all?  Maybe.  There are four areas where I see some hope:
language model compression, language model adapatation, new
applications of language modeling, and basic research.

Language model compression or pruning is an area that has received
relatively little research, although good work has been done by
\namecite{Kneser:96a}, \namecite{Seymore:96a}, \namecite{Stolcke:98a},
\namecite{Siu:00a}, and by us \mycite{Goodman:00a}.  Care is needed
here.  For instance, the techniques of Seymore \myetal and Stolcke can
be used with almost any recognizer, but our own techniques use
clustering, and interact poorly with the search in some speech
recognizers.  Still, this seems like an area where progress can be
made.  Similarly, there has been no comprehensive work showing the
space/perplexity tradeoffs of all of the techniques we examined in
this paper, nor has there been any work on pruning interpolated
models, such as skipping models or sentence mixture models.

Another area for language modeling research is language model
adaptation.  A very common product scenario involves only a very
small amount of in-domain training data, and lots of out-of-domain
data.  There has been a moderate amount of research in this area.
Much of the research does not compare to simply interpolating together
the in-domain and out-of-domain language models, which in our
experiments works quite well.  The best research in this area is
probably that of \namecite{Iyer:97a}.  We suspect that better
techniques could be found, but our own attempts have failed.

Language models work very well and are well understood, and can be
applied to many domains
\mycite{Church:88a,Brown:90b,Hull:92a,Kernighan:90a,Srihari:92a}.
Rather than trying to improve them, trying to use them in new ways can
be fruitful.  Almost any machine learning problem could potentially
use language modeling techniques as a solution, and identifying new
areas where language models work well is likely to be as useful as
trying to do basic research.

There are many unfinished areas of language modeling basic research.
None of these will have a huge practical impact, but they will help
advance the field.  One area is a continuous version of Kneser-Ney
smoothing.  Interpolated Kneser-Ney smoothing is wonderful.  No matter
what kind of model we have used, it has worked better with Kneser-Ney
smoothing than with any other smoothing technique we have tried.  But
Kneser-Ney smoothing is limited to discrete distributions; it cannot
handle fractional counts.  Fractional counts are very common with, for
instance, Expectation Maximization (EM) algorithms.  This means that
we do not currently know how to do smoothing for distributions
learned through EM, such as most instances of Hidden Markov Models or
of Probabilistic Context-Free Grammars.  A continuous version of
Kneser-Ney could be used in all of these areas.  A related area that
needs more research is soft-clustering versus hard clustering.  Hard
clustering assigns each word to a single cluster, while soft
clustering allows the same word to be placed in different clusters.
There has been essentially no work comparing hard clustering to soft
clustering, but several soft-style techniques, including the work of
\namecite{Bengio:00a} and of \namecite{Bellegarda:00a} have had good
success, hinting that these techniques may be more effective.  One
reason we have not tried soft clustering is because we do not know how
to properly smooth it: soft clustered models have fractional counts
and would work best with a continuous version of Kneser-Ney smoothing.

\subsubsection{Some hope is left}
To summarize, language modeling is a very difficult area, but not one
that is completely hopeless.  Basic research is still possible, and
there continue to be new, if not practical, then certainly interesting
language modeling results.  There also appear to be a few areas in
which useful language modeling research is promising.  But language
modeling, like most research, but perhaps more so, is not an area for
the faint of heart or easily depressed.
\mymarginpar{End long rambling cynical diatribe}
}

\subsection{Discussion}

We believe our results -- a 50\% perplexity reduction on a very small data
set, and a 41\% reduction on a large one (38\% for data without
punctuation) -- are the best ever reported for language modeling, as
measured by improvement from a fair baseline, a Katz smoothed trigram
model with no count cutoffs.  We also systematically explored
smoothing, higher order n-grams, skipping, sentence mixture models,
caching, and clustering.

Our most important result is perhaps the superiority of Interpolated
Kneser-Ney smoothing in every situation we have examined.  We
previously showed \mycite{Chen:98a} that Kneser-Ney smoothing is always
the best technique across training data sizes, corpora types, and
n-gram order.  We have now shown that it is also the best across
clustering techniques, and that it is one of the most important
factors in building a high performance language model, especially one
using 5-grams.

We have carefully examined higher-order n-grams, showing that
performance improvements plateau at about the 5-gram level, and we
have given the first results at the 20-gram level, showing that there
is no improvement to be gained past 7 grams.

We have systematically examined skipping techniques.  We examined
trigram-like models, and found that using pairs through to the 5-gram
level captures almost all of the benefit.  We also performed 
experiments on 5-gram skipping models, finding a combination of 3
contexts that captures most of the benefit.

We carefully explored sentence mixture models, showing that
much more improvement can be had than was previously expected by
increasing the number of mixtures.  In our experiments, increasing the
number of sentence types to 64 allows nearly twice the improvement over a
small number of types.

Our caching results show that caching is by far the most useful
technique for perplexity reduction at small and medium training data
sizes.  They also show that a trigram cache can lead to almost twice
the entropy reduction of a unigram cache.

Next, we systematically explored clustering, trying 9 different
techniques, finding a new clustering technique, fullibmpredict, that
is a bit better than standard ibm clustering, and examining the
limits of improvements from clustering.  \longversion{We also showed
that clustering performance may depend on smoothing in some cases.}


Our word-error rate reduction of 8.9\% from combining all techniques
except caching is also very good.  

Finally, we put all the techniques together, leading to a 38\%-50\%
reduction in perplexity, depending on training data size.  The results
compare favorably to other recently reported combination results
\mycite{Martin:99a}, where, essentially using a subset of these
techniques, from a comparable baseline (absolute discounting), the
perplexity reduction is half as much.  Our results show that smoothing
can be the most important factor in language modeling, and its
interaction with other techniques cannot be ignored.

In some ways, our results are a bit discouraging.  The overall model
we built is so complex, slow and large that it would be completely
impractical for a product system.  Despite this size and complexity,
our word error rate improvements are modest.  To us, this implies that
the potential for practical benefit to speech recognizers from
language model research is limited.  On the other hand, language
modeling is useful for many fields beyond speech recognition, and is
an interesting test bed for machine learning techniques in general.

Furthermore, parts of our results are very encouraging.  First, they
show that progress in language modeling continues to be made.  For
instance, one important technique in our system, sentence mixture
models, is only a few years old, and, as we showed, its potential has
only been partially tapped.  Similarly, the combination of so many
techniques is also novel.  Furthermore, our results show that the
improvements from these different techniques are roughly additive: one
might expect an improvement of .9 bits for the largest training size
based on simply adding up the results of Figure \ref{fig:combinekneser}, and instead
the total is about .8 bits -- very similar.  This means that further
incremental improvements may also lead to improvements in the best
models, rather than simply overlapping or being redundant.

As we noted in Section \ref{sec:othertechniques}, there are many other
promising language modeling techniques currently being pursued, such
as maximum entropy models, neural networks, latent semantic analysis,
and structured language models.  Figuring out how to combine these
techniques with the ones we have already implemented should lead to
even larger gains, but also yet more complex models.

\section*{Acknowledgements}

I would like to thank the entire Microsoft Speech.Net Research Team
for their help, especially Milind Mahajan, X. D. Huang, Alex Acero,
Ciprian Chelba, as well as Jianfeng Gao.  I would also like to thank
the anonymous reviewers, Sarah Schwarm, Roland Kuhn, Eric Brill,
Hisami Suzuki, and Shaojun Wang for their comments on drafts of this
paper.  I would like to especially thank Stanley Chen for useful
discussions; in addition, small amounts of text and code used for this
implementation and paper irrespectively were originally coauthored
with Stanley Chen.

\shortversion{\bibliographystyle{plainnat}}
\longversion{\bibliographystyle{acl}}

\bibliography{master}

%

\longversion{

\clearpage

\appendix

\section{Kneser-Ney Smoothing}

In this section, we first briefly justify interpolated Kneser-Ney
smoothing, giving a proof that helps explain why preserving marginals
is useful.  Then, we give some implementation tips for Kneser-Ney
smoothing.  

\subsection{Proof}

First, we give a short theoretical argument for why Kneser-Ney
smoothing.  Most of our argument is actually given by
\namecite{Chen:99a}, derived from work done by \namecite{Kneser:95a}.
We prove one additional fact that was previously an assumption,
somewhat strengthening our argument.  In particular, we will show that
any optimal smoothing algorithm must preserve known marginal
distributions.

First, we mention that it is impossible to prove that any smoothing
technique is optimal, without making at least a few assumptions.  For
instance, work by \namecite{MacKay:95a} proves the optimality of a
particular, simple interpolated model, subject to the assumption of a
Dirichlet prior.  Here, we prove the optimality of Kneser-Ney
smoothing, based on other assumptions.  We note however, that the
assumption of a Dirichlet prior is probably, empirically, a bad one.
The Dirichlet prior is a bit odd, and, at least in our work on
language modeling, does not seem to correspond well to reality.  The
assumptions we make are empirically based, rather than based on
mathematical convenience.  We note also that we are not aware of any
case in which another smoothing method outperforms Kneser-Ney
smoothing, which is another excellent piece of evidence.

The argument for Kneser-Ney smoothing is as follows.  First, absolute
discounting empirically works reasonably well, and approximates the
true discounts (which can be measured on real data.)  Second,
interpolated techniques fairly consistently outperform backoff
techniques, especially with absolute-discounting style smoothing.
These are empirical facts based on many experiments \mycite{Chen:99a}.
Next, \namecite{Chen:99a} and \namecite{Kneser:95a} make an assumption
that preserving marginal distributions is good.  It is this assumption
that we will prove below.  Finally, \namecite{Chen:99a} prove
(following an argument of \namecite{Kneser:95a}) that the smoothing
technique that uses absolute discounting, interpolation, and preserves
marginals is Interpolated Kneser-Ney smoothing.  Thus, the
contribution of this section is to prove what was previously an
assumption, strengthening the argument for Interpolated Kneser-Ney
smoothing to one that relies on empirical facts and proofs derived
from those facts, with only a very small amount of hand waving.

We previously assumed that smoothing techniques that preserve marginal
distributions (e.g. smoothing the bigram such that the unigram
distribution is preserved) is a good thing.  In this section, we prove
that any modeling technique that does not preserve known marginal
distributions can be improved by preserving those distributions.
Admittedly, when we smooth a Kneser-Ney bigram while preserving the
observed unigram distribution, the observed unigram distribution is
not the true unigram distribution; it is simply a good approximation
to the true unigram.\footnote{In fact, in practice, we smooth the
unigram distribution with the uniform distribution because of its
divergence from the true distribution.}\footnote{This is the very
small amount of hand waving.  Whether or not one agrees with this
``proof'', it is certainly an improvement over the previous argument
which was simply ``Preserving the marginal distribution is good.''}

The proof is as follows.  Let $p$ represent an estimated distribution,
and let $P$ represent the true probability.  Consider an estimated bigram
distribution of the form $p(z|y)$ such that for some $z'$, $\sum_{y}
p(z'|y) P(y) \neq P(z')$.  Then, we show that the entropy with respect
to $P$ is reduced by using a new distribution, with a multiplier $\lambda$.  We simply take the old distribution, multiply the probability of $p(z'|y)$ by $\lambda$, and renormalize:
$$
p'(z|y) = \left\{ \begin{array}{ll}
\frac{p(z|y)}{1+(\lambda-1)p(z'|y)} & \mbox{if $z \neq z'$} \\
\frac{\lambda p(z'|y)}{1+(\lambda-1)p(z'|y)} & \mbox{if $z = z'$} 
\end{array} \right.
$$
It will turn out that the optimal value for $\lambda$ is the one such
that $\sum_{y} p'(z'|y) P(y) = P(z')$.

The proof is as follows: consider the entropy of $p'$ with respect to
$P$.  It
will be convenient to measure entropy in nats instead of bits -- that
is, using $e$ for the logarithm, instead of $2$.  (1 nat =
$\frac{1}{\ln 2}$ bits).  The entropy in nats is
$$
\sum_{y,z} - P(y,z) \ln p'(z|y) 
$$
Consider the derivative of the entropy with respect to $\lambda$.  
\begin{eqnarray*} 
\lefteqn {\frac{d}{d\lambda}  \sum_{y,z} - P(y,z) \ln p'(z|y) } \\
&=& \frac{d}{d\lambda}  \sum_{y,z} - P(y,z) \ln \left\{ \begin{array}{ll}
\frac{p(z|y)}{1+(\lambda-1)p(z'|y)} & \mbox{if $z \neq z'$} \\
\frac{\lambda p(z'|y)}{1+(\lambda-1)p(z'|y)} & \mbox{if $z = z'$} 
\end{array} \right. \\
&=&  \sum_{y,z} P(y,z) - \frac{d}{d\lambda} \ln   \left\{ \begin{array}{ll}
\frac{p(z|y)}{1+(\lambda-1)p(z'|y)} & \mbox{if $z \neq z'$} \\
\frac{\lambda p(z'|y)}{1+(\lambda-1)p(z'|y)} & \mbox{if $z = z'$} 
\end{array} \right. \\
&=&  \sum_{y,z}  \frac{-P(y,z)}{ \left\{ \begin{array}{ll}
\frac{p(z|y)}{1+(\lambda-1)p(z'|y)} & \mbox{if $z \neq z'$} \\
\frac{\lambda p(z'|y)}{1+(\lambda-1)p(z'|y)} & \mbox{if $z = z'$} 
\end{array} \right.} 
\frac{d}{d\lambda}
\left\{ \begin{array}{ll}
\frac{p(z|y)}{1+(\lambda-1)p(z'|y)} & \mbox{if $z \neq z'$} \\
\frac{\lambda p(z'|y)}{1+(\lambda-1)p(z'|y)} & \mbox{if $z = z'$} 
\end{array} \right.
 \\
&=&  \sum_{y,z}  \frac{-P(y,z)}{ \left\{ \begin{array}{ll}
\frac{p(z|y)}{1+(\lambda-1)p(z'|y)} & \mbox{if $z \neq z'$} \\
\frac{\lambda p(z'|y)}{1+(\lambda-1)p(z'|y)} & \mbox{if $z = z'$} 
\end{array} \right.} 
\left\{ \begin{array}{ll}
-\frac{p(z|y) \frac{d}{d\lambda} (1+(\lambda-1)p(z'|y))}{(1+(\lambda-1)p(z'|y))^2} & \mbox{if $z \neq z'$} \\
\frac{(1+(\lambda-1)p(z'|y)) \frac{d}{d\lambda} \lambda p(z'|y) - \lambda p(z'|y) \frac{d}{d\lambda} (1+(\lambda-1)p(z'|y))}{(1+(\lambda-1)p(z'|y))^2} & \mbox{if $z = z'$} 
\end{array} \right.
 \\
&=&  \sum_{y,z}\frac{- P(y,z) }{ \left\{ \begin{array}{ll}
\frac{p(z|y)}{1+(\lambda-1)p(z'|y)} & \mbox{if $z \neq z'$} \\
\frac{\lambda p(z'|y)}{1+(\lambda-1)p(z'|y)} & \mbox{if $z = z'$} 
\end{array} \right.} 
\left\{ \begin{array}{ll}
-\frac{p(z|y) p(z'|y)}{(1+(\lambda-1)p(z'|y))^2} & \mbox{if $z \neq z'$} \\
\frac{(1+(\lambda-1)p(z'|y)) p(z'|y) - \lambda p(z'|y) p(z'|y)}{(1+(\lambda-1)p(z'|y))^2} & \mbox{if $z = z'$} 
\end{array} \right.
 \\
&=&  \sum_{y,z}\frac{- P(y,z) }{ \left\{ \begin{array}{ll}
\frac{p(z|y)}{1+(\lambda-1)p(z'|y)} & \mbox{if $z \neq z'$} \\
\frac{\lambda p(z'|y)}{1+(\lambda-1)p(z'|y)} & \mbox{if $z = z'$} 
\end{array} \right.} 
\left\{ \begin{array}{ll}
-\frac{p(z|y) p(z'|y)}{(1+(\lambda-1)p(z'|y))^2} & \mbox{if $z \neq z'$} \\
\frac{(1-p(z'|y)) p(z'|y)}{(1+(\lambda-1)p(z'|y))^2} & \mbox{if $z = z'$} 
\end{array} \right.
 \\
&=&  \sum_{y,z}- P(y,z) 
\left\{ \begin{array}{ll}
-\frac{ p(z'|y)}{1+(\lambda-1)p(z'|y)} & \mbox{if $z \neq z'$} \\
\frac{1-p(z'|y) }{\lambda(1+(\lambda-1)p(z'|y))} & \mbox{if $z = z'$} 
\end{array} \right.
 \\
\end{eqnarray*}
Now, if $\lambda=1$, this reduces to
\begin{eqnarray*}
\sum_{y,z}- P(y,z) 
\left\{ \begin{array}{ll}
-p(z'|y) & \mbox{if $z \neq z'$} \\
1-p(z'|y)  & \mbox{if $z = z'$} 
\end{array} \right.
& = & -P(z') + \sum_y P(y) \sum_z P(z|y) p(z'|y) \\
& = & -P(z') + \sum_y P(y) p(z'|y) \sum_z P(z|y) \\
& = & -P(z') + \sum_y P(y) p(z'|y)\\
& = & -P(z') + \sum_y P(y) p(z'|y)\\
\end{eqnarray*}
which will be zero only when the true marginal, $P(z')$ is equal to
the marginal of $p$.  If $\sum_y P(y) p(z'|y)  > P(z')$, then the derivative will be $>0$, meaning that we can decrease the entropy by decreasing $\lambda$, and if $\sum_y P(y) p(z'|y)  < P(z')$, then the derivative will be $<0$, meaning that we can decrease the entropy by increasing $\lambda$.  Thus, whenever $\sum_y P(y) p(z'|y)  \neq P(z')$, there will be some value of $\lambda$ other than 1 that leads to lower entropy.  

This means that if we have some smoothing algorithm that does not
preserve the known marginal distributions, we can modify the resulting
distribution by multiplying it in such a way that the marginal is
preserved.  Thus, any smoothing algorithm that does not preserve the
known marginals is suboptimal.  

(Notice that our argument was not really about smoothing, but just
about preserving known marginal distributions in general.  This
argument is the same one used to justify maximum entropy models: the
best model will be the one that preserves all the known marginals.
Thus, this proof is not really new; the only novel part is noticing
that this fact applies to smoothing algorithms, as well as to other
distributions.)

\subsection{Implementing Kneser-Ney Smoothing}

The formula for Kneser-Ney smoothing is a bit complex looking.
However, the algorithm for actually finding the counts for Kneser-Ney
smoothing is actually quite simple, requiring only trivial changes
from interpolated absolute discounting.  In Figure \ref{fig:abstrain}
we give the code for training an interpolated absolute discounting
model, and in Figure \ref{fig:kntrain} we give the corresponding code
for an interpolated Kneser-Ney model.  The lines that change are
marked with an asterisk (*) -- notice that the only change is in the
placement of two right curly braces (\}).  For completeness, Figure
\ref{fig:testboth} gives the code for testing either interpolated
absolute discounting or interpolated Kneser-Ney smoothing -- the code
is identical.

} 

\begin{figure}
\begin{verbatim}
# code for implementing interpolated absolute discounting
# usage: discount training test
# training data and test data are one word per line

$discount = shift; # must be between 0 and 1
$train = shift;
$test = shift;

$w2 = $w1 = "";

open TRAIN, $train;
while (<TRAIN>) {         # For each line in the training data
    $w0 = $_;                         
    $TD{$w2.$w1}++;       # increment Trigram Denominator
    if ($TN{$w2.$w1.$w0}++ == 0) { # increment Trigram Numerator
        $TZ{$w2.$w1}++;   # if needed, increment Trigram non-Zero counts
*   }                     # This curly brace will move for Kneser-Ney
    
    $BD{$w1}++;           # increment Bigram Denominator
    if ($BN{$w1.$w0}++==0){ # increment Bigram Numerator
        $BZ{$w1}++;       # if needed, increment Bigram non-Zero counts
*   }                     # This curly brace will move for Kneser-Ney
    
    $UD++;                # increment Unigram Denominator
    $UN{$w0}++;           # increment Unigram Numerator 
    
    $w2 = $w1;
    $w1 = $w0;
}
close TRAIN;
\end{verbatim}
\caption{Interpolated Absolute Discounting Perl code}
\label{fig:abstrain}
\end{figure}

\begin{figure}
\begin{verbatim}
# code for implementing interpolated Kneser-Ney
# usage: discount training test
# training data and test data are one word per line

$discount = shift; # must be between 0 and 1
$train = shift;
$test = shift;

$w2 = $w1 = "";

open TRAIN, $train;
while (<TRAIN>) {           # For each line in the training data
    $w0 = $_;                   
    $TD{$w2.$w1}++;         # increment Trigram Denominator
    if ($TN{$w2.$w1.$w0}++ == 0) { # increment Trigram Numerator
        $TZ{$w2.$w1}++;     # if needed, increment Trigram non-Zero counts
    
        $BD{$w1}++;         # increment Bigram Denominator
        if ($BN{$w1.$w0}++==0){ # increment Bigram Numerator
            $BZ{$w1}++;     # if needed, increment Bigram non-Zero counts
            
            $UD++;          # increment Unigram Denominator
            $UN{$w0}++;     # increment Unigram Numerator 
*       }                   # This curly brace moved for Kneser-Ney
*   }                       # This curly brace moved for Kneser-Ney
    
    $w2 = $w1;
    $w1 = $w0;
}
close TRAIN;
\end{verbatim}
\caption{Interpolated Kneser-Ney Perl Code}
\label{fig:kntrain}
\end{figure}

\begin{figure}
\begin{verbatim}
$w2 = $w1 = "";

open TEST, $test;
while (<TEST>) {              # For each line in the test data
    $w0 = $_;
    $unigram = $UN{$w0} / $UD;        # compute unigram probability
    if (defined($BD{$w1})) {  # non-zero bigram denominator 
        if (defined($BN{$w1.$w0})) { # non-zero bigram numerator
            $bigram = ($BN{$w1.$w0} - $discount) /
                      $BD{$w1};
        } else {
            $bigram = 0;
        }
        $bigram = $bigram +
                  $BZ{$w1}* $discount / $BD{$w1} * $unigram;

        # Now, add in the trigram probability if it exists
        if (defined($TD{$w2.$w1})) { # non-zero trigram denominator
            if (defined($TN{$w2.$w1.$w0})) { # non-zero trigram numerator
                $trigram = ($TN{$w2.$w1.$w0} - $discount) /
                           $TD{$w2.$w1};
            } else {
                $trigram = 0;
            }
            $trigram = $trigram +
                       $TZ{$w2.$w1} * $discount / $TD{$w2.$w1} * $bigram;
            $probability = $trigram;
        } else {
            $probability = $bigram;
        }
    } else {
        $probability = $unigram;
    }

    print "Probability:\n  $w0    $w2    $w1       = $probability\n";
    
    $w2 = $w1;
    $w1 = $_;
}
close TEST;
\end{verbatim}
\caption{Testing Code for Interpolated Absolute Discounting or Kneser-Ney}
\label{fig:testboth}
\end{figure}

\longversion{

\section {Implementation Notes}
\label{sec:appendiximplementationnotes}
While this paper makes a number of contributions, they are much less
in the originality of the ideas, and much more in the completeness of
the research: its thoroughness, size,
combination, and optimality. This research was in many areas more
thorough, than any other previous research: trying more variations on
ideas; able to exceed previous size boundaries, such as examining
20-grams; able to combine more ideas; and more believable, because
nearly all parameters were optimized jointly on heldout data.  These
contributions were made possible by a well designed tool, with many
interesting implementation details.  It is perhaps these
implementation details that were the truly novel part of the work.

In this section, we sketch some of the most useful implementation
tricks needed to perform this work.

\subsection{Store only what you need}
\label{sec:storeonly}
The single most important trick used here, part of which was
previously used by \namecite{Chen:99a}, was to store only those parameters
that are actually needed for the models.  In particular, the very
first action the program takes is to read through the heldout and test
data, and determine which counts will be needed to calculate the
heldout and test probabilities.  Then, during the training phase, only
the needed counts are recorded.  Note that the number of necessary
counts may be much smaller than the total number of counts in the test
data.  For instance, for a test instance $P(z|xy)$, we typically need
to store all training counts $C(xy\ast)$.  However, even this depends
on the particular model type.  For instance, for our baseline
smoothing, simple interpolation, it is only necessary to know $C(xy)$
and $C(xyz)$ (as well as the lower order counts).  This fact was
useful in our implementation of caching, which used the baseline
smoothing for efficiency.

For Kneser-Ney smoothing, things become more complex: for the trigram,
we need to temporarily store $C(xy\ast)$ so that we can compute $\vert
\{ v | C(xyv) > 0 \} \vert$.  Actually, we only need to store for each
xyv whether we have seen it or not.  Since we do not need to store the
actual count, we can save a fair amount of space.  Furthermore, once
we know $\vert \{ v | C(xyv) > 0 \} \vert$, we can discard the table
of which ones occurred.  Partially for this reason, when we calculate
the parameters of our models, we do it in steps.  For instance, for a
skipping model that interpolates two different kinds of trigrams, we
first find the needed parameters for the first model, then discard any
parts of the model we no longer need, then find the parameters for the
second model.  This results in a very significant savings in storage.

One problem with this ``store only what you need'' technique is that
it interacts poorly with Katz smoothing.  In particular, Katz
smoothing needs to know the discount parameters.  Recall that $n_r$
represents the number of n-grams that occur $r$ times, i.e.
$$
n_r = \vert \{ w_{i-n+1}...w_i | C(w_{i-n+1}...w_i) = r \} \vert
$$
and that the Good-Turing formula states that for any n-gram that
occurs $r$ times, we should pretend that it occurs $\disc(r)$
times where 
$$
\disc(r) = (r+1)\frac{n_{r+1}}{n_r}
$$
On its face, computing the $n_r$ requires knowing all of the counts.
However, typically, discounting is only done for counts that
occur 5 or fewer times.  In that case, we only need to know $n_1$,
$n_2$, ..., $n_6$.  In fact, however, we do not need to know $n_r$ and
$n_{r+1}$ but only their ratio.  This means that we can sample the
counts and estimate their relative frequencies.  By sampling the
counts, we do {\it not} mean sampling the data: we mean sampling the
counts.  In particular, we define a random function of a sequence of
words (a simple hash function); we store the count for a sequence
$w_1w_2...w_n$ if and only if ${\mbox{\it hash}}(w_1w_2...w_n) \bmod s
= 0$, where $s$ is picked in such a way that we get 25,000 to 50,000
counts.  This does not save any time, since we still need to consider
ever count $C(w_1w_2...w_n)$ but it saves a huge amount of space.  In
pilot experiments, the effect of this sampling on accuracy was
negligible.

One problem with the ``store only what you need'' technique is that it
requires reading the test data.  If there were a bug, then excellent
(even perfect) performance could be achieved.  It was therefore
important to check that no such bugs were introduced.  This can be
done in various ways.  For instance, we added an option to the program
that would compute the probability of all words, not just the words in
the test data.  We then verified that the sum of the probabilities of
all words was within roundoff error ($10^{-9}$) of 1.  We also
verified that the probability of the correct word in the special
score-everything mode was the same as the probability in the normal
mode.  Also, none of the individual results we report are
extraordinarily good; all are in the same range as typically reported
for these techniques.  The only extraordinarily good result comes from
combining all technique, and even here, the total improvement is less
than the total of the individual improvements, implying it is well
within the reasonable range.  Furthermore, when our code is used to
rescore n-best lists, it reads through the lists and figures out
which counts are needed to score them.  However, the
n-best lists have no indication of which is the correct output.  (A
separate, standard program scores the output of our program against
the correct transcription.)  Since we get good improvement on n-best
rescoring, and the improvement is consistent with the perplexity
reductions, there is further evidence that our implementation of
``store only what you need'' did not lead to any cheating.

\subsection{System architecture}

It turns out that language modeling is an ideal place to use object
oriented methods, and inheritance.  In particular, our language
modeling tool was based on a virtual base class called Model.  The
single most important member function of a model provides the
probability of a word or a cluster given a context.  Models have all
sorts of other useful functions; for instance, they can return the set
of parameters that can be optimized.  They can go through the test
data and determine which counts they will need.  They can go through
the training data and accumulate those counts.  

From this base class, we can construct all sorts of other subclasses.
In particular, we can construct models that store probabilities --
DistributionModels -- and models that contain other models --
ListModels.

\begin{figure}
$$\psfig{figure=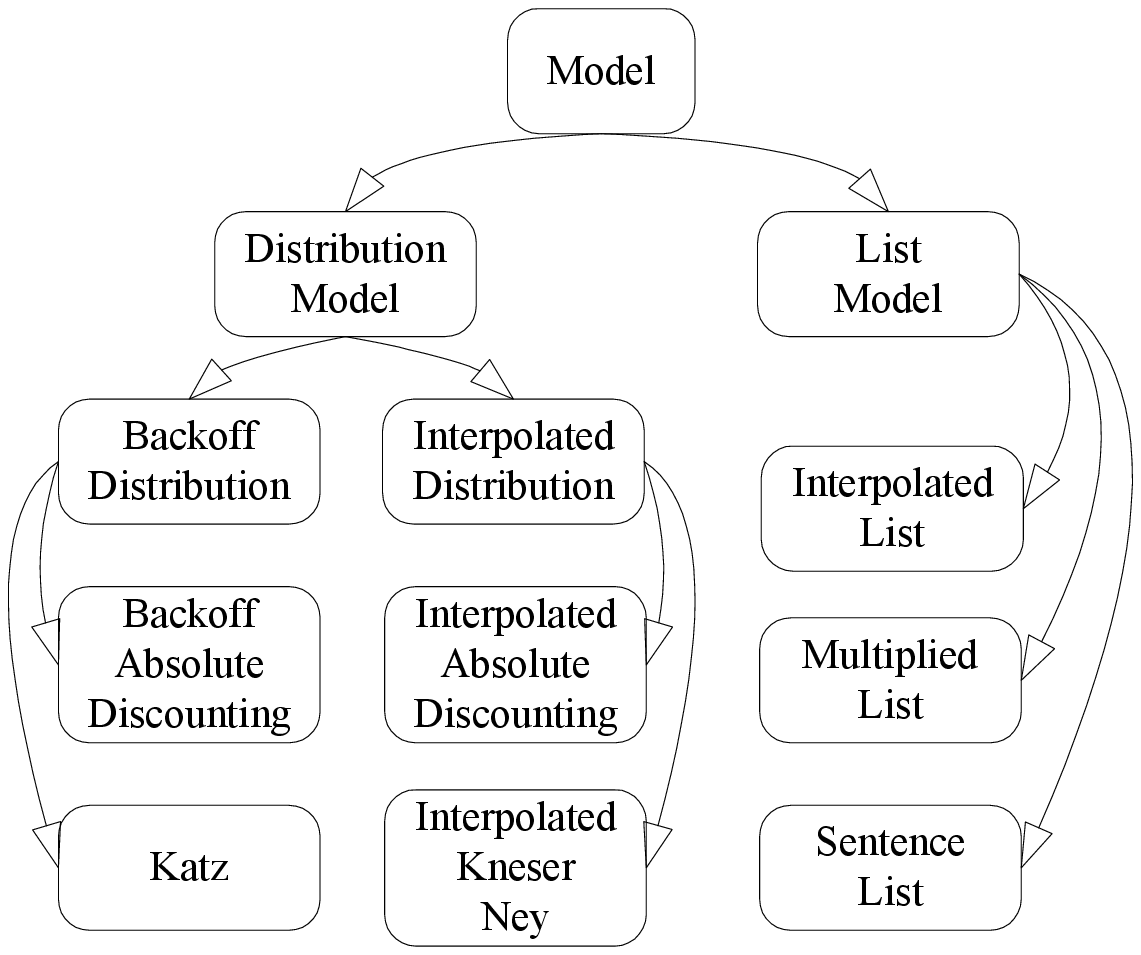}$$
\caption{Class Structure}
\label{fig:classes}
\end{figure}

The DistributionModels are, for instance, Kneser-Ney smoothed models,
Katz smoothed models, simple interpolated models, and absolute
discounting models.  Even within these models, there is some
substructure; for instance, absolute discounting backoff models, Kneser-Ney
backoff models, and Katz models can share most of their code, while
interpolated Kneser-Ney models and interpolated absolute discounting
models can also share code.

The other type of model is a model containing other models.  For
instance, an InterpolateModel contains a list of other models, whose
probability it interpolates.  For clustered models of the form
$P(Z|XY) \times P(z|Z)$, there is a model that contains a list of
other models, and multiplies their probabilities.  There is also a
SentenceInterpolateModel that contains a list of other models, and
interpolates them at the sentence level, instead of the word level.
All of these container type models have similar behavior in many ways,
and much of their code can be shared as well.

It was in part this architecture, providing a set of basic building
blocks and tools for combining them, that allowed so many different
experiments to be performed.

\subsection{Parameter search}

Another key tool was our parameter search engine.  In many cases, it
is possible to find EM algorithms to estimate the various parameters
of models.  In other cases it is possible to find good theoretical
approximate closed form solutions, using leaving-one-out, such as for
discounting parameters.  But for the large number of different models
used here, this approach would have lead to endless coding, and, for
some of the more complex model types, perhaps to suboptimal parameter
settings.  Instead, we used Powell's method \mycite{Press:88a} to
search all parameters of our models.  This had many advantages.
First, there are some parameters, such as $\beta$ in Katz smoothing,
that are hard to find except through a general search procedure.
Second, it meant that we could optimize all parameters jointly, rather
than separately.  For instance, consider a skipping model that
interpolates two models together.  We could, traditionally, optimize
the smoothing parameters for one model, optimize the smoothing
parameters for the other model, and then find the optimal
interpolation interpolation of the two models, without reestimating
the smoothing parameters.  Instead, we find jointly optimal settings.
Of course, like all gradient descent techniques, the parameters we
find are probably only locally optimal, but that is still an
improvement over traditional techniques.

We used certain tricks to speed up and improve the parameter search.
While in theory Powell search should always find local minima, it can
run into certain problems.  Consider interpolating two distributions,
$\lambda P_1(w) + (1-\lambda)P_2(w)$.  Imagine that the parameters for
$P_2$ have not yet been optimized.  Now, the Powell search routine
optimizes $\lambda$.  Since $P_2$ is awful, $\lambda$ is set at 1 or
nearly 1.  Eventually, the routine searches the parameters of $P_2$,
but, finding essentially no effect from changing these parameters,
quickly gives up searching them.  Instead, we search interpolation
parameters last, and start with all distributions interpolated evenly.
This ensures that each distribution has its parameters optimized.

Furthermore, searching the parameters of $P_2$ interpolated with $P_1$
is slow, since both distributions must be evaluated.  We typically
search the parameters of each one separately first, potentially
halving the computation.  Only once each one is optimized do we
optimize them jointly, but since they are often already nearly optimal
this search goes faster.  In the case of two distributions, the
savings may be small, but in the case of 10, it may be almost an order
of magnitude.  This technique must however be used with care, since
sometimes the optimal settings of one distribution differ wildly from
their settings when combined.  In particular, with sentence mixture
models, we found that an individual mixture component might
significantly oversmooth, unless it was trained jointly.

\subsection{Clustering}
\label{sec:clusterdetails}

There is no shortage of techniques for generating clusters, and there
appears to be little evidence that different techniques that optimize
the same criterion result in a significantly different quality of
clusters.  We note, however, that different algorithms may require
significantly different amounts of run time.  We used several
techniques to speed up our clustering significantly.  

The basic criterion we followed was to minimize entropy.  In
particular, assume that the model we are using is of the form
$P(z|Y)$; we want to find the placement of words $y$ into clusters $Y$
that minimizes the entropy of this model.  This is typically done by
swapping words between clusters whenever such a swap reduces the
entropy.

The first important approach we took for speeding up clustering was to
use a top-down approach.  We note that agglomerative clustering
algorithms -- those which merge words bottom up -- may require
significantly more time than top-down, splitting algorithms.  Thus,
our basic algorithm is top-down.  However, at the end, we perform four
iterations of swapping all words between all clusters.  This final
swapping is typically the most time consuming part of the algorithm.

Another technique we use is Buckshot \mycite{Cutting:92a}.  The basic
idea is that even with a small number of words, we are likely to have
a good estimate of the parameters of a cluster.  So, we proceed top
down, splitting clusters.  When we are ready to split a cluster, we
randomly pick a few words, and put them into two random clusters,
and then swap them in such a way that entropy is decreased, until
convergence (no more decrease can be found).  Then we add a few more
words, typically $\sqrt(2)$ more, and put each into the best bucket,
then swap again until convergence.  This is repeated until all words
in the current cluster have been added and split.  We haven't tested
this particularly thoroughly, but our intuition is that it should lead
to large speedups.

We use one more important technique that speeds computations, adapted
from earlier work of \namecite{Brown:90a}.  We attempt to minimize the
entropy of our clusters.  Let $v$ represent
words in the vocabulary, and $W$ represent a potential cluster.  We
minimize
$$
\sum_{v} C(Wv) \log P(v|W)
$$
The inner loop of this minimization considers adding (or removing) a
word $x$ to  cluster $W$.  What will the new entropy be?  On it's face,
this would appear to require computation proportional to the
vocabulary size to recompute the sum.  However, letting the new
cluster, $W+x$ be called $X$,
\begin{equation}
\sum_v C(Xv) \log P(v|X)  = \!\!\!\!\!\!\!\! \sum_{v|C(xv) \neq 0} \!\! \!\!\!\! C(Xv) \log P(v|X) +
\!\!\!\!\!\!\! \sum_{v|C(xv)=0} \!\! \!\!\!\! C(Xv) \log P(v|X) 
\label{eqn:wholeclusters}
\end{equation}
The first summation in Equation \ref{eqn:wholeclusters} can be computed
relatively efficiently, in time proportional to the number of
different words that follow $x$; it is the second summation that
needs to be transformed:
\begin{eqnarray}
\lefteqn{\sum_{v|C(xv)=0} C(Xv) \log P(v|X)} \nonumber \\
& = & \sum_{v|C(xv)=0} C(Wv) \log \left( P(v|W) \frac{C(W)}{C(X)} \right)  \nonumber \\
& = & 
\sum_{v|C(xv)=0} C(Wv) \log P(v|W)  +
 \left( \log \frac{C(W)}{C(X)} \right)  \sum_{v|C(xv)=0}  C(Wv) 
  \label{eqn:partialcluster}
\end{eqnarray}
Now, notice that 
\begin{equation}
\sum_{v|C(xv)=0} \!\!\!\! C(Wv) \log P(v|W)  = \sum_v C(Wv) \log P(v|W) - \!\! \sum_{v|C(xv) \neq 0} \!\!\!\! C(Wv) \log P(v|W)
\label{eqn:clustera}
\end{equation}
and that
\begin{equation}
\sum_{v|C(xv)=0}  C(Wv) = \left( C(W) - \sum_{v|C(xv) \neq 0} C(Wv)  \right)
\label{eqn:clusterb}
\end{equation}
Substituting Equations \ref{eqn:clustera} and \ref{eqn:clusterb} into Equation \ref{eqn:partialcluster}, we get
\begin{eqnarray*}
\lefteqn{\sum_{v|C(xv)=0} C(Xv) \log P(v|X)} \\
&=&
\sum_v C(Wv) \log P(v|W) - \sum_{v|C(xv) \neq 0} C(Wv) \log P(v|W)  \\
&&+
\left( \log \frac{C(W)}{C(X)} \right)  \left( C(W) - \sum_{v|C(xv) \neq 0} C(Wv)  \right)
\end{eqnarray*}
Now, notice that $\sum_v C(Wv) \log P(v|W)$ is just the old entropy,
before adding $x$.  Assuming that we have precomputed/recorded this
value, all the other summations only sum over words $v$ for which
$C(xv) > 0$, which, in many cases, is much smaller than the vocabulary
size.


Many other clustering techniques \mycite{Brown:90a} attempt to
maximize $\sum_Y,Z P(YZ) \log \frac{P(Y|Z)}P(Z)$, where the same
clusters are used for both.  The original speedup formula uses this
version, and is much more complex to minimize.  Using different
clusters for different positions not only leads to marginally lower
entropy, but also leads to simpler clustering.

\subsection{Smoothing}

Although all smoothing algorithms were reimplemented for this
research, the details closely follow \namecite{Chen:99a}.  This
includes our use of additive smoothing of the unigram distribution for
both Katz smoothing and Kneser-Ney smoothing.  That is, we found a
constant $b$ which was added to all unigram counts; this leads to
better performance in small training-data situations, and allowed us
to compare perplexities across different training sizes, since no
unigram received 0 counts, meaning 0 probabilities were never
returned.  For Katz smoothing, we found a maximum count to discount,
based on when data sparsity prevented good estimates of the discount.
As is typically done, we corrected the lower discounts so that the
total probability mass for 0 counts was unchanged by the cutoff, and
we also included a parameter $\beta$, which was added to a
distribution whenever no counts were discounted.

We used a novel technique for getting the counts for Katz smoothing.
As described in Appendix \ref{sec:storeonly}, we do not record all
counts, but only those needed for our experiments.  This is
problematic for Katz smoothing, where we need counts of counts ($n_1$,
$n_2$, etc.) in order to compute the discounts.  Actually, all we
really need is the ratio between these counts.  We use a statistical
sampling technique, in which we randomly sample 25,000 to 50,000
counts (not word histories, but actual counts); this allows us to
accurately estimate the needed ratios, with much less storage.

} 
\end{document}